\documentclass{article} % For LaTeX2e
\usepackage{iclr2026_conference,times}

\usepackage{amsmath,amsfonts,bm}

\def\eqref#1{equation~\ref{#1}}
\def\1{\bm{1}}

\def\va{{\bm{a}}}

\def\vq{{\bm{q}}}

\def\vs{{\bm{s}}}

\def\vu{{\bm{u}}}

\def\vy{{\bm{y}}}

\DeclareMathAlphabet{\mathsfit}{\encodingdefault}{\sfdefault}{m}{sl}
\SetMathAlphabet{\mathsfit}{bold}{\encodingdefault}{\sfdefault}{bx}{n}

\newcommand{\softmax}{\mathrm{softmax}}

\usepackage{url}
\definecolor{iccvblue}{rgb}{0.21,0.49,0.74}
\usepackage[
    pagebackref,
    breaklinks,
    colorlinks,
    citecolor=iccvblue
]{hyperref}
\usepackage{graphicx}
\usepackage{algorithm,algcompatible,amsmath}
\usepackage{algpseudocode}
\usepackage{amssymb}
\usepackage{booktabs}
\usepackage{siunitx}
\usepackage{multirow}
\usepackage{pdflscape}
\usepackage{rotating}
\usepackage{xspace}
\usepackage{wrapfig}
\usepackage[dvipsnames,table]{xcolor}
\usepackage{pgfplots}
\usepgfplotslibrary{groupplots,colormaps,colorbrewer}
\usepackage{pgfplotstable}
\usetikzlibrary{plotmarks}
\usepackage{tikz}
\usepackage{subcaption}
\usepackage[section]{placeins}  % inserts \FloatBarrier at each \section
\usepackage{makecell}
\usepackage{titletoc}

\definecolor{qcol1}{RGB}{247,237,123}
\definecolor{qcol2}{RGB}{244,181,110}
\definecolor{qcol3}{RGB}{239,115,115}
\definecolor{qcol4}{RGB}{125,164,242}
\definecolor{qcol5}{RGB}{135,224,242}
\definecolor{qcol6}{RGB}{124,239,143}
\definecolor{qcol7}{RGB}{195,137,237}
\definecolor{qcol8}{RGB}{244,146,195}

\makeatletter
\@namedef{ver@everyshi.sty}{}
\makeatother

\PassOptionsToPackage{dvipsnames,svgnames,x11names,table}{xcolor}
\usepackage{tikz}
\usetikzlibrary{arrows.meta,shapes,calc,matrix,fit,positioning,backgrounds,decorations.markings,fadings}
\usepackage{pgfplots}
\usepgfplotslibrary{fillbetween}
\usepackage{pgfplotstable}
\pgfplotsset{compat=1.18}
\usepackage{xstring}

\usepgfplotslibrary{external}
\IfBeginWith*{\jobname}{fig/extern/}{\finalcopy}{}

\tikzstyle{every picture}+=[
	remember picture,
	every text node part/.style={align=center},
	every matrix/.append style={ampersand replacement=\&},
]
\tikzstyle{tight} = [inner sep=0pt,outer sep=0pt]
\tikzstyle{node}  = [draw,circle,tight,minimum size=12pt,anchor=center]
\tikzstyle{op}    = [draw,circle,tight]
\tikzstyle{dot}   = [fill,draw,circle,inner sep=1pt,outer sep=0]
\tikzstyle{pt}    = [fill,draw,circle,inner sep=1.5pt,outer sep=.2pt]
\tikzstyle{box}   = [draw,rectangle,inner sep=3pt]
\tikzstyle{high}  = [black!60]
\tikzstyle{group} = [high,box,opacity=.5]
\tikzstyle{dim1}  = [fill opacity=.3,text opacity=1]
\tikzstyle{dim2}  = [fill opacity=.5,text opacity=1]
\tikzstyle{dim3}  = [fill opacity=.7,text opacity=1]
\tikzstyle{rectc} = [tight,transform shape]
\tikzstyle{rect}  = [rectc,anchor=south west]

\tikzset{every mark/.append style={solid}}
\pgfplotsset{%smooth,
	grid=both, width=\columnwidth, try min ticks=5,
	every axis/.append style={font=\small},
	every axis plot/.append style={thick,mark=none,mark size=1.8,tension=0.18},
	legend cell align=left, legend style={fill opacity=0.8},
	nodes near coords math/.style={
		nodes near coords={\pgfmathprintnumber[assume math mode=true]{\pgfplotspointmeta}},
	},
}

\pgfplotsset{
	dash/.style={mark=o,dashed,opacity=0.6},
	dott/.style={mark=o,dotted,opacity=0.6},
	nolim/.style={enlargelimits=false},
	plain/.style={every axis plot/.append style={},nolim,grid=none},
}

\pgfdeclarelayer{bg4}
\pgfdeclarelayer{bg3}
\pgfdeclarelayer{bg2}
\pgfdeclarelayer{bg1}
\pgfdeclarelayer{fg1}
\pgfdeclarelayer{fg2}
\pgfdeclarelayer{fg3}
\pgfdeclarelayer{fg4}
\pgfsetlayers{bg4,bg3,bg2,bg1,main,fg1,fg2,fg3,fg4}

\pgfkeys{/tikz/.cd, aspect/.store in=\aspect, aspect=1}
\pgfkeys{/tikz/.cd, depth/.store in=\depth, depth=.5}
\pgfkeys{/tikz/.cd, stepx/.store in=\step, stepx=1}

\tikzstyle{geom} = [line join=bevel,aspect=1,depth=.5,z={(\depth*\aspect,\depth)}]
\tikzstyle{wire} = [geom,draw,thick]

\def\cx[#1,#2,#3]{#1}
\def\cy[#1,#2,#3]{#2}
\def\cz[#1,#2,#3]{#3}
\def\ex[#1,#2,#3]{#1,0,0}
\def\ey[#1,#2,#3]{0,#2,0}
\def\ez[#1,#2,#3]{0,0,#3}

\DeclareRobustCommand{\equ}[1]{(\ref{equ:#1})\xspace}

\DeclareRobustCommand{\real}{\ensuremath{\mathbb{R}}}
\DeclareRobustCommand{\lone}{\ensuremath{\ell_1}}
\input{tex/defn}

\title{Attention, Please! Revisiting Attentive \\ Probing through the lens of Efficiency}

\author{
\textbf{Bill Psomas}$^{1}$\thanks{Equal contribution} \quad 
\textbf{Dionysis Christopoulos}$^{2}$\footnotemark[1] \quad
\textbf{Eirini Baltzi}$^{2}$ \quad
\textbf{Ioannis Kakogeorgiou}$^{7}$ \quad \\
\textbf{Tilemachos Aravanis}$^{1}$ \quad
\textbf{Nikos Komodakis}$^{3,4,5}$ \quad
\textbf{Konstantinos Karantzalos}$^{2}$ \quad \\
\textbf{Yannis Avrithis}$^{6}$ \quad
\textbf{Giorgos Tolias}$^{1}$ \\
[0.5em]
$^1$VRG, FEE, Czech Technical University in Prague \quad
$^2$National Technical University of Athens \\
$^3$University of Crete \quad
$^4$Archimedes, Athena RC \quad 
$^5$IACM-FORTH \quad 
$^6$IARAI \quad \\
$^7$IIT, NCSR ``Demokritos''
\vspace{-14pt}
}

\iclrfinalcopy % Uncomment for camera-ready version, but NOT for submission.
\begin{document}

\maketitle

\begin{abstract}
As fine-tuning becomes impractical at scale, probing is emerging as the preferred evaluation protocol. However, standard linear probing can understate the capability of models whose pre-training optimizes local representations rather than an explicit global representation. This motivates attentive probing, an alternative that uses attention to selectively aggregate patch-level features. Despite growing adoption, attentive probing is still underexplored: existing approaches are often over-parameterized and computationally inefficient. In this work, we revisit attentive probing through the lens of the accuracy \vs parameter-efficiency trade-off. We present the first comprehensive study of existing methods, analyzing their design choices and benchmarking their performance. Building on these insights, we propose \our (\OUR), a lightweight yet effective multi-query cross-attention mechanism that eliminates redundant projections and reduces the number of trainable parameters. Across multiple benchmarks and pre-training paradigms, \OUR consistently outperforms linear probing and previous attentive probing methods, and remains effective when combined with parameter-efficient fine-tuning. Beyond evaluation, our analysis uncovers emerging properties of \OUR, including complementary attention maps, which open new directions for leveraging probing beyond protocol design. Project page: \url{https://vrg.fel.cvut.cz/ep/}. 
\end{abstract}

\section{Introduction}
\label{sec:intro}

The past few years have witnessed remarkable progress in \emph{representation learning}, with pre-training paradigms ranging from self-supervised learning~\citep{simclr, caron2021emerging, attmask}, to vision–language models~\citep{radford2021clip, jia2021scaling}, and auto-regressive architectures~\citep{aim,aimv2}. These diverse approaches share a \emph{common goal}: learning rich, transferable visual representations that minimize reliance on task-specific labels and scale to large datasets. Evaluating the quality of such pre-trained representations is therefore \emph{central} to measuring progress. Conventional evaluation protocols include $k$-NN classification, linear probing (LP), and full fine-tuning (FT). While $k$-NN and LP assess the quality of the learned representations under a \emph{frozen} backbone, FT measures the utility of pre-training as \emph{initialization} for downstream tasks. Although FT achieves the highest performance, it is increasingly viewed as unsustainable and prohibitive at scale~\citep{xin2024parameter, shuttleworth2024lora, zou2023comprehensive}. As a result, probing is emerging as a practical evaluation protocol~\citep{dinov2, v-jepa, capi}.

However, probing protocols remain \emph{misaligned} with many pre-training approaches. Standard LP typically relies on a single \emph{global} representation. \eg, the \cls, which is well-suited for architectures trained with global objectives, but poorly reflects the potential of models such as masked image modeling~\citep{mae}, autoregressive~\citep{aim}, or diffusion~\citep{sit}, where valuable information is distributed across \emph{local} representations. This gap motivates the rise of \emph{attentive probing}~\citep{capi, chen2023cae, v-jepa}. Despite its promise, attentive probing remains underexplored. Existing methods vary significantly in design, often suffering from \emph{excessive parameterization} and \emph{inefficiency}. Moreover, the connection between how attention aggregates features and why it improves predictive performance remains unclear.

In this work, we address these limitations by conducting the first comprehensive study of attentive probing, revisiting its design through the lens of the \emph{accuracy \vs parameter-efficiency trade-off}. We introduce a unified framework that encompasses a wide range of attention-based aggregation methods—including those proposed for probing~\citep{aim, v-jepa, capi} and others from unrelated tasks~\citep{psomas2023simpool, delf, rymarczyk2021abmilp}. Through this framework, we derive \emph{\our} (\OUR), a simple multi-query cross-attention mechanism that eliminates redundant projections, reduces parameter count and computational cost, while matching or surpassing prior state-of-the-art performance. Moving beyond the parameter-efficient probing (PEP), we compare \OUR against parameter-efficient fine-tuning (PEFT) methods. We find that \OUR remains beneficial even when PEFT is allowed: combining EP with LoRA~\citep{lora} yields configurations that dominate both pure probing and pure LoRA.

%\OUR strikes a strong accuracy–efficiency balance, offering a lightweight yet expressive alternative for probing frozen models.

Beyond efficiency and accuracy, a common ingredient of both existing methods and \OUR is the use of multiple independent \emph{attention predictors} (\eg, heads or queries). We show that a predictor’s contribution to classification accuracy correlates with its localization quality: predictors with sharper, foreground-focused attention drive larger accuracy gains. Rather than resorting to shortcut learning, \eg, leveraging background cues like water to classify a ``fish", \OUR's predictors consistently attend to the object, improving interpretability, robustness, and performance. Notably, the attention maps of \OUR are more diverse and complementary than those of existing methods, with different predictors specializing in distinct object regions.  Our contributions are threefold:

%We establish a connection between the contribution of each predictor to the classification accuracy and its \emph{localization quality}. Our analysis shows that, rather than resorting to shortcut learning, \eg, leveraging background cues like water to classify a fish, the predictors of \OUR focus on foreground object(s), enhancing interpretability, robustness, and performance. Notably, each attention predictor in \OUR specializes in distinct object regions, ensuring complementary feature extraction and a more structured representation.

% We validate our findings through extensive experiments on seven classification benchmarks, across four MIM frameworks, using backbones of varying size. Beyond its original motivation for MIM, \OUR generalizes well to other pre-training paradigms, as shown by results on joint-embedding, hybrid, and vision-language models. 

% TODO: Maybe put this sentence back again?
%\OUR also delivers strong gains over LP in low-shot and layer-wise settings.

%We validate our findings through extensive experiments on seven classification benchmarks, across four MIM frameworks (MAE~\citep{mae}, SimMIM~\citep{simMIM}, BEiTv2~\citep{beitv2}, CAPI~\citep{capi}) using backbones of varying size. Beyond its original motivation for MIM, \OUR generalizes well to other pretraining paradigms, as shown by results on joint-embedding (\eg, BYOL~\citep{byol}), hybrid (\eg, DINOv2~\citep{dinov2}), and vision-language models (\eg, CLIP~\citep{radford2021clip}). Notably, \OUR also delivers strong gains over LP in low-shot and layer-wise settings.
\vspace{-8pt}
\begin{enumerate}\setlength{\itemsep}{0pt}\setlength{\topsep}{2pt}
    \item We conduct the first systematic benchmark and analysis of attentive probing methods across diverse pre-training paradigms, comparing their accuracy, efficiency, and design choices.
    \item We introduce \emph{\our} (\OUR), which achieves state-of-the-art accuracy, while bringing substantial gains in compute, memory, and parameter efficiency.
    \item We uncover a correlation between spatial localization and predictive performance, and show that \OUR produces diverse, complementary, and interpretable attention maps.
\end{enumerate}
\section{Related Work}
\label{sec:related}

%\paragraph{Evaluation protocols, pre-training paradigms.}

Self-supervised learning (SSL) has transformed visual representation learning, with evaluation typically performed via (i) \emph{$k$-NN} on frozen features, (ii) \emph{linear probing} (LP) using a shallow classifier on a frozen encoder, or (iii) \emph{fine-tuning} (FT) the entire model. Although FT achieves the highest accuracy, it is \emph{computationally expensive}, motivating the evaluation under frozen backbones. Two dominant SSL paradigms are joint embedding architectures (JEA) and masked image modeling (MIM). JEA methods (\eg, DINO~\citep{caron2021emerging}) contrast or cluster augmentations to learn global representations via a \cls token or pooled features. In contrast, MIM methods (\eg, MAE~\citep{mae}) reconstruct masked regions, yielding localized, patch-distributed representations. This global \vs local distinction affects evaluation: LP is effective for JEA~\citep{caron2021emerging} but under-performs for MIM~\citep{mae,przewikezlikowski2024beyond}, where discriminative information is not concentrated in a single token.

Beyond SSL, \emph{vision–language models} (VLMs) pre-train on web-scale image–text corpora (\eg, LAION~\citep{schuhmann2022laion}) and optimize cross-modal alignment (\eg, CLIP~\citep{openclip,radford2021clip}). Although these models expose strong global descriptors, much of the signal remains distributed across patch tokens, making attentive aggregation appealing at probe time. Likewise, \emph{auto-regressive} (AR) families (\eg, AIM/AIMv2~\citep{aim,aimv2}) and \emph{diffusion-based} transformers (\eg, DiT~\citep{dit}) are primarily trained for generation (next-token prediction or denoising), not representation learning per se; nevertheless, their frozen features can be probed to assess representation quality. In all these cases, protocols that assume a single discriminative token may under-utilize the information spread across patches.

To address this, recent work explores \emph{attentive probing}~\citep{aim,chen2023cae,v-jepa,capi}, which learns attention to selectively aggregate patch tokens into informative descriptors for a linear classifier. While methods like AIM~\citep{aim}, CAE~\citep{chen2023cae}, and V-JEPA~\citep{v-jepa} adopt this idea, no unified evaluation exists. We fill this gap with a comprehensive benchmark and introduce a novel attention mechanism achieving a strong accuracy–efficiency trade-off. Additional related work appears in~\autoref{sec:more_related}.

% Additional related work appears in~\autoref{sec:more_related}.

%This global \vs local distinction affects evaluation: LP is effective for JEA~\citep{caron2021emerging} but underperforms for MIM~\citep{mae, przewikezlikowski2024beyond}, where discriminative information is not concentrated in a single token. Consequently, FT remains the preferred strategy for MIM~\citep{mae, simMIM}. To overcome this, recent work explores \emph{attentive probing}~\citep{aim, chen2023cae, v-jepa, capi}, where attention is used to aggregate patch tokens into informative descriptors. While methods like AIM~\citep{aim}, CAE~\citep{chen2023cae}, and V-JEPA~\citep{v-jepa} adopt this idea, no unified evaluation exists. We fill this gap with a comprehensive benchmark and introduce a novel attention mechanism achieving a strong accuracy–efficiency trade-off.

%---------------------------------------------------------------------

\section{Method}
\label{sec:method}

%\subsection{Preliminaries}
%------------------------------------------------------------------------------

\subsection{Attentive Pooling}
\label{sec:att-pool}

\paragraph{Preliminaries.} 

Let $X \in \real^{D_i \times N}$ be the \emph{feature matrix} obtained from a \emph{pre-trained} and \emph{frozen} ViT backbone, where $D_i$ is the number of feature channels  and $N = W \times H$ the number of features, one per image patch across the spatial dimensions $W \times H$. Given the \emph{input features} $X$, the goal is to generate an \emph{output image-level feature} $\vy \in \real^{D_o}$ by applying an \emph{attentive pooling} mechanism. The output feature is used to train a $C$-way linear classifier with $D_o (C+1)$ parameters.

We consider $M$ \emph{attention predictors}, to be discussed in \autoref{sec:att-pred}. For each predictor $j \in \{1, \ldots, M\}$, let $\va_j \in \real^N$ be the $\lone$-normalized \emph{attention vector} it generates. Each vector, reshaped to $W \times H$, is an attention map indicating the locations on which the predictor focuses. Let $V \in \real^{D_o \times N}$ be the \emph{value features}, commonly obtained by a linear transformation $V=W_V X$, where $W_V \in \real^{D_o \times D_i}$ is a learnable \emph{projection matrix}.

Let the output feature $\vy$, value features $V$ and projection matrix $W_V$ be partitioned into $M$ subvectors / submatrices according to
\begin{equation}
	\vy = \begin{bmatrix} \vy_1 \\ \vdots \\ \vy_M \end{bmatrix},
	V =  \begin{bmatrix} V_1 \\ \vdots \\ V_M \end{bmatrix},
	W_V =  \begin{bmatrix} W_{V_1} \\ \vdots \\ W_{V_M} \end{bmatrix},
\end{equation}
with $\vy_j \in \real^{d_o}$, $V_j \in \real^{d_o \times N}$, $W_{V_j} \in \real^{d_o \times D_i}$ and $d_o = \frac{D_o}{M}$.

The attentive pooling operation is then given by
\begin{equation}
	\vy_j = V_j \va_j = W_{V_j} X \va_j.
	\label{equ:att-pool}
\end{equation}
Each attention predictor is responsible for the weighted pooling of $N$ features into a $d_o$-dimensional subspace of the final representation space. In the following, we explore existing and novel ways for designing these attention predictors. We focus on the number of additional parameters to be learnt on top of the frozen backbone and the computational complexity of the pooling operation.

\subsection{Attention Predictors}
\label{sec:att-pred}

\paragraph{Multi-Head Cross-Attention (MHCA).}

A vanilla approach is to perform multi-head cross-attention between the input features and an \emph{input vector} $\vu \in \real^{D_i}$, where each head corresponds to a separate attention predictor. The \emph{query feature} $\vq \in \real^{D_a}$ and \emph{key features} $K \in \real^{D_a \times N}$ are obtained by linear transformations $\vq = W_Q \vu$, $K = W_K X$ with projection matrices $W_Q, W_K \in \real^{D_a \times D_i}$.

Let the query feature $\vq$ and projection matrix $W_Q$ be partitioned into $M$ subvectors / submatrices according to
\begin{equation}
	\vq =  \begin{bmatrix} \vq_1 \\ \vdots \\ \vq_M \end{bmatrix},
	W_Q =  \begin{bmatrix} W_{Q_1} \\ \vdots \\ W_{Q_M} \end{bmatrix},
\end{equation}
with $\vq_j = W_{Q_j} \vu \in \real^{d_a}$, $W_{Q_j} \in \real^{d_a \times D_i}$ and $d_a = \frac{D_a}{M}$. Similarly, let the key features $K$ and projection matrix $W_K$ be partitioned according to
\begin{equation}
	K =  \begin{bmatrix} K_1 \\ \vdots \\ K_M \end{bmatrix},
	W_K =  \begin{bmatrix} W_{K_1} \\ \vdots \\ W_{K_M} \end{bmatrix},
\end{equation}
with $K_j = W_{K_j} X \in \real^{d_a \times N}$ and $W_{K_j} \in \real^{d_a \times D_i}$.

The attention vector for head $j$ is then given by
\begin{equation}
	\va_j = \softmax(\hat{\va}_j)
	\label{equ:softmax}
\end{equation}
with
\begin{equation}
	\hat{\va}_j = K_j^\top \vq_j = (W_{K_j} X)^\top (W_{Q_j} \vu).
	\label{equ:mha}
\end{equation}

%------------------------------------------------------------------------------
\begin{figure*}[ht]
\vspace{-10pt}
    \centering
    \includegraphics[width=\textwidth]{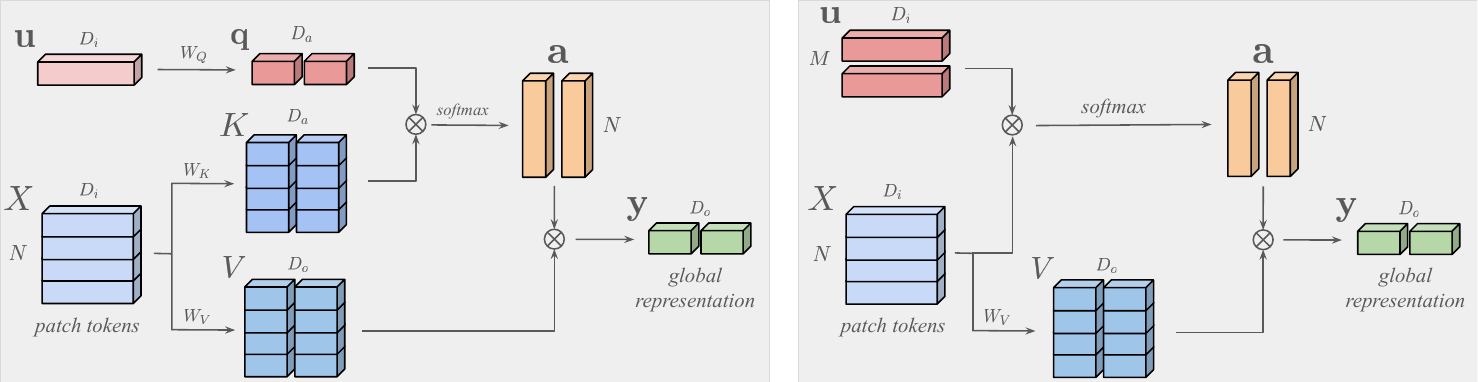}
    \vspace{-14pt}
    \caption{\emph{Comparison of multi-head cross-attention (MHCA, left) \vs our multi-query cross-attention (\OUR, right)}. 
    MHCA uses an input vector $\vu$ projected into query space and interacts with key features $K$ in (two) separate subspaces, each corresponding to an attention predictor.
    Attention predictor outputs $\va_j$ are used to aggregate value features $V$ into sub-vectors $\vy_j$, forming the final output $\vy$.
    In contrast, \OUR employs (two) \emph{learnable queries} $\vq_j$, one per attention predictor, to compute attention with input features directly in the full representation space. Attention predictor outputs $\va_j$ are used as in MHCA to perform the aggregation.
    }
    % \OUR reduces parameters and computation while obtaining strong representations.
    % \caption{\todo{Multi-head cross attention (MHCA) vs. transformation-free cross-attention, \ie the mechanism of \our. MHCA computes attention using a single \emph{query} \( \mathbf{q} \in \mathbb{R}^{D_a} \) derived from an \emph{input vector} \( \mathbf{u} \in \mathbb{R}^{D_i} \) via a learned projection \( W_Q \). Attention scores \( \mathbf{a} \) are obtained via a dot product with the \emph{key features} \( K = W_K X \) and applied to the \emph{value features} \( V = W_V X \) to produce the output \( \mathbf{y} \in \mathbb{R}^{D_o} \). This requires projection matrices (\( W_Q, W_K, W_V \)), increasing parameters and compute. Efficient Probing (\OUR) replaces the projected query with \emph{multiple learnable queries} \( Q \in \mathbb{R}^{M \times D_i} \), removing \( W_Q \) and computing attention directly on \( X \). This reduces parameters and computation while maintaining strong representation power. The output \( \mathbf{y} \) is obtained via efficient weighted aggregation over \( V = W_V X \).}}
    \label{fig:two_figs}
    \vspace{-6pt}
\end{figure*}
%------------------------------------------------------------------------------

That is, the input features $X$ and input vector $\vu$ are projected to $d_a$-dimensional subspaces where attention subvectors are computed via dot product followed by softmax normalization over patches. This attention predictor requires $D_a(2D_i+1)$ parameters and has complexity $\mathcal{O}(ND_a D_i)$. As discussed in \autoref{sec:var}, there are several existing methods that fit within this generic framework.

%------------------------------------------------------------------------------

\paragraph{MHCA with a learnable query.}

If we consider input vector $\vu$ to be learnable, then there is no need for the projection matrix $W_Q$ in \equ{mha}. Instead, we can set the query feature $\vq$ to be learnable, thus absorbing $W_Q$ and $\vu$:
\begin{equation}
	\hat{\va}_j = (W_{K_j} X)^\top \vq_j = X^\top W_{K_j}^\top \vq_j
\label{equ:mha-lrn-q}
\end{equation}
where the query feature $\vq_j \in \real^{d_a}$ is learnable.

We observe that $W_{K_j}^\top$ maps $\vq_j$ to the $D_i$-dimensional space of input features to compute the attention vector. Thus, standard MHCA ensures that each query subvector
% consequently each attention predictor (head),
is interacting with the full representation space of the input features, despite being defined in a smaller dimensional space. Using a learnable query feature directly simplifies the architecture, reduces the amount of computations and the number of parameters to $D_a (D_i+1)$.

In order to explore ways for reducing the total number of parameters and to better understand the role of key transformation, we simplify the architecture by removing it.
% the key transformation
Letting $W_K$ be fixed to the identity matrix, \equ{mha-lrn-q} becomes
\begin{equation}
	\hat{\va}_j = X_j^\top \vq_j
\label{equ:mha-lrn-k-id}
\end{equation}
where the feature matrix $X$ is partitioned into $M$ submatrices according to
\begin{equation}
	X = \begin{bmatrix} X_1 \\ \vdots \\ X_M \end{bmatrix},
\end{equation}
with $X_j \in \real^{d_i \times N}$ and $d_i = \frac{D_i}{M}$. We thus observe that the query feature only interacts  with a $d_i$-dimensional subspace of the input features for $M>1$, which is a \emph{limitation}. 
We experimentally verify that setting $W_K$ to identity matrix results in a noticeable performance drop.
Based on this observation, in the following, we revisit 
(\ref{equ:mha-lrn-q}) and design attention predictors 
that have less parameters and require less compute but do not experience a performance drop due to mathematical equivalence.
% In the following, we suggest a new way to design attention predictors that have less parameters, require less compute and overcome this limitation, making them perform better.

%------------------------------------------------------------------------------

\paragraph{Parameter-efficient Multi-Query Cross-Attention (MQCA).} Instead of using the key submatrices $W_{K_j}$ to project the query subvectors $\vq_j$ to the $D_i$-dimensional input feature space, we propose to learn $M$ effective query features $\vu_j := W_{K_j}^\top \vq_j \in \real^{D_i}$ in that space directly (\autoref{fig:two_figs}). Thus, $\vu_j$ absorbs $W_{K_j}$ and $\vq_j$ and attention prediction becomes
\begin{equation}
	\hat{\va}_j = X^\top \vu_j
	\label{equ:ours}
\end{equation}
for $j \in \{1, \ldots, M\}$. As a result, there are no projection matrices and there are no parameters other than the learnable query features $\vu_j$.

This choice reduces the number of additional parameters to be learned and saves from one more matrix-vector multiplication. In particular, it requires $D_i M$ parameters for the attention compared to $D_a(D_i+1)$ for \equ{mha}, while the number of operations drops to $ND_i M$ compared to $ND_a(D_i+1)$. Typically, $M$ is one to two orders of magnitude smaller than $D_i$ and $D_a$, which are commonly equal to each other, making the proposed approach more efficient in parameters and operations.

There is a connection between \OUR and \emph{slot attention}~\citep{locatello2020object}, where slots are also multiple vectors in the input feature space. To derive \OUR from slot attention, one needs to perform only a single iteration; remove LayerNorm, GRU and MLP; make slot vectors learnable rather than initialized at random; and concatenate the output features into a global representation of appropriate dimension. Thus, \OUR can be seen as a lightweight counterpart of slot attention, where the absence of interactions is compensated by the query features being learned.

%------------------------------------------------------------------------------

\subsection{Existing variants}
\label{sec:var}

We analyze existing methods as instances of the presented framework (\autoref{tab:pooling-variants}), and examine common variants, considering their relationship to the framework despite slight deviations. Additional methods considered in experiments are presented in~\autoref{sec:more_methods}.

\textbf{AbMILP}~\citep{rymarczyk2021abmilp} is the simplest variant. It fixes $W_K$ and $W_V$ to identity and is equivalent to MHCA with a learnable query feature framework in \equ{mha-lrn-q}, with a single head ($M=1$). It can also be seen as a special case of our proposed method in \equ{ours} with one learnable query feature, \ie $M=1$. AbMILP requires only $D_i$ parameters and computes attention with a single matrix-vector multiplication, but its performance is limited by the single head/query.

\textbf{AIM}~\citep{aim} is an instance of MHCA with a learnable query feature. It deviates from the generic framework by applying batch normalization on the input features. It does follow \equ{att-pool} and \equ{mha-lrn-q} with $M$ heads and $D_a = D_i = D_o$, but replaces $X$ by $\mathrm{BN}(X)$. Batch normalization introduces minor additional parameters and a slight computational overhead compared to the default variant of the framework.

\textbf{DELF}~\citep{delf} feeds each of the $N$ input features to a MLP whose output is a scalar attention value in $[0,1]$. It can be viewed as an instance of the MHCA with a learnable query feature and $M = 1$ with the following modifications. Key and value projection matrices share the same weights, \ie $W_K = W_V = W$, a non-linearity is introduced in equation \equ{mha-lrn-q} by $\hat{\va} = \relu(W X)^\top \vq $, where subscript $j$ is skipped due to $M=1$, and softmax in \equ{softmax} is replaced by element-wise softplus, $\va = \sigma_{p}(\hat{\va})$. In the context of DELF, the query feature $\vq$ can be seen as the parameter of a $1 \times 1$ convolutional layer. DOLG~\citep{yang2021dolg} adopts a similar design choice.
% DELF has fewer parameters and less complexity than the default variant of the framework.

\textbf{SimPool}~\citep{psomas2023simpool} can be seen as an instance of MHCA with a single head ($M = 1$) that uses a data-dependent input vector $\vu$, $W_V$ fixed to identity, and layer normalization on the input features. Specifically, the query feature is obtained as $\vq \!=\! W_Q \vu,\ \ \vu \!=\! \tfrac{1}{N}\,X^\top\,\mathbf{1}\ \ (M = 1)$. $X$ is replaced by $\LN(X)$ for key and value transforms. Compared to MHCA with a learnable query, SimPool saves $D_i$ parameters and has the same complexity.

\textbf{\vjepa.} The first part of \vjepa~\citep{v-jepa} is identical to the MHCA framework but applies layer normalization on the input features for key and value transforms, like SimPool. Its second part is an MLP with GeLU activation~\citep{gelu} and residual connections, making the overall process equivalent to a transformer block.

\begin{table}[h]
\centering
\scriptsize
\setlength{\tabcolsep}{2.5pt}
\renewcommand{\arraystretch}{1.15}
\caption{\emph{Attentive pooling variants as aligned algorithmic steps,} fitting the framework of Sec.~\ref{sec:att-pool}--\ref{sec:att-pred}. We list (i) how the query is formed, (ii) key/value transforms, (iii) attention predictor outputs, and (iv) pooling operation. $\sigma_{m}$: softmax, $\sigma_{p}$: softplus. $\mathrm{\phi}(x) := W_2\,\mathrm{GeLU}(W_1 x)$. \textcolor{blue}{blue}: learnable.}
%Acc@1 is on ImageNet-1k with MAE ViT-B.
\label{tab:pooling-variants}
\vspace{-6pt}
\begin{tabular}{l l l l l l}
\toprule
\textsc{Method}
& \textsc{Query source}
& \textsc{Key transform}
& \textsc{Value transform}
& \textsc{Attention}
& \textsc{Pooling}
%& \bf Params (attn + \(W_V\))
%& \bf Acc. 
\\ % MAE-B, IN-1k
\midrule
%MHCA$_{\text{van}}$
%& \(\vq_j{=}\textcolor{blue}{W_{Q_j}}\textcolor{blue}{\vu}\in\real^{d_a},\textcolor{blue}{\vu}\in\real^{D_i}\)
%& \(K_j{=}\textcolor{blue}{W_{K_j}}X\)
%& \(V_j{=}\textcolor{blue}{W_{V_j}}X\)
%& \(\va_j{=}\sigma_{m}(K_j^\top\vq_j)\)
%& \(\vy_j{=}V_j\va_j\)
%& \(D_a(2D_i{+}1)\ +\ D_oD_i\)
%& 75.3 
%\\
%
MHCA
& \(\textcolor{blue}{\vq_j}\in\real^{d_a}\)
& \(K_j{=}\textcolor{blue}{W_{K_j}}X\)
& \(V_j{=}\textcolor{blue}{W_{V_j}}X\)
& \(\va_j{=}\sigma_{m}(K_j^\top\vq_j)\)
& \(\vy_j{=}V_j\va_j\)
%& \(D_a(D_i{+}1)\ +\ D_oD_i\)
%& 75.2 
\\
AbMILP
& \(\textcolor{blue}{\vq}\in\real^{D_i}\)
& \(K{=}X\) %\ (\(W_K{=}I\))
& \(V{=}X\) %\ (\(W_V{=}I\))
& \(\va{=}\sigma_{m}(K^\top\vq)\)
& \(\vy{=}V\va\)
%& \(\color{gray}{\underbrace{D_i}_{\text{query}}} \ +\ \color{gray}{\underbrace{0}_{W_V}}\)
%& 71.7 
\\
AIM
& \(\textcolor{blue}{\vq_j}\in\real^{d_a}\)
& \(K_j{=}\textcolor{blue}{W_{K_j}}\mathrm{BN}(X)\)
& \(V_j{=}\textcolor{blue}{W_{V_j}}\mathrm{BN}(X)\)
& \(\va_j{=}\sigma_{m}(K_j^\top\vq_j)\)
& \(\vy_j{=}V_j\va_j\)
%& \(D_a(D_i{+}1)+D_oD_i\ +\ \color{gray}{2D_i\ (\mathrm{BN})}\)
%& 75.2 
\\
DELF
& \(\textcolor{blue}{\vq}\in\real^{D_i}\)
& \(K{=}\mathrm{ReLU}(\textcolor{blue}{W}X)\)
& \(V{=}\textcolor{blue}{W}X\)
& \(\va{=}\sigma_{m}(K^\top\vq)\)
& \(\vy{=}V\va\)
%& \(D_a(D_i{+}1)\ +\ D_oD_i\)
%& 72.9 
\\
SimPool
& \(\vq \!=\! \textcolor{blue}{W_Q} \vu\in\real^{D_i},\vu \!=\! \tfrac{1}{N}X^\top\mathbf{1}\)
& \(K{=}\textcolor{blue}{W_{K}}\mathrm{LN}(X)\)
& \(V{=}\mathrm{LN}(X)\) %\ (\(W_V{=}I\))
& \(\va{=}\sigma_{m}(K^\top\vq)\)
& \(\vy{=}V\va\)
%& \(\color{gray}{2D_i\ (\mathrm{LN})}\ +\ \color{gray}{0\ (W_V)}\)
%& 72.6 
\\
V\text{-}JEPA
& \(\vq_j{=}\textcolor{blue}{W_{Q_j}}\textcolor{blue}{\vu}\in\real^{d_a},\textcolor{blue}{\vu}\in\real^{D_i}\)
& \(K_j{=}\textcolor{blue}{W_{K_j}}\mathrm{LN}(X)\)
& \(V_j{=}\textcolor{blue}{W_{V_j}}\mathrm{LN}(X)\)
& \(\va_j{=}\sigma_{m}(K_j^\top\vq_j)\)
%& \(\vy{=}\mathrm{\textcolor{blue}{MLP}}(V_j\va_j)\)
& \(\vy{=}\mathrm{\textcolor{blue}{\phi}} (\textcolor{blue}{W_P}V_j\va_j)\)
%& \(D_a(D_i{+}1)+D_oD_i\ +\ \color{gray}{2D_i\ (\mathrm{LN})}\ +\ \color{gray}{\#\text{MLP}}\)
%& 74.1 
\\
EP (ours)
& \(\textcolor{blue}{\vq_j}\in\real^{D_i}\)
& \(K{=}X\) %\ (\(W_K{=}I\))
& \(V_j{=}\textcolor{blue}{W_{V_j}}X\)
& \(\va_j{=}\sigma_{m}(K^\top\vq_j)\)
& \(\vy_j{=}V_j\va_j\)
%& \(\color{blue}{D_iM}\ +\ D_oD_i\)
%& \textbf{75.6} 
\\
\bottomrule
\end{tabular}
%\vspace{-2mm}
%\begin{flushleft}
%\footnotesize
%\textbf{Notes.} Parameter counts exclude the $C$-way linear classifier. By default we set \(D_a{=}D_i{=}D_o\).
%AIM/SimPool/V-JEPA apply BN/LN as shown (each adds \(\approx 2D_i\) scale/shift params).
%EP replaces \(W_Q,W_K\) with \(M\) learnable queries \(\{\vu_j\}\), yielding \(\mathcal{O}(ND_iM)\) attention compute vs.\ \(\mathcal{O}(ND_a(D_i{+}1))\) for MHCA.
%\end{flushleft}
\end{table}

%Additional methods considered in experiments like CLIP~\citep{radford2021clip} and CoCa~\citep{yu2022coca} are further variants of the MHCA framework with slight differences compared to the variants presented above. For brevity, we present these in the supplementary material.

\section{Experiments}
\label{sec:exp}

\subsection{Experimental setup}
\label{sec:setup}

\textbf{Datasets.} We evaluate attentive probing across diverse image classification benchmarks, including ImageNet-1K (IN-1K)~\citep{imagenet}, CIFAR-100~\citep{cifar}, Places365~\citep{places365}, CUB-200~\citep{cub}, FGVC Aircraft~\citep{fgvc}, Stanford Cars~\citep{cars}, and Food-101~\citep{food}. More details in~\autoref{subsec:more_exp_setup}.

\textbf{Pre-training methods.} We conduct attentive probing experiments with frozen models from diverse pre-training paradigms: MIM (\eg, MAE and CAPI), JEA (\eg, BYOL and DINO), hybrid (\eg, iBOT and DINOv2), VLMs (\eg, CLIP and SigLIP), and generative (\eg, DiT and AIMv2). Model sizes vary from small (\eg ViT-S for MAE) to extra large (\eg DiT-XL for DiT).

\textbf{Pooling/probing methods.} We compare attentive probing against a diverse set of methods, covering different paradigms. First, we evaluate attentive poolings originally designed for probing, including AIM~\citep{aim}, CAE~\citep{chen2023cae}, CAPI~\citep{capi}, and V-JEPA~\citep{v-jepa}. Second, we include attentive poolings originally proposed in other contexts but applicable to probing, such as AbMILP~\citep{rymarczyk2021abmilp}, SimPool~\citep{psomas2023simpool}, CLIP~\citep{radford2021clip}, SigLIP~\citep{siglip, timm}, CoCa~\citep{yu2022coca}, CaiT~\citep{cait}, and DELF~\citep{delf}. Additionally, we include feature re-weighting methods like CBAM~\citep{woo2018cbam}, applying global average pooling to obtain the global descriptor. As baselines, we use \cls, which corresponds to standard linear probing when the backbone provides a classification token, and \gap, which serves as linear probing when no \cls token is available, but can also be interpreted as a baseline attentive probing method with uniform attention over patch tokens. To establish a reference, we also evaluate a ViT~\citep{vit} block, applying global average pooling to extract the global representation, and MHCA. All methods operate on the same input features—namely, the patch tokens extracted from the frozen backbone—ensuring a fair and consistent comparison. Unless otherwise stated, $D_o{=}D_i{=}D_a$.

\textbf{Parameter-efficient fine-tuning (PEFT) methods.} We compare attentive probing against PEFT methods on ImageNet-1K: LoRA~\citep{lora}, BitFit~\citep{bitfit}, and LayerNorm tuning~\citep{layernorm}.

\textbf{Evaluation protocols.} Attentive probing is performed for 90 epochs. We evaluate top-1 classification accuracy on the validation set of each dataset. Additionally, we compute the number of parameters and measure the FLOPs to assess computational efficiency and scalability.

%------------------------------------------------------------------------------
\begin{figure*}[ht]
    \centering
    \begin{subfigure}[t]{0.48\textwidth}
        \centering
        \caption{MAE ViT-B ImageNet-1K}
        % Scatter Plot
%\begin{figure}[ht]
%\centering
\begin{tikzpicture}
    \begin{axis}[
        width=7.3cm, % Adjust the width as needed
        height=6cm,  % Adjust the height as needed
        ylabel={top-1 accuracy (\%)},
        grid=major,
        enlargelimits=true,
        xmode=log,
%         log ticks with fixed point,
        legend pos=south east,
        legend style={font=\tiny}
    ]

%-----------------------------------------------------------------
% 1) Baseline Methods (Black)
%-----------------------------------------------------------------
\addplot[
    only marks,
    mark=*,
    draw=black,
    fill=black,
    mark size=1.5pt,
    nodes near coords,
    every node near coord/.append style={font=\tiny,anchor=south, yshift=-1pt, xshift=-8pt},
    point meta=explicit symbolic, % Set meta data source
]
table [
    meta=Method,
    x=Parameters,
    y=Accuracy,
    col sep=space,
] {
    Method Accuracy Parameters
    \textsc{[CLS]} 67.66 769000
    GAP 66.72 769000
    %CLIP 74.84 3282664
    %SigLIP 74.74 7854568
    %SimPool 72.64 1950184
    %AIM 75.18 1949416
    %CBAM 67.45 842828
    %CoCa 73.53 1851880
    CaiT 73.69 7860712
    ViT 75.25 7854568
    V-JEPA 74.12 7857640
    DELF 72.86 1361897
    AbMILP 71.74 769769
    %CAE 74.93 3135976
    %EP$_2$ 73.18 1360360
    EP$_4$ 74.14 1361896
    EP$_{16}$ 75.08 1368040
    EP$_{64}$ 75.58 1395688
};
\addlegendentry{$D_o = D_i$}

% 3) D_i/2 (blue)
%-----------------------------------------------------------------
\addplot[
    only marks,
    mark=*,
    draw=blue,
    fill=blue,
    mark size=1.5pt,
    nodes near coords,
    every node near coord/.append style={font=\tiny,anchor=south, yshift=-1pt, xshift=-8pt},
    point meta=explicit symbolic,
]
table [
    meta=Method,
    x=Parameters,
    y=Accuracy,
    col sep=space,
] {
    Method Accuracy Parameters
    EP$_2$ 72.72 681448
    EP$_4$ 73.56 682984
    EP$_{16}$ 74.41 692200
    EP$_{96}$ 74.86 753640
};
\addlegendentry{$D_o = D_i / 2$}

%-----------------------------------------------------------------
% 4) D_i/4 (orange)
%-----------------------------------------------------------------
\addplot[
    only marks,
    mark=*,
    draw=orange,
    fill=orange,
    mark size=1.5pt,
    nodes near coords,
    every node near coord/.append style={font=\tiny,anchor=south, yshift=-1pt, xshift=-8pt},
    point meta=explicit symbolic,
]
table [
    meta=Method,
    x=Parameters,
    y=Accuracy,
    col sep=space,
] {
    Method Accuracy Parameters
    EP$_2$ 71.31 341992
    EP$_4$ 72.05 343528
    EP$_{64}$ 72.88 389608
};
\addlegendentry{$D_o = D_i / 4$}

%-----------------------------------------------------------------
% 5) D_o/8 (purple)
%-----------------------------------------------------------------
\addplot[
    only marks,
    mark=*,
    draw=purple,
    fill=purple,
    mark size=1.5pt,
    nodes near coords,
    every node near coord/.append style={font=\tiny,anchor=south, yshift=-1pt, xshift=-8pt},
    point meta=explicit symbolic,
]
table [
    meta=Method,
    x=Parameters,
    y=Accuracy,
    col sep=space,
] {
    Method Accuracy Parameters
    EP$_2$ 68.99 172264
    EP$_8$ 69.59 176872
    EP$_{48}$ 70.27 207592
};
\addlegendentry{$D_o = D_i / 8$}

%-----------------------------------------------------------------
% Manual positions
%-----------------------------------------------------------------
\addplot[
    only marks,
    mark=*,
    draw=black,
    fill=black,
    mark size=1.5pt,
]
coordinates {
    (842828, 67.45) % CBAM
    (1950184, 72.64) % SimPool
    (1949416, 75.18) % AIM
    (1851880, 73.53) % CoCa
    %(2539240, 75.24) % MHCA
    (7854568, 74.74) % SigLIP
    (3282664, 74.84) % CLIP
    (3135976, 74.93) % CAE
};

\node[anchor=north, font=\tiny, yshift=1pt, xshift=7pt] at (842828, 67.45) {CBAM};
%\node[anchor=south, font=\tiny, yshift=0pt, xshift=10pt] at (2539240, 75.24) {MHCA};
\node[anchor=north, font=\tiny, yshift=0pt, xshift=9pt] at (1950184, 72.64) {SimPool};
\node[anchor=north, font=\tiny, yshift=1pt, xshift=8pt] at (1851880, 73.53) {CoCa};
\node[anchor=south, font=\tiny, yshift=-3pt, xshift=-9pt] at (7854568, 74.74) {SigLIP};
\node[anchor=south, font=\tiny, yshift=0pt, xshift=8pt] at (1949416, 75.18) {AIM/MHCA};
\node[anchor=north, font=\tiny, yshift=2pt, xshift=-9pt] at (3135976, 74.93) {CAE};
\node[anchor=north, font=\tiny, yshift=0pt, xshift=2pt] at (3282664, 74.84) {CLIP};

%-----------------------------------------------------------------
% Pareto Front
%-----------------------------------------------------------------
\path [name path=pareto]
    (769000, 64)
    -- (769769, 71.74)
    -- (1361897, 72.86)
    -- (1949416, 75.18)
    %-- (2539240, 75.24)
    -- (7854568, 75.25)
    -- (12000000, 75.25);

% Upper boundary (extend down to lowest y-value)
\path [name path=top]
    (12000000, 100) -- (12000000, 50);

% Fill the area to the right of the Pareto front
\addplot [fill=blue!30, opacity=0.3]
    fill between [of=pareto and top];

% --- Pareto front line ---
\addplot[thick, dashed, blue, mark=none] coordinates {
    (769000, 67.66)
    (769769, 71.74)
    (1361897, 72.86)
    (1949416, 75.18)
    %(2539240, 75.24)
    (7854568, 75.25)
};

    \end{axis}
\end{tikzpicture}
%\caption{Scatter plot of Top-1 Accuracy (\%) vs. Number of Parameters on ImageNet-1k.}
%\label{fig:accuracy_parameters_scatter}
%\end{figure}
        \label{fig:bench-imagenet}
    \end{subfigure}
    \vspace{-12pt}
    \begin{subfigure}[t]{0.48\textwidth}
        \centering
        \caption{MAE ViT-B Food-101}
        % Scatter Plot
%\begin{figure}[ht]
%\centering
\begin{tikzpicture}
    \begin{axis}[
        width=7.3cm, % Adjust width as needed
        height=6cm,  % Adjust height as needed
        grid=major,
        enlargelimits=true,
        xmode=log,
%         log ticks with fixed point,
        legend pos=south east,
        legend style={font=\tiny}
    ]

%-----------------------------------------------------------------
% 1) D_o (Black)
%-----------------------------------------------------------------
\addplot[
    only marks,
    mark=*,
    draw=black,
    fill=black,
    mark size=1.5pt,
    nodes near coords,
    every node near coord/.append style={font=\tiny,anchor=south, yshift=-1pt, xshift=-8pt},
    point meta=explicit symbolic, % Set meta data source
]
table [
    meta=Method,
    x=Parameters,
    y=Accuracy,
    col sep=space,
] {
    Method Accuracy Parameters
    [CLS] 69.87 77669
    GAP 70.87 77669
    SimPool 78.04 1258853
    %CLIP 83.47 2592869
    %SigLIP 85.05 7163237
    %AIM 84.46 1258085
    CBAM 72.46 151497
    %CoCa 84.16 1113701
    %CaiT 83.89 7169381
    %ViT 85.49 7163237
    %V-JEPA 85.51 7166309
    DELF 80.41 670566
    %AbMILP 78.53 78438
    %CAE 84.50 2444645
    %MHCA 84.74 1847909
    %EP$_2$ 81.52 669029
    EP$_4$ 82.79 670565
    EP$_8$ 84.44 673637
    %EP$_{64}$ 85.28 716645
};
\addlegendentry{$D_o = D_i$}

%-----------------------------------------------------------------
% 2) D_o / 4 (orange)
%-----------------------------------------------------------------
\addplot[
    only marks,
    mark=*,
    draw=orange,
    fill=orange,
    mark size=1.5pt,
    nodes near coords,
    every node near coord/.append style={font=\tiny,anchor=south, yshift=-1pt, xshift=-8pt},
    point meta=explicit symbolic,
]
table [
    meta=Method,
    x=Parameters,
    y=Accuracy,
    col sep=space,
] {
    Method Accuracy Parameters
    EP$_2$ 81.10 168485
    EP$_4$ 82.96 170021
    EP$_8$ 83.94 173093
    EP$_{64}$ 84.64 216101
};
\addlegendentry{$D_o = D_i / 4$}

%-----------------------------------------------------------------
% 3) D_o / 8 (purple)
%-----------------------------------------------------------------
%\addplot[
%    only marks,
%    mark=*,
%    draw=purple,
%    fill=purple,
%    mark size=2pt,
%    nodes near coords,
%    every node near coord/.append style={font=\tiny,anchor=south, yshift=1pt},
%    point meta=explicit symbolic,
%]
%table [
%    meta=Method,
%    x=Parameters,
%    y=Accuracy,
%    col sep=space,
%] {
%    Method Accuracy Parameters
%    EP$_2$ 81.10 85061
%    EP$_{16}$ 82.96 95813
%    EP$_{48}$ 83.50 120389
%};
%\addlegendentry{$D_o = \frac{D_i}{8}$}

%-----------------------------------------------------------------
% 4) D_o / 16 (brown)
%-----------------------------------------------------------------
\addplot[
    only marks,
    mark=*,
    draw=brown,
    fill=brown,
    mark size=1.5pt,
    nodes near coords,
    every node near coord/.append style={font=\tiny,anchor=south, yshift=-1pt, xshift=-8pt},
    point meta=explicit symbolic,
]
table [
    meta=Method,
    x=Parameters,
    y=Accuracy,
    col sep=space,
] {
    Method Accuracy Parameters
    EP$_2$ 78.06 43349
    EP$_4$ 79.20 44885
    EP$_{24}$ 80.27 54101
};
\addlegendentry{$D_o = D_i / 16$}

%-----------------------------------------------------------------
% Manual positions
%-----------------------------------------------------------------
\addplot[
    only marks,
    mark=*,
    draw=black,
    fill=black,
    mark size=1.5pt,
]
coordinates {
    (1258085, 84.46) % AIM
    %(1847909, 84.74) % MHCA
    (78438, 78.53) % AbMILP
    (716645, 85.28) % EP$_{64}$
    (7169381, 83.39) % CaiT
    (7163237, 85.05) % SigLIP
    (7163237, 85.49) % ViT
    (7166309, 85.51) % V-JEPA
    (2592869, 83.47) % CLIP
    (2444645, 84.50) % CAE
    (1113701, 84.16) % CoCa
};

\node[anchor=south, font=\tiny, yshift=2pt, xshift=7pt] at (1258085, 84.46) {AIM/MHCA};
%\node[anchor=south, font=\tiny, yshift=0pt, xshift=6pt] at (1847909, 84.74) {MHCA};
\node[anchor=south, font=\tiny, yshift=0pt, xshift=0pt] at (78438, 78.53) {AbMILP};
\node[anchor=south, font=\tiny, yshift=0pt, xshift=0pt] at (716645, 85.28) {EP$_{64}$};
\node[anchor=south, font=\tiny, yshift=-5pt, xshift=-8pt] at (7169381, 83.39) {CaiT};
\node[anchor=north, font=\tiny, yshift=0pt, xshift=0pt] at (7163237, 85.05) {SigLIP};
\node[anchor=south, font=\tiny, yshift=-1pt, xshift=6pt] at (7163237, 85.49) {ViT};
\node[anchor=south, font=\tiny, yshift=0pt, xshift=-8pt] at (7166309, 85.51) {V-JEPA};
\node[anchor=north, font=\tiny, yshift=0pt, xshift=0pt] at (2592869, 83.47) {CLIP};
\node[anchor=south, font=\tiny, yshift=-6pt, xshift=9pt] at (2444645, 84.50) {CAE};
\node[anchor=north, font=\tiny, yshift=1pt, xshift=6pt] at (1113701, 84.16) {CoCa};

%-----------------------------------------------------------------
% Pareto Front
%-----------------------------------------------------------------
\path [name path=pareto]
    (77669, 67.87)
    -- (78438, 78.53)
    -- (670566, 80.41)
    -- (1113701, 84.16)
    -- (1258085, 84.46)
    %-- (1847909, 84.74)
    -- (7166309, 85.51)
    -- (12000000, 85.51);

% Upper boundary (extend down to lowest y-value)
\path [name path=top]
    (12000000, 85.51) -- (12000000, 50);

% Fill the area to the right of the Pareto front
\addplot [fill=blue!30, opacity=0.3]
    fill between [of=pareto and top];

% --- Pareto front line ---
\addplot[thick, dashed, blue, mark=none] coordinates {
    (77669, 70.87)
    (78438, 78.53)
    (670566, 80.41)
    (1113701, 84.16)
    (1258085, 84.46)
    %(1847909, 84.74)
    (7166309, 85.51)
};

    \end{axis}
\end{tikzpicture}
%\caption{Scatter plot of Top-1 Accuracy (\%) vs. Number of Parameters on Food101 for various pooling methods. Each point is labeled with its corresponding method name positioned above the point.}
%\label{fig:accuracy_parameters_scatter}
%\end{figure}
        \label{fig:bench-food}
    \end{subfigure}
    %\vspace{-12pt}
    \begin{subfigure}[t]{0.48\textwidth}
        \centering
        \caption{BEiTv2 ViT-B ImageNet-1K}
        % Scatter Plot
%\begin{figure}[ht]
%\centering
\begin{tikzpicture}
    \begin{axis}[
        width=7.3cm, % Adjust width
        height=6cm,  % Adjust height
        xlabel={number of parameters},
        ylabel={top-1 accuracy (\%)},
        grid=both,
        enlargelimits=true,
        xmode=log,
%         log ticks with fixed point,
        legend pos=south east,
        legend style={font=\tiny}
    ]

%-----------------------------------------------------------------
% D_o (black)
%-----------------------------------------------------------------
\addplot[
    only marks,
    mark=*,
    draw=black,
    fill=black,
    mark size=1.5pt,
    nodes near coords,
    every node near coord/.append style={font=\tiny,anchor=south, yshift=-1pt, xshift=-8pt},
    point meta=explicit symbolic,
]
table [
    meta=Method,
    x=Parameters,
    y=Accuracy,
    col sep=space,
] {
    Method Accuracy Parameters
    [CLS] 78.95 769000
    GAP 78.55 769000
    %CLIP 81.48 3282664
    %SigLIP 80.59 7854568
    %SimPool 81.56 1950184
    %AIM 81.32 1949416
    %CBAM 78.53 842828
    CoCa 79.25 1851880
    %CaiT 80.76 7860712
    ViT 81.09 7854568
    V-JEPA 80.79 7857640
    DELF 81.46 1361897
    %AbMILP 81.11 769769
    %CAE 81.54 3135976
    %EP$_{2}^{\mathrm{D_o}}$ 81.38 1360360
    %EP$_{8}^{\mathrm{D_o}}$ 81.48 1360360
    %EP$_{16}$ 81.54 1371112
    EP$_{32}$ 81.66 1371112
    %EP$_{64}^{\mathrm{D_o}}$ 81.43 1407976
};
\addlegendentry{$D_o = D_i$}

%-----------------------------------------------------------------
% D_o / 2 (Blue)
%-----------------------------------------------------------------
\addplot[
    only marks,
    mark=*,
    draw=blue,
    fill=blue,
    mark size=1.5pt,
    nodes near coords,
    every node near coord/.append style={font=\tiny,anchor=south, yshift=-1pt, xshift=-8pt},
    point meta=explicit symbolic,
]
table [
    meta=Method,
    x=Parameters,
    y=Accuracy,
    col sep=space,
] {
    Method Accuracy Parameters
    EP$_{2}$ 80.90 681448
    EP$_{4}$ 81.13 682984
    EP$_{16}$ 81.27 692200
};
\addlegendentry{$D_o = D_i / 2$}

%-----------------------------------------------------------------
% D_o / 4 (Orange)
%-----------------------------------------------------------------
\addplot[
    only marks,
    mark=*,
    draw=orange,
    fill=orange,
    mark size=1.5pt,
    nodes near coords,
    every node near coord/.append style={font=\tiny,anchor=south, yshift=-1pt, xshift=-8pt},
    point meta=explicit symbolic,
]
table [
    meta=Method,
    x=Parameters,
    y=Accuracy,
    col sep=space,
] {
    Method Accuracy Parameters
    EP$_{2}$ 80.54 341992
    EP$_{4}$ 80.71 343528
    EP$_{8}$ 80.86 346600
    EP$_{32}$ 81.00 365032
};
\addlegendentry{$D_o = D_i / 4$}

%-----------------------------------------------------------------
% D_o / 8 (Red)
%-----------------------------------------------------------------
\addplot[
    only marks,
    mark=*,
    draw=purple,
    fill=purple,
    mark size=1.5pt,
    nodes near coords,
    every node near coord/.append style={font=\tiny,anchor=south, yshift=-1pt, xshift=-8pt},
    point meta=explicit symbolic,
]
table [
    meta=Method,
    x=Parameters,
    y=Accuracy,
    col sep=space,
] {
    Method Accuracy Parameters
    EP$_{2}$ 79.64 172264
    EP$_{16}$ 79.78 183016
};
\addlegendentry{$D_o = D_i / 8$}

%-----------------------------------------------------------------
% Manual positions
%-----------------------------------------------------------------
\addplot[
    only marks,
    mark=*,
    draw=black,
    fill=black,
    mark size=1.5pt,
]
coordinates {
    (842828, 78.53) % CBAM
    (769769, 81.11) % AbMILP
    (1950184, 81.56) % SimPool
    (3135976, 81.54) % CAE
    (3282664, 81.48) % CLIP
    (1949416, 81.32) % AIM
    %(2539240, 81.38) % MHCA
    (7854568, 80.59) % SigLIP
    (7860712, 80.76) % CaiT
};

\node[anchor=north, font=\tiny, yshift=1pt, xshift=8pt] at (842828, 78.53) {CBAM};
\node[anchor=north, font=\tiny, yshift=1pt, xshift=12pt] at (769769, 81.11) {AbMILP};
\node[anchor=south, font=\tiny, yshift=0pt, xshift=0pt] at (1950184, 81.56) {SimPool};
\node[anchor=south, font=\tiny, yshift=0pt, xshift=0pt] at (3135976, 81.54) {CAE};
\node[anchor=north, font=\tiny, yshift=1pt, xshift=7pt] at (3282664, 81.48) {CLIP};
\node[anchor=north, font=\tiny, yshift=0pt, xshift=-2pt] at (1949416, 81.32) {AIM/MHCA};
%\node[anchor=north, font=\tiny, yshift=0pt, xshift=4pt] at (2539240, 81.38) {MHCA};
\node[anchor=north, font=\tiny, yshift=5pt, xshift=-11pt] at (7854568, 80.59) {SigLIP};
\node[anchor=north, font=\tiny, yshift=5pt, xshift=-9pt] at (7860712, 80.76) {CaiT};

%-----------------------------------------------------------------
% Pareto Front + Transparent Region
%-----------------------------------------------------------------
\path [name path=pareto]
    (769000, 78.95)
    -- (769769, 81.11)
    -- (1361897, 81.46)
    -- (1950184, 81.56)
    -- (12000000, 81.56); % Extends right to show region

% Upper boundary for filling
\path [name path=top]
    (12000000, 100) -- (12000000, 50);

% Fill the area to the right of the Pareto front
\addplot [fill=blue!30, opacity=0.3]
    fill between [of=pareto and top];

% Pareto front line
\addplot[thick, dashed, blue, mark=none] coordinates {
    (769000, 78.95)
    (769769, 81.11)
    (1361897, 81.46)
    (1950184, 81.56)
};

    \end{axis}
\end{tikzpicture}
%\caption{Scatter plot of Top-1 Accuracy (\%) vs. Number of Parameters on ImageNet-1k for various pooling methods. Each point is labeled with its corresponding method name positioned above the point. The blue-shaded area represents the dominated region.}
%\label{fig:accuracy_parameters_scatter_beitv2}
%\end{figure}
        \label{fig:bench-beitv2}
    \end{subfigure}
    \begin{subfigure}[t]{0.48\textwidth}
        \centering
        \caption{CAPI ViT-L ImageNet-1K}
        % Scatter Plot
%\begin{figure}[ht]
%\centering
\begin{tikzpicture}
    \begin{axis}[
        width=7.3cm, % Adjust width
        height=6cm,  % Adjust height
        xlabel={number of parameters},
        grid=both,
        enlargelimits=true,
        xmode=log,
%         log ticks with fixed point,
        legend pos=south east,
        legend style={font=\tiny}
    ]

%-----------------------------------------------------------------
% D_o (black)
%-----------------------------------------------------------------

\addplot[
    only marks,
    mark=*,
    draw=black,
    fill=black,
    mark size=1.5pt,
    nodes near coords,
    every node near coord/.append style={font=\tiny,anchor=south, yshift=-1pt, xshift=-8pt},
    point meta=explicit symbolic,
]
table [
    meta=Method,
    x=Parameters,
    y=Accuracy,
    col sep=space,
] {
    Method Accuracy Parameters
    [CLS] 81.5 1025000
    GAP 76.3 1025000
    %CBAM 73.21 1156172
    % CLIP 82.7 5427176
    % SigLIP 83.0 13618152
    %SimPool 78.09 3124200
    % AIM 83.1 3123176
    % CoCa 82.5 2406376
    % CaiT 82.7 13626344
    % ViT 83.2 13618152
    % V-JEPA 82.8 13622248
    % DELF 82.6 2077673
    AbMILP 82.4 1026025
    % CAE 82.7 5229544
    % \textcolor{pink}{CAPI} 82.9 4219880
    %EP$_2$ 82.78 2075624
    %EP$_4$ 83.09 2077672
    %EP$_8$ 83.20 2081768
    EP$_{16}$ 83.39 2089960
    EP$_{64}$ 83.69 2139112
    %EP$_{128}$ 83.50 2204648
};
\addlegendentry{$D_o = D_i$}
%-----------------------------------------------------------------
% D_o / 2 (Blue)
%-----------------------------------------------------------------
\addplot[
    only marks,
    mark=*,
    draw=blue,
    fill=blue,
    mark size=1.5pt,
    nodes near coords,
    every node near coord/.append style={font=\tiny,anchor=south, yshift=-1pt, xshift=-8pt},
    point meta=explicit symbolic,
]
table [
    meta=Method,
    x=Parameters,
    y=Accuracy,
    col sep=space,
] {
    Method Accuracy Parameters
    %EP$_{2}$ 80.90 1039336
    %EP$_{4}$ 81.13 1041384
    EP$_{8}$ 83.13 1045480
    %EP$_{16}$ 83.37 1053672
    EP$_{64}$ 83.52 1151976
};
\addlegendentry{$D_o = D_i / 2$}

%-----------------------------------------------------------------
% D_o / 4 (Orange)
%-----------------------------------------------------------------
\addplot[
    only marks,
    mark=*,
    draw=orange,
    fill=orange,
    mark size=1.5pt,
    nodes near coords,
    every node near coord/.append style={font=\tiny,anchor=south, yshift=-1pt, xshift=-8pt},
    point meta=explicit symbolic,
]
table [
    meta=Method,
    x=Parameters,
    y=Accuracy,
    col sep=space,
] {
    Method Accuracy Parameters
    %EP$_{2}$ 80.54 341992
    %EP$_{4}$ 80.71 343528
    EP$_{8}$ 82.76 527336
    %EP$_{16}$ 83.0 535528
    EP$_{64}$ 83.13 584680
};
\addlegendentry{$D_o = D_i / 4$}

%-----------------------------------------------------------------
% D_o / 8 (Red)
%-----------------------------------------------------------------
\addplot[
    only marks,
    mark=*,
    draw=purple,
    fill=purple,
    mark size=1.5pt,
    nodes near coords,
    every node near coord/.append style={font=\tiny,anchor=south, yshift=-1pt, xshift=-8pt},
    point meta=explicit symbolic,
]
table [
    meta=Method,
    x=Parameters,
    y=Accuracy,
    col sep=space,
] {
    Method Accuracy Parameters
    EP$_{8}$ 82.68 268264
    %EP$_{16}$ 82.80 276456
    EP$_{64}$ 82.98 281064
};
\addlegendentry{$D_o = D_i / 8$}

\addplot[
    only marks,
    mark=*,
    draw=black,
    fill=black,
    mark size=1.5pt,
    nodes near coords,
    every node near coord/.append style={font=\tiny,anchor=north west, yshift=1pt, xshift=-4pt},
    point meta=explicit symbolic,
]
table [
    meta=Method,
    x=Parameters,
    y=Accuracy,
    col sep=space,
] {
    Method Accuracy Parameters
    CoCa 82.5 2406376
    CLIP 82.7 5427176
};

\addplot[
    only marks,
    mark=*,
    draw=black,
    fill=black,
    mark size=1.5pt,
    nodes near coords,
    every node near coord/.append style={font=\tiny,anchor=south west, yshift=0pt, xshift=-3pt},
    point meta=explicit symbolic,
]
table [
    meta=Method,
    x=Parameters,
    y=Accuracy,
    col sep=space,
] {
    Method Accuracy Parameters
    CAPI 82.9 4219880
};

\addplot[
    only marks,
    mark=*,
    draw=black,
    fill=black,
    mark size=1.5pt,
    nodes near coords,
    every node near coord/.append style={font=\tiny,anchor=south west, yshift=1.5pt, xshift=-12pt},
    point meta=explicit symbolic,
]
table [
    meta=Method,
    x=Parameters,
    y=Accuracy,
    col sep=space,
] {
    Method Accuracy Parameters
    AIM/MHCA 83.1 3123176
};

\addplot[
    only marks,
    mark=*,
    draw=black,
    fill=black,
    mark size=1.5pt,
    nodes near coords,
    every node near coord/.append style={font=\tiny,anchor=south west, yshift=-5pt, xshift=-2pt},
    point meta=explicit symbolic,
]
table [
    meta=Method,
    x=Parameters,
    y=Accuracy,
    col sep=space,
] {
    Method Accuracy Parameters
    ViT 83.2 13618152
    V-JEPA 82.8 13622248
};

\addplot[
    only marks,
    mark=*,
    draw=black,
    fill=black,
    mark size=1.5pt,
    nodes near coords,
    every node near coord/.append style={font=\tiny,anchor=south east, yshift=-6pt, xshift=1pt},
    point meta=explicit symbolic,
]
table [
    meta=Method,
    x=Parameters,
    y=Accuracy,
    col sep=space,
] {
    Method Accuracy Parameters
    SigLIP 83.0 13618152
    CaiT 82.7 13626344
};

\addplot[
    only marks,
    mark=*,
    draw=black,
    fill=black,
    mark size=1.5pt,
    nodes near coords,
    every node near coord/.append style={font=\tiny,anchor=north east, yshift=2pt, xshift=1.5pt},
    point meta=explicit symbolic,
]
table [
    meta=Method,
    x=Parameters,
    y=Accuracy,
    col sep=space,
] {
    Method Accuracy Parameters
    DELF 82.6 2077673
    CAE 82.7 5229544
};
%-----------------------------------------------------------------
% Pareto Front + Transparent Region
%-----------------------------------------------------------------
\path [name path=pareto]
    (1025000, 75.0)
    (1025000, 76.3)
    -- (1025000, 81.5)
    -- (1026025, 82.4)
    -- (3123176, 83.1)
    -- (13618152, 83.2); % Extends right to show region

% Upper boundary for filling
\path [name path=top]
    (102500000, 83.2) -- (102500000, 50);

% Fill the area to the right of the Pareto front
\addplot [fill=blue!30, opacity=0.3]
    fill between [of=pareto and top];

% Pareto front line
\addplot[thick, dashed, blue, mark=none] coordinates {
    (1025000, 76.3)
    (1025000, 81.5)
    (1026025, 82.4)
    (3123176, 83.1)
    (13618152, 83.2)
};

    \end{axis}
\end{tikzpicture}
%\caption{Accuracy (\%) vs. \# of parameters on IN-1k with CAPI ViT-L for various pooling methods.}
%\label{fig:accuracy_parameters_scatter_capi}
%\end{figure}
        \label{fig:bench-capi}
    \end{subfigure}
    \vspace{-14pt}
    \caption{\emph{Top-1 classification accuracy \vs number of parameters} for various self-supervised pre-training methods across different datasets. We evaluate both dedicated probing mechanisms (\eg, V-JEPA) and repurposed attentive pooling methods (\eg, CLIP). \OUR variants are marked with different colors for different output dimensionalities $D_o$. $\mOUR_M$: \our with $M$ learnable queries. \cls: linear probing using the classification token; \gap: global average pooling over patch tokens; \textsc{mhca}: multi-head cross-attention; ViT: default transformer block.}
    \label{fig:bench_datasets}
    \vspace{-6pt}
\end{figure*}
%\OUR consistently achieves better performance \emph{with up to 4$\times$ fewer parameters} than linear probing.
%------------------------------------------------------------------------------

\subsection{Experimental Results}
\label{sec:results}

\paragraph{Accuracy \vs parameters.}

In~\autoref{fig:bench_datasets}, we compare \our (\OUR) with baseline and competitor methods using MAE ViT-B, BEiTv2 ViT-B, and CAPI ViT-L on ImageNet-1K, and MAE ViT-B on Food-101. We plot top-1 accuracy against the number of trainable parameters, including both attentive pooling and classifier parameters, and overlay the Pareto frontier to highlight optimal trade-offs. The two primary baselines, \cls and \gap, are the most parameter-efficient, as they introduce no overhead beyond the classifier parameters, but yield noticeably lower accuracy. In contrast, methods like \vjepa, \cait, \siglip, and the reference ViT block employ significantly more parameters, though within the attentive probing setting, their increased complexity provides mostly marginal accuracy improvements. Among the existing attentive probing or pooling methods, \simpool provides moderate accuracy but is not particularly parameter-efficient, while \cae and \clip achieve stronger performance at the cost of higher parameter counts. \abmilp, \delf, \aim, and MHCA lie primarily on the Pareto frontier, striking the optimal balance.

\OUR consistently achieves the best accuracy–parameter trade-off, positioning itself on the left or upper-left region of the Pareto frontier across pre-training methods and datasets. A key factor is its flexibility in controlling the number of queries $M$ and the output dimensionality $D_o$ (because of $W_V$), allowing adaptation to different parameter constraints. Notably, on ImageNet-1K with MAE ViT-B, $\mOUR_{64}$ (64 queries) achieves a state-of-the-art top-1 accuracy of 75.6\% with less than 1.4M parameters. $\mOUR_{48}$ with 48 queries and $D_o = D_i / 8$ achieves 70.3\% top-1 accuracy, while having a little more than 200k parameters, \ie almost 4$\times$ less than linear probing (\cls). On ImageNet-1K with BEiTv2 ViT-B and with CAPI ViT-L, $\mOUR_{32}$ and $\mOUR_{64}$ achieve a state-of-the-art accuracy of 81.7\% and 83.7\% respectively. On Food-101, a dataset smaller than ImageNet-1K, we observe that reducing $D_o$ even to $D_i / 16$ does not significantly hurt performance. More results in~\autoref{subsec:more_exp_results}.

\begin{wrapfigure}{r}{0.48\textwidth}
    \vspace{-4pt}
    \centering
    \begin{tikzpicture}
    \begin{axis}[
        width=7cm,
        height=5.5cm,
        xlabel={GFLOPs (beyond backbone)},
        ylabel={top-1 accuracy (\%)},
        grid=major,
        enlargelimits=true,
        legend pos=south east,
        legend style={font=\tiny}
    ]

% 1) D_o (black)
\addplot[
    only marks,
    mark=*,
    draw=black,
    fill=black,
    mark size=1.5pt,
    nodes near coords,
    every node near coord/.append style={font=\tiny,anchor=south, yshift=-1pt, xshift=-8pt},
    point meta=explicit symbolic,
]
table [
    meta=Method,
    x=Flops,
    y=Accuracy,
    col sep=space,
] {
    Method Accuracy Flops
    \textsc{[CLS]} 67.66 0
    GAP 66.72 0
    CoCa 73.53 0.213500896
    ViT 75.25 1.447778304
    EP$_{64}$ 75.58 0.125389824
};
\addlegendentry{$D_o = D_i$}

% 2) D_o = D_i / 2 (Blue)
\addplot[
    only marks,
    mark=*,
    draw=blue,
    fill=blue,
    mark size=1.5pt,
    nodes near coords,
    every node near coord/.append style={font=\tiny,anchor=south, yshift=-1pt, xshift=-8pt},
    point meta=explicit symbolic,
]
table [
    meta=Method,
    x=Flops,
    y=Accuracy,
    col sep=space,
] {
    Method Accuracy Flops
    EP$_{96}$ 74.86 0.07194432
};
\addlegendentry{$D_o = D_i / 2$}

% 3) D_o = D_i / 4 (Orange)
\addplot[
    only marks,
    mark=*,
    draw=orange,
    fill=orange,
    mark size=1.5pt,
    nodes near coords,
    every node near coord/.append style={font=\tiny,anchor=south, yshift=-1pt, xshift=-8pt},
    point meta=explicit symbolic,
]
table [
    meta=Method,
    x=Flops,
    y=Accuracy,
    col sep=space,
] {
    Method Accuracy Flops
    EP$_{64}$ 72.88 0.037996224
};
\addlegendentry{$D_o = D_i / 4$}

% Manual positions (other methods)
\addplot[
    only marks,
    mark=*,
    draw=black,
    fill=black,
    mark size=1.5pt,
]
coordinates {
    (0.525144576, 74.84) % CLIP
    (0.231662592, 75.18) % AIM
    (0.237865728, 74.12) % V-JEPA
    (0.117001664, 72.64) % SimPool
    (0.000150528, 71.74) % AbMILP
};

\node[anchor=north, font=\tiny, yshift=-1pt] at (0.525144576, 74.84) {CLIP};
\node[anchor=south, font=\tiny, yshift=1pt, xshift=10pt] at (0.231662592, 75.18) {AIM/MHCA};
\node[anchor=north, font=\tiny, yshift=1pt, xshift=10pt] at (0.237865728, 74.12) {V-JEPA};
\node[anchor=north, font=\tiny, yshift=1pt, xshift=10pt] at (0.117001664, 72.64) {SimPool};
\node[anchor=north, font=\tiny, yshift=1pt, xshift=10pt] at (0.000150528, 71.74) {AbMILP};

% Pareto region and line
\path [name path=pareto]
    (0, 64)
    -- (0.000150528, 71.74)
    -- (0.117001664, 72.64)
    -- (0.213500896, 73.53)
    -- (0.231662592, 75.18)
    -- (1.6, 75.25);

\path [name path=top]
    (1.6, 100) -- (1.6, 50);

\addplot [fill=blue!30, opacity=0.3]
    fill between [of=pareto and top];

\addplot[thick, dashed, blue, mark=none] coordinates {
    (0, 66.72)
    (0.000150528, 71.74)
    (0.117001664, 72.64)
    (0.213500896, 73.53)
    (0.231662592, 75.18)
    (1.447778304, 75.25)
};

    \end{axis}
\end{tikzpicture}
    \vspace{-18pt}
    \caption{\emph{Top-1 classification accuracy \vs GFLOPs} for MAE ViT-B with different probings on ImageNet-1K.}
    \label{fig:flops}
\end{wrapfigure}

\paragraph{Accuracy \vs computational cost.}

In~\autoref{fig:flops}, we compare different pooling/probing methods in terms of top-1 accuracy against computational cost, measured in GFLOPs. Baselines \cls and \gap are the most efficient but less accurate, while self-attention methods (\eg, ViT, CLIP) incur higher cost due to additional attention computations. \OUR achieves better accuracy than a ViT block with over 10$\times$ less compute, and consistently lies on the Pareto frontier, combining high accuracy with low cost. By adjusting the output dimensionality $D_o$, \OUR scales to different computational budgets without significant accuracy loss. Note that backbone FLOPs (\eg, 17.58 GFLOPs for MAE ViT-B) are not shown, since they are identical across all methods and thus omitted for fairness. Here we only compare the additional cost of the pooling/probing methods. 

%An alternative evaluation protocol using frozen and pre-stored features, requiring a single backbone forward pass, remains an interesting direction for future work.

%For clarity, we mostly retain only those defining the Pareto frontier.

%------------------------------------------------------------------------------
%\begin{figure}[ht]
%    \centering
%    \input{plots/mae_vitb_imagenet1k_flops_colors}
%    \vspace{-4pt}
%    \caption{\emph{Top-1 classification accuracy \vs GFLOPS} for MAE ViT-B with different probings on ImageNet-1k.}
%    \label{fig:flops}
%    \vspace{-10pt}
%\end{figure}
%------------------------------------------------------------------------------

\textbf{Comparison of pre-training methods.} \autoref{tab:extended_comparisons} compares methods from diverse pre-training paradigms under multiple evaluation protocols on ImageNet-1K. Fine-tuning (FT) achieves the highest accuracy but is compute-intensive and unsustainable at scale, making it increasingly rare in recent evaluations (\textcolor{red}{$\times$}). $k$-NN consistently underperforms, reflecting the weak separability of raw features. Linear probing (LP) offers a stronger baseline, yet \OUR provides consistent gains across all paradigms, often with only a small increase in parameters. Notably, \OUR changes the relative ranking of methods compared to LP and $k$-NN: for example, MAE surpasses BYOL, and CAPI outperforms CLIP, challenging the view that MIM methods are weaker. In general, methods whose pre-training optimizes representations of patch tokens rather than an explicit global representation are benefit the most from attentive probing (\eg, SimMIM +13.6\%, DiT +24.3\%). We further analyze this trend in~\autoref{subsec:more_exp_analysis}. Across the board, \OUR narrows the gap to full fine-tuning while remaining lightweight and scalable, demonstrating both its generality and effectiveness.

\begin{table}[t]
\renewcommand{\arraystretch}{.8} %
\vspace{-8pt}
    \centering
    \scriptsize
        \caption{\emph{Comparison of pre-training methods in terms of different evaluation protocols} on ImageNet-1K. \textsc{\# Par.}: number of trainable parameters including classifier; $^{\dagger}$Results reported from the official paper under different than probing setup (\eg, augmentations); $^{\ddagger}$Evaluated using \gap.}
    \vspace{-6pt}
    \setlength{\tabcolsep}{2pt}
        \begin{tabular}{c|lcc|c|cc|cc|c|cc}
        \toprule
         & \textsc{Method} &  \textsc{Arch} & \textsc{Pre-training} & k-NN  & LP & \textsc{\# Par.} & EP & \textsc{\# Par.} & \textsc{Gain} & FT$^{\dagger}$ & \textsc{\# Par.} \\
        \midrule
        \multirow{6}{*}{\rotatebox{90}{MIM}} 
        & \multirow{3}{*}{MAE~\citep{mae}}      & ViT-S/16 & \multirow{3}{*}{IN-1K} & 26.7 & 47.4 & 0.4M & 64.6 & 0.5M & \textcolor{ForestGreen}{+17.2} & 80.6 & 22M \\
        & &  ViT-B/16 & & 46.1 & 67.7 & 0.8M & 75.6 & 1.4M & \textcolor{ForestGreen}{+7.9}  & 83.6 & 87M \\
        & &  ViT-L/16 &  & 58.2 & 76.0 & 1.0M & 79.3 & 2.1M & \textcolor{ForestGreen}{+3.3} & 85.9 & 304M \\
        \cline{2-12} \\ [-4pt]
        & BEiTv2~\citep{beitv2}   & ViT-B/16 & IN‑1K & 74.8 & 79.0 & 0.8M & 81.7 & 1.4M & \textcolor{ForestGreen}{+2.7}  & 85.0 & 87M \\
        & SimMIM~\citep{simMIM}   & ViT-B/16 & IN‑1K & 15.1 & 51.5 & 0.8M & 65.1 & 1.4M & \textcolor{ForestGreen}{+13.6} & 83.8 &  87M \\
        & CAPI~\citep{capi}     & ViT-L/14 & IN‑1K & 76.7  & 81.5  & 1.0M & 83.6 & 2.1M & \textcolor{ForestGreen}{+2.1} & \textcolor{red}{$\times$} & 304M \\
        \midrule
        \multirow{2}{*}{\rotatebox{90}{JEA}} & BYOL~\citep{byol}     & RN-50 & IN‑1K & 64.8  & 74.3 & 2.0M & 75.1 & 6.3M & \textcolor{ForestGreen}{+0.8} & 77.7 & 26M \\
        & DINO~\citep{caron2021emerging}     & ViT-B/16 & IN‑1K & 76.1  & 77.3 & 0.8M & 77.8 & 1.4M & \textcolor{ForestGreen}{+0.5} & 82.8 & 87M\\
        \midrule
        \multirow{7}{*}{\rotatebox{90}{HYBRID}} & iBOT~\citep{ibot}     & ViT-B/16 & IN‑1K & 77.0 & 78.7 & 0.8M & 79.2  & 1.4M & \textcolor{ForestGreen}{+0.5} & 84.0 & 87M \\
        \cline{2-12} \\ [-4pt]
        & \multirow{2}{*}{DINOv2~\citep{dinov2}}   & ViT-B/14 & \multirow{2}{*}{LVD-142M} & 81.8  & 83.2 & 0.8M & 84.0 & 1.4M &\textcolor{ForestGreen}{+0.8} & \textcolor{red}{$\times$} & 87M \\
        &  & ViT-L/14 & & 83.5 & 85.2 & 1.0M & 85.6 & 2.1M & \textcolor{ForestGreen}{+0.4} & \textcolor{red}{$\times$} & 304M \\
        \cline{2-12} & \\ [-4pt]
        & Franca~\citep{franca}  &  ViT-L/14 & IN-21k & 82.2 & 83.8 & 1.0M & 84.3 & 2.1M & \textcolor{ForestGreen}{+0.5} & \textcolor{red}{$\times$} & 304M \\
        \cline{2-12} \\ [-4pt]
        & \multirow{2}{*}{DINOv3~\citep{dinov3}}   & ViT-B/16 & \multirow{2}{*}{LVD-1689M} & 83.0 & 84.0 & 0.8M & 84.4 & 1.4M &\textcolor{ForestGreen}{+0.4} & \textcolor{red}{$\times$} & 87M \\
        & & ViT-L/16 & & 85.3 & 86.6 & 1.0M & 87.1 & 2.1M & \textcolor{ForestGreen}{+0.5} & \textcolor{red}{$\times$} & 304M \\
        \midrule
        \multirow{3}{*}{\rotatebox{90}{VLM}} & CLIP~\citep{radford2021clip}     & ViT-L/14 & WIT & 77.2  & 82.3 & 0.8M & 83.4  & 2.1M & \textcolor{ForestGreen}{+1.1} & \textcolor{red}{$\times$} & 305M \\
        & SigLIP~\citep{siglip} & ViT-L/16 & WebLI & 83.7 & 84.1$^{\ddagger}$ & 1.0M & 86.1 & 2.1M & \textcolor{ForestGreen}{+2.0} & \textcolor{red}{$\times$} & 305M \\
        & SigLIP2~\citep{siglip2}  & ViT-L/16 & WebLI & 84.4 & 85.2$^{\ddagger}$ & 1.0M & 87.0 & 2.1M & \textcolor{ForestGreen}{+1.8} & \textcolor{red}{$\times$} & 305M \\
        \midrule
        \multirow{2}{*}{\rotatebox{90}{GEN}} & DiT~\citep{dit}    & DiT-XL/2 & IN‑1K & 8.3 & 32.7$^{\ddagger}$ & 1.2M & 57.0 & 2.5M & \textcolor{ForestGreen}{+24.3} & \textcolor{red}{$\times$} & 676M \\
        %& SiT~\citep{sit}    & DiT-XL/2 & IN‑1K & -- & -- & 1.2M & -- & 2.5M &  -- & & -- & 676M \\
        & AIMv2~\citep{aimv2} & ViT-L/14 & custom$^{*}$ & 80.8 & 84.8$^{\ddagger}$ & 1.0M & 85.9 & 2.1M & \textcolor{ForestGreen}{+1.1} & \textcolor{red}{$\times$} & 304M \\
        \bottomrule
    \end{tabular}
    \begin{minipage}{0.9\linewidth}
    \scriptsize \emph{Note.} custom$^{*}$: DFN-2B~\citep{dfn}, COYO~\citep{coyo}, HQITP~\citep{aimv2}. Default EP $=$ EP$_{32}$.
    \end{minipage}
    \label{tab:extended_comparisons}
    \vspace{-6pt}
\end{table}

\begin{figure*}[t]
    \centering
    \begin{tikzpicture}
    \begin{axis}[
        width=11cm,
        height=6.1cm,
        xlabel={number of parameters},
        ylabel={top-1 accuracy (\%)},
        grid=major,
        enlargelimits=true,
        xmode=log,
        xtick={250000,500000,1000000,2000000},
        xticklabels={$2.5\cdot10^5$, $5\cdot10^5$, $10^6$, $2\cdot10^6$},
          legend style={
            font=\scriptsize,
            at={(1.37,0.00)},   % (x,y) position
            anchor=south east,  % which corner of the legend sits at that point
          },
    ]

% HELPER
\pgfplotsset{
  starx legend/.style={
    legend image code/.code={
      \draw[mark=*, mark size=1.5pt, draw=orange, fill=orange]
        plot coordinates {(0.3cm,0cm)};
      \draw[mark=x, mark size=3.5pt, draw=orange]
        plot coordinates {(0.3cm,0cm)};
    },
  },
}

%------------------------------------------------------------
% BASELINE POOLINGS (BLUE POINTS)
%------------------------------------------------------------
\addplot[
    only marks,
    mark=*,
    draw=blue,
    fill=blue,
    mark size=1.5pt,
    %forget plot,
]
coordinates {
    (769000, 67.66)   % [CLS]
    (769769, 71.74)   % AbMILP
    (1361897, 72.86)  % DELF
    (1949416, 75.18)  % AIM
};

% Labels for baseline poolings
\node[anchor=north, font=\tiny] at (axis cs:769000, 67.66) {[CLS]};
\node[anchor=north, font=\tiny] at (axis cs:769769, 71.74) {AbMILP};
\node[anchor=north, font=\tiny] at (axis cs:1361897, 72.86) {DELF};
\node[anchor=south, font=\tiny] at (axis cs:1949416, 75.18) {AIM};

\addlegendentry{Baseline probings}

%------------------------------------------------------------
% OLD PARETO CURVE (BLUE DASHED)
%------------------------------------------------------------
\addplot[
    thick,
    dashed,
    blue,
    mark=none
]
coordinates {
    (769000, 67.66)
    (769769, 71.74)
    (1361897, 72.86)
    (1949416, 75.18)
};
\addlegendentry{Baseline Pareto}

%------------------------------------------------------------
% EP PARETO POINTS (ORANGE POINTS)
%------------------------------------------------------------
\addplot[
    only marks,
    mark=*,
    draw=orange,
    fill=orange,
    mark size=1.5pt,
    %forget plot,
]
coordinates {
    (207592, 70.27)   % EP eighth
    (389608, 72.88)   % EP quarter
    (753640, 74.86)   % EP half
    (1395688, 75.58)  % EP full
};

% Labels for EP points
\node[anchor=north, font=\tiny] at (axis cs:207592, 70.27) {EP$^{D_i/8}$};
\node[anchor=north, font=\tiny] at (axis cs:389608, 72.88) {EP$^{D_i/4}$};
\node[anchor=south, font=\tiny] at (axis cs:753640, 74.86) {EP$^{D_i/2}$};
\node[anchor=south, font=\tiny] at (axis cs:1395688, 75.58) {EP$^{D_i}$};

\addlegendentry{EP at various $D_o$}

%------------------------------------------------------------
% NEW EP PARETO CURVE (ORANGE DASHED)
%------------------------------------------------------------
\addplot[
    thick,
    dashed,
    orange,
    mark=none
]
coordinates {
    (207592, 70.27)
    (389608, 72.88)
    (753640, 74.86)
    (1395688, 75.58)
};
\addlegendentry{EP Pareto}

%------------------------------------------------------------
% LoRA Layer 4 (RED)
%------------------------------------------------------------
\addplot[
    only marks,
    mark=x,
    draw=red,
    fill=red,
    mark size=2.0pt,
]
coordinates {
    (781288, 69.11)
    (781288, 69.09)
    (781288, 70.99)
    (793576, 69.23)
    (805864, 71.48)
    (818152, 71.94)
    (867304, 71.89)
    (965608, 72.31)
    (1162216, 72.27)
};
\addlegendentry{LoRA at layer 4}

%------------------------------------------------------------
% LoRA Layer 8 (CYAN)
%------------------------------------------------------------
\addplot[
    only marks,
    mark=x,
    draw=cyan,
    fill=cyan,
    mark size=2.0pt,
]
coordinates {
    (781288, 70.00)
    (781288, 69.70)
    (781288, 71.89)
    (793576, 70.25)
    (805864, 72.83)
    (818152, 72.84)
    (867304, 73.31)
    (965608, 73.52)
    (1162216, 73.73)
};
\addlegendentry{LoRA at layer 8}

%------------------------------------------------------------
% LoRA Layer 12 (MAGENTA)
%------------------------------------------------------------
\addplot[
    only marks,
    mark=x,
    draw=magenta,
    fill=magenta,
    mark size=2.0pt,
]
coordinates {
    (781288, 71.21)
    (781288, 71.33)
    (781288, 68.67)
    (793576, 71.79)
    (805864, 72.71)
    (818152, 72.75)
    (867304, 73.65)
    (965608, 74.30)
    (1162216, 74.90)
};
\addlegendentry{LoRA at layer 12}

%------------------------------------------------------------
% LoRA: All Layers (GREEN)
%------------------------------------------------------------
\addplot[
    only marks,
    mark=x,
    draw=orange,
    fill=orange,
    mark size=2.0pt,
]
coordinates {
    (916456, 74.29)   % LoRA_Qall
    (916456, 74.49)   % LoRA_Kall
    (916456, 74.89)   % LoRA_Vall
    (1063912, 74.96)  % LoRA_QKall
    (1211368, 76.01)  % LoRA_QKV_all
    (1358824, 76.36)  % LoRA_QKVO_all
    (1063912, 74.62)  % LoRA_Qall_rho16
    (1358824, 75.05)  % LoRA_QKall_rho16
    (1653736, 76.46)  % LoRA_QKVall_rho16
    (1948648, 76.72)  % LoRA_QKVOall_rho16
    % (3128296, 77.06)  % LoRA_QKVOall_rho32
    % (5487592, 77.18)  % LoRA_QKVOall_rho64
};
\addlegendentry{LoRA on all layers}

%------------------------------------------------------------
% BitFit (ORANGE-BROWN)
%------------------------------------------------------------
\addplot[
    only marks,
    mark=x,
    draw=orange!80!black,
    fill=orange!80!black,
    mark size=2.0pt,
]
coordinates {
    (975824, 74.71)   % BitFit
};
\addlegendentry{BitFit}

%------------------------------------------------------------
% Tuning Layernorm (green-black)
%------------------------------------------------------------
\addplot[
    only marks,
    mark=x,
    draw=green!60!black,
    fill=green!60!black,
    mark size=2.0pt,
]
coordinates {
    (845800, 72.81)   % BitFit
};
\addlegendentry{LayerNorm tuning}

%------------------------------------------------------------
% BitFit + EP Hybrid (BROWN)
%------------------------------------------------------------
%\addplot[
%    only marks,
%    mark=*,
%    draw=brown,
%    fill=brown,
%    mark size=1.5pt,
%]
%coordinates {
%    (1590224, 77.28)   % BitFit + EP_full_32
%    (911312,  77.02)   % BitFit + EP_d2_32
%    (571856,  75.41)   % BitFit + EP_d4_32
%    (402128,  73.54)   % BitFit + EP_d8_32
%};
%\addlegendentry{BitFit+EP}

%------------------------------------------------------------
% Hybrid LoRA + EP (PURPLE)
%------------------------------------------------------------
\addplot[
    only marks,
    mark=*,
    draw=orange,
    fill=orange,
    mark size=1.5pt,
    forget plot,
]
coordinates {
    (1530856, 77.23)   % LoRA_Vall + EP_full_32
    (1825768, 77.52)   % LoRA_QKVall + EP_full_32
    % (1973224, 77.51)   % LoRA_QKVOall + EP_full_32
    
    (851944,  76.99)   % LoRA_Vall + EP_half_32
    (512488,  75.84)   % LoRA_Vall + EP_quarter_32
    (342760,  74.17)   % LoRA_Vall + EP_d8_32
    (251752,  71.99)   % LoRA_Vall + EP_d16_24
};

%\addlegendimage{starx legend}
%\addlegendentry{LoRA $+$ EP}

%------------------------------------------------------------
% Hybrid LoRA + EP (PURPLE)
%------------------------------------------------------------
\addplot[
    only marks,
    mark=x,
    draw=orange,
    fill=orange,
    mark size=3.5pt,
    forget plot,
]
coordinates {
    (1530856, 77.23)   % LoRA_Vall + EP_full_32
    (1825768, 77.52)   % LoRA_QKVall + EP_full_32
    % (1973224, 77.51)   % LoRA_QKVOall + EP_full_32
    
    (851944,  76.99)   % LoRA_Vall + EP_half_32
    (512488,  75.84)   % LoRA_Vall + EP_quarter_32
    (342760,  74.17)   % LoRA_Vall + EP_d8_32
    (251752,  71.99)   % LoRA_Vall + EP_d16_24
};

\addlegendentry{LoRA $+$ EP}
\addlegendimage{starx legend}

%------------------------------------------------------------
% BitFit + EP Pareto (BROWN DASHED, NO LEGEND)
%------------------------------------------------------------
%\addplot[
%    thick,
%    dashed,
%    brown,
%    mark=none,
%]
%coordinates {
%    (402128, 73.54)
%    (571856, 75.41)
%    (911312, 77.02)
%    (1590224, 77.28)
%};

%------------------------------------------------------------
% Hybrid LoRA+EP Pareto curve (PURPLE DASHED, NO LEGEND)
%------------------------------------------------------------
\addplot[
    thick,
    solid,
    orange,
    mark=none
]
coordinates {
    (251752, 71.99)
    (342760, 74.17)
    (512488, 75.84)
    (851944, 76.99)
    (1530856, 77.23)
    (1825768, 77.52)
    % (1973224, 77.51)
};

\addlegendentry{LoRA $+$ EP Pareto}

\path [name path=pareto]
    (769000, 64)
    -- (769769, 71.74)
    -- (1361897, 72.86)
    -- (1949416, 75.18)
    -- (2549416, 75.18);

% Upper boundary (extend down to lowest y-value)
\path [name path=top]
    (2549416, 100) -- (2549416, 50);

% Fill the area to the right of the Pareto front
\addplot [fill=blue!30, opacity=0.3]
    fill between [of=pareto and top];

    \end{axis}
\end{tikzpicture}
    \vspace{-18pt}
    \caption{\emph{Accuracy–parameter trade-off of probing and parameter-efficient fine-tuning methods} on ImageNet-1K using a frozen MAE ViT-B/16 backbone. The \emph{hybrid} LoRA $+$ EP configurations achieve a \emph{strictly better trade-off} than both pure EP and pure LoRA.}
    \vspace{-18pt}
    \label{fig:lora_merged}
\end{figure*}

\textbf{Comparison against PEFT methods.} In~\autoref{fig:lora_merged} we compare the most efficient baseline probing methods (\cls, AbMILP, DELF, AIM), EP at different output dimensionalities $D_o$, more than 40 LoRA variants, BitFit, and LayerNorm tuning, all on a frozen MAE ViT-B/16. We sweep LoRA across individual transformer layers (4, 8, 12) as well as all layers, and across multiple configurations (\eg $W_Q$-only, $W_K$-only, $W_V$-only, $W_Q{+}W_K$, $W_Q{+}W_K{+}W_V$) with ranks $\rho \in \{8,16,32,64\}$.

Overall, LoRA applied to a small subset of layers (red, cyan, magenta crosses) lies roughly on or slightly above the baseline Pareto front, but is consistently dominated by EP in the accuracy–parameter plane. For instance, EP$^{D_i/2}$ attains $74.9\%$ top-1 accuracy with only $750$K parameters, whereas the best single-layer LoRA configuration around that region requires around $1.2$M parameters. LayerNorm tuning and BitFit (green and brown crosses) also outperform the pure probing baselines, yet EP remains strictly more efficient. All-layer LoRA configurations (orange crosses) move closer to the EP Pareto curve and some even surpass it in accuracy (up to $76.7\%$). In~\autoref{subsec:more_exp_results}, we present an analysis with in- and out-of-domain $k$-NN experiments on frozen features, showing that all-layer LoRA modifies the representation more strongly than EP, consistent with their roles as task-adaptive low-rank fine-tuning (LoRA) versus representation-preserving probing (EP).

Motivated by these observations, we finally combine one of the most parameter-efficient LoRA variants (LoRA on all $W_V$ matrices across layers) with EP for different output dimensionalities $D_o$ (orange star–cross markers). The resulting LoRA${+}$EP configurations form a new dominant region in the accuracy–parameter plane, \emph{strictly improving over both pure EP and pure LoRA}. For example, a hybrid setting with $850$K parameters achieves $76.99\%$ top-1 accuracy, improving over both the best pure EP configuration ($75.58\%$ with $1.38$M parameters) and the best all-layer LoRA variant ($76.72\%$ with $1.95$M parameters). At the low-parameter end, a LoRA${+}$EP configuration with only $250$K parameters already reaches $71.99\%$ accuracy, \ie about $4.3$\% above \cls linear probing ($67.66\%$) while using over $3\times$ fewer parameters. These results indicate that \emph{EP captures information that LoRA alone does not}, and vice versa: rather than being redundant with PEFT, EP provides a complementary, parameter-efficient probing (PEP) mechanism that remains beneficial.

\subsection{Experimental Analysis}
\label{sec:analysis}

\textbf{Impact of $W_K$ and $W_V$.}
% Our analysis in \autoref{sec:att-pred} posits that while a single learnable query $\vq$ can effectively absorb the key transformation $W_K$ in single-head attention, the same does not hold in multi-head. To empirically validate this, we probe MAE ViT-B pre-trained on IN-1K with four variants: single-head  \vs multi-head, each with and without $W_K$. Specifically, we evaluate single-head AbMILP and AIM with 12 heads (AIM$_{12}$). 
% In single-head attention, removing $W_K$ has minimal impact on performance (71.8\% $\to$ \ 71.7\%), while in multi-head the drop is noticeable (75.1\% $\to$ 72.9\%). 
% %
Our analysis in \autoref{sec:att-pred} posits that while a single learnable query $\vq$ effectively absorbs the key transformation $W_K$ in single-head attention, the same does not hold in multi-head. To empirically validate this, we probe MAE ViT-B pre-trained on ImageNet-1K with four variants: single-head ($M{=}1$)  \vs multi-head ($M{>}1$), each with (\ref{equ:mha-lrn-q}) \vs without $W_K$ (\ref{equ:mha-lrn-k-id}). Specifically, we evaluate  AbMILP and AIM for the single-head and multi-head (AIM$_{12}$ with 12 heads) case, respectively. 
In single-head attention, removing $W_K$ has minimal impact on performance (71.8\% $\to$ \ 71.7\%), while in multi-head the drop is noticeable (75.1\% $\to$ 72.9\%), since each query now interacts only with a subspace. Note that MHCA (\ref{equ:mha-lrn-q}), that natively includes $W_K$, in AIM$_{12}$ and our MQCA (\ref{equ:ours}), that natively does not includes $W_K$, in EP$_{12}$, configured with the same number of heads/queries ($M{=}12$), attain identical accuracy (75.1\%), reflecting their mathematical equivalence. However, EP$_{12}$ reaches this performance with far fewer parameters (1.36M for EP$_{12}$ \vs 1.95M for AIM$_{12}$), making its design more parameter-efficient.
We also ablate the effect of the value transformation $W_V$, which operates on patch tokens~\equ{att-pool}, by adding or removing it across pooling methods. Introducing $W_V$ to \gap results in a top-1 accuracy improvement from 66.7\% $\to$ 68.0\%. Conversely, removing $W_V$ from $\mOUR_{12}$ degrades performance from 75.1\% $\to$ 72.1\%. A similar accuracy drop is observed for other methods, such as AIM (75.1\% $\to$ 72.0\%) and CAE (74.9\% $\to$ 72.2\%), confirming that $W_V$ is a critical component.

%------------------------------------------------------------------------------
\begin{figure*}[ht]
\vspace{-4pt}
\centering
\begin{tabular}{cccc}
\begin{tikzpicture}
    \begin{axis}[
        width=4cm, height=4cm,
        ylabel={$\Delta$ accuracy},
        xlabel={localization},      
        grid=both,
        legend pos=south east,
        legend style={
            font=\tiny,
            row sep=0pt,
            inner xsep=2pt,
            inner ysep=0.5pt,
            nodes={scale=0.9, transform shape}
            },
        % Define custom colors
        cycle list={
            {cyan!70!black}, % AIM (Teal)
            {orange!80!black}, % EP (Deep Orange)
            {magenta!80!black} % V-Jepa (Dark Magenta)
        }
    ]

    % AIM (Teal/Blue)
    \addplot[only marks, color=cyan!70!black, mark=*] coordinates {
        (0.895, 1.120) (0.94571, 1.674) (0.95662, 1.760) (0.88158, 0.964) (0.93466, 1.460) (0.94453, 1.634)
        (0.76658, 0.668) (0.79474, 0.838) (0.87887, 1.066) (0.88662, 0.952) (0.88496, 1.460) (0.96872, 2.308)
    };
    \addlegendentry{AIM}

    % EP (Deep Orange)
    \addplot[only marks, color=orange!80!black, mark=square*] coordinates {
        (0.76840, 0.824) (0.83402, 0.896) (0.93551, 1.922) (0.99174, 1.752) (0.84595, 0.948) (0.86708, 1.040)
        (0.92411, 1.512) (0.93061, 1.426) (0.86252, 1.068) (0.93495, 1.450) (0.94457, 1.270) (0.97862, 1.604)
    };
    \addlegendentry{EP}

    % V-Jepa (Dark Magenta)
    \addplot[only marks, color=magenta!80!black, mark=triangle*] coordinates {
        (0.91864, 1.864) (0.87630, 1.256) (0.93334, 1.352) (0.93241, 1.124) (0.79426, 0.538) (0.96196, 2.080)
        (0.91032, 1.250) (0.98784, 1.744) (0.96148, 1.864) (0.91404, 1.380) (0.97769, 2.378) (0.76384, 0.370)
    };
    \addlegendentry{V-JEPA}

    \addplot[dashed, color=gray] coordinates { (0.75, 0) (1.05, 0) }; % Reference line at Δacc1 = 0

    \end{axis}
\end{tikzpicture}&
\begin{tikzpicture}
    \begin{axis}[
width=4cm, height=4cm,
        grid=both,
        xlabel={entropy},    
        legend pos=south west,
        legend style={
            font=\tiny,
            row sep=0pt,
            inner xsep=2pt,
            inner ysep=0.5pt,
            nodes={scale=0.9, transform shape}
            },
        % Define custom colors
        cycle list={
            {cyan!70!black}, % AIM (Teal)
            {orange!80!black}, % EP (Deep Orange)
            {magenta!80!black} % V-Jepa (Dark Magenta)
        }
    ]

    % AIM (Teal/Blue)
    \addplot[only marks, color=cyan!70!black, mark=*] coordinates {
        (5.43168, 1.120) (5.22279, 1.674) (5.12907, 1.760) (5.54051, 0.964) 
        (5.22057, 1.460) (5.19683, 1.634) (5.70033, 0.668) (5.66339, 0.838) 
        (5.44345, 1.066) (5.45223, 0.952) (5.38595, 1.460) (4.76420, 2.308)
    };
    \addlegendentry{AIM}

    % EP (Deep Orange)
    \addplot[only marks, color=orange!80!black, mark=square*] coordinates {
        (5.59843, 0.824) (5.60367, 0.896) (5.08845, 1.922) (4.65599, 1.752) 
        (5.60393, 0.948) (5.37573, 1.040) (5.20931, 1.512) (5.24228, 1.426) 
        (5.46595, 1.068) (5.07543, 1.450) (5.15721, 1.270) (4.94809, 1.604)
    };
    \addlegendentry{EP}

    % V-Jepa (Dark Magenta)
    \addplot[only marks, color=magenta!80!black, mark=triangle*] coordinates {
        (5.47964, 1.864) (5.63809, 1.256) (5.36647, 1.352) (5.44121, 1.124) 
        (5.75248, 0.538) (5.27593, 2.080) (5.50275, 1.250) (5.00644, 1.744) 
        (5.31168, 1.864) (5.49079, 1.380) (5.14293, 2.378) (5.76049, 0.370)
    };
    \addlegendentry{V-JEPA}

    \addplot[dashed, color=gray] coordinates { (4.5, 0) (5.8, 0) }; % Reference line at Δacc1 = 0

    \end{axis}
\end{tikzpicture}&
\begin{tikzpicture}
    \begin{axis}[
width=4cm, height=4cm,
        xlabel={localization},    
        grid=both,
        legend pos=south east,
        legend style={
            font=\tiny,
            row sep=0pt,
            inner xsep=2pt,
            inner ysep=0.5pt,
            nodes={scale=0.9, transform shape}
            }
    ]

    % EP D (Blue)
    \addplot[only marks, color=blue, mark=*] coordinates {
        (0.84203, 1.382)
        (0.81963, 1.682)
        (0.98276, 3.042)
        (0.87578, 1.628)
        (0.94905, 2.466)
        (0.90010, 1.988)
        (0.96586, 2.928)
        (0.98257, 2.398)
    };
    \addlegendentry{$D_o = D_i$}

    % EP D/2 (Purple)
    \addplot[only marks, color=purple, mark=triangle*] coordinates {
        (0.92017, 2.004)
        (0.94754, 2.360)
        (0.98200, 2.568)
        (0.97869, 2.688)
        (0.83448, 1.750)
        (0.89494, 1.516)
        (0.95120, 2.178)
        (0.87708, 1.478)
    };
    \addlegendentry{$D_o = D_i / 2$}

    % EP D/4 (Orange)
    \addplot[only marks, color=brown!60!orange, mark=square*] coordinates {
        (0.97872, 2.936)
        (0.87572, 1.690)
        (0.92744, 1.796)
        (0.93330, 2.006)
        (0.93970, 2.168)
        (0.94548, 2.344)
        (0.92897, 2.068)
        (0.95090, 2.826)
    };
    \addlegendentry{$D_o = D_i / 4$}

    \addplot[dashed, color=gray] coordinates { (0.75, 0) (1.0, 0) }; % Reference line at Δacc1 = 0

    \end{axis}
\end{tikzpicture}&
\begin{tikzpicture}
    \begin{axis}[
width=4cm, height=4cm,
        xlabel={entropy},            
        grid=both,
        legend pos=south west,
        legend style={
            font=\tiny,
            row sep=0pt,
            inner xsep=2pt,
            inner ysep=0.5pt,
            nodes={scale=0.9, transform shape}
            }
    ]

    % EP D (Blue)
    \addplot[only marks, color=blue, mark=*] coordinates {
        (5.64549, 1.382) (5.60842, 1.682) (4.93765, 3.042) (5.61278, 1.628)
        (5.21849, 2.466) (5.41558, 1.988) (4.99797, 2.928) (5.01118, 2.398)
    };
    \addlegendentry{$D_o = D_i$}

    % EP D/2 (Purple)
    \addplot[only marks, color=purple, mark=triangle*] coordinates {
        (5.56017, 2.004) (5.40978, 2.360) (5.02644, 2.568) (5.16251, 2.688)
        (5.68091, 1.750) (5.68105, 1.516) (5.47525, 2.178) (5.72602, 1.478)
    };
    \addlegendentry{$D_o = D_i / 2$}

    % EP D/4 (Orange)
    \addplot[only marks, color=brown!60!orange, mark=square*] coordinates {
        (5.23622, 2.936) (5.80825, 1.690) (5.73023, 1.796) (5.66501, 2.006)
        (5.64199, 2.168) (5.53140, 2.344) (5.61468, 2.068) (5.46096, 2.826)
    };
    \addlegendentry{$D_o = D_i / 4$}

    \addplot[dashed, color=gray] coordinates { (4.9, 0) (5.9, 0) }; % Reference line at Δacc1 = 0

    \end{axis}
\end{tikzpicture}\\
\end{tabular}
    \vspace{-6pt}
    \caption{\emph{Classification accuracy \vs attention quality} on ImageNet-1K. Each point corresponds to an attention predictor (head or query). $\Delta$ accuracy measures the drop when replacing an attention predictor’s distribution with uniform. Plots show relations to localization quality (1st, 3rd) and entropy (2nd, 4th). Left: different attentive probing methods; Right: varying $D_o$ for \OUR.}
    \vspace{-6pt}
    \label{fig:delta_acc_subplots}
\end{figure*}
%------------------------------------------------------------------------------

\textbf{Classification \vs localization.} We investigate whether the quality of attention maps in terms of localization contributes positively to classification accuracy. Localization quality is measured by (i) the attention mass within the ground-truth bounding box~\citep{imagenet} and (ii) the entropy of the attention distribution, averaged over the validation set. To estimate each predictor’s contribution to classification, we measure the accuracy drop when replacing its attention with a uniform distribution. \autoref{fig:delta_acc_subplots} shows a clear correlation: predictors with better localization and lower entropy exert stronger influence on accuracy. This holds across different attentive methods as well as \OUR with reduced output dimensionality $D_o$. As also seen in \autoref{fig:bench_datasets}, lowering $D_o$ degrades performance; the rightmost plots further suggest that this effect arises not only from reduced representational capacity but also from diminished attention quality, as reflected in higher entropy.

\begin{figure*}[t]
\centering
\scriptsize
\input{tab/imagenet_val_attmap_ours_bird}
\vspace{-8pt}
%\caption{\emph{Attention maps of \our} ($\mOUR_{8}$). Each query $\vq_i$ captures distinct and complementary regions, enabling a structured and semantic decomposition of visual information. MAE ViT-B pre-trained on ImageNet-1K, probed with \OUR.}
\caption{\emph{Attention maps of} $\mOUR_{8}$. Each query is assigned a distinct color. Semantic correspondences emerge (\eg, \textcolor{qcol3}{tails}, \textcolor{qcol4}{beaks}, \textcolor{qcol2}{feet}), with each query capturing complementary regions and enabling a structured decomposition of visual cues. MAE ViT-B pre-trained on IN-1K, probed with EP.}
\label{fig:attention-maps-ours-bird}
\end{figure*}
%------------------------------------------------------------------------------

\begin{figure}
    \centering
    \begin{tikzpicture}
% --- shared legend container name
\def\compllegend{compl_legend}

\begin{groupplot}[
  group style={group size=2 by 1, horizontal sep=1.3cm},
  height=4.5cm,
  ymin=0, ymax=0.92,             % <-- same y-scale for BOTH plots
  ymajorgrids,
  ybar,
  enlarge x limits=0.05
]

% ===== Left plot ============================================================
\nextgroupplot[
  width=0.76\textwidth,
  ylabel={Complementarity},
  symbolic x coords={MAE\\ViT-L,BEiTv2\\ViT-B,SimMIM\\ViT-B,iBOT\\ViT-B,
                     DINOv2\\ViT-B,DINOv3\\ViT-B,DINOv3\\ViT-L,Franca\\ViT-L,CLIP\\ViT-L,SigLIP\\ViT-L},
  xtick=data, xtick align=center,
  x tick label style={font=\scriptsize},
  legend to name=\compllegend,  % collect legend entries to a shared box
  nodes near coords,
  nodes near coords align={vertical},
  nodes near coords style={font=\tiny, scale=0.8, transform shape,/pgf/number format/fixed},
]
% blue bars
\addplot+[draw=blue,   fill=blue!70, bar width=9pt,
          every node near coord/.append style={font=\tiny, color=blue}]
  coordinates {(MAE\\ViT-L,0.27) (BEiTv2\\ViT-B,0.36) (SimMIM\\ViT-B,0.37)
               (iBOT\\ViT-B,0.44) (DINOv2\\ViT-B,0.24) (DINOv3\\ViT-B,0.27)
               (DINOv3\\ViT-L,0.18) (Franca\\ViT-L,0.16) (CLIP\\ViT-L,0.22) (SigLIP\\ViT-L,0.46)};
% orange bars
\addplot+[draw=orange, fill=orange!70, bar width=9pt,
          every node near coord/.append style={font=\tiny, color=orange}]
  coordinates {(MAE\\ViT-L,0.69) (BEiTv2\\ViT-B,0.77) (SimMIM\\ViT-B,0.67)
               (iBOT\\ViT-B,0.61) (DINOv2\\ViT-B,0.75) (DINOv3\\ViT-B,0.76)
               (DINOv3\\ViT-L,0.84) (Franca\\ViT-L,0.47) (CLIP\\ViT-L,0.75) (SigLIP\\ViT-L,0.75)};

% ===== Right plot ===========================================================
\nextgroupplot[
  width=0.24\textwidth,
  symbolic x coords={MAE\\ViT-B},
  xtick=data, xtick align=center,
  x tick label style={font=\scriptsize},
  yticklabels={},               % keep right axis clean
  legend to name=\compllegend,  % send legend entries here
  legend style={
    font=\tiny, draw=black, fill=white,
    inner xsep=2pt, inner ysep=0.5pt, row sep=0pt, nodes={scale=0.7, transform shape}
  },
  legend columns=1,
  nodes near coords,
  nodes near coords align={vertical},
  nodes near coords style={font=\tiny, scale=0.8, transform shape,/pgf/number format/fixed},
]
% order of plots = order of legend entries:
\addplot+[draw=blue,    fill=blue!70, bar width=9pt,
          every node near coord/.append style={font=\tiny, color=blue}]
  coordinates {(MAE\\ViT-B,0.24)};
\addplot+[draw=magenta, fill=magenta!70, bar width=9pt,
          every node near coord/.append style={font=\tiny, color=magenta}]
  coordinates {(MAE\\ViT-B,0.55)};
\addplot+[draw=cyan,    fill=cyan!70, bar width=9pt,
          every node near coord/.append style={font=\tiny, color=cyan}]
  coordinates {(MAE\\ViT-B,0.59)};
\addplot+[draw=orange,  fill=orange!70, bar width=9pt,
          every node near coord/.append style={font=\tiny, color=orange}]
  coordinates {(MAE\\ViT-B,0.65)};

\legend{MHSA, V-JEPA, AIM, EP} % <- shared legend entries

\end{groupplot}

% ===== Place the shared legend *inside the left plot* =======================
\node at ($(group c1r1.north east)+(+0.66cm,-1.35cm)$) {\ref{\compllegend}};
% tweak the shift (-0.55cm,-0.55cm) to sit in your desired blank area.

\end{tikzpicture}
    \vspace{-8pt}
    \caption{\emph{Complementarity scores of attention maps across different backbones and probings}. We compare the diversity of the internal \textcolor{blue}{MHSA heads in the last block} against the external \textcolor{orange}{\OUR\ queries} (left), \textcolor{magenta}{V-JEPA heads}, and \textcolor{cyan}{AIM heads} (right). Number of predictors are matched for fairness.}
    \vspace{-8pt}
    \label{fig:complementarity}
\end{figure}

\textbf{Attention complementarity.} 
In~\autoref{fig:attention-maps-ours-bird}, we jointly visualize the attention maps of $\OUR_8$. An emerging property of \OUR is that its queries specialize in different object regions, yielding complementary and interpretable attention patterns. Queries consistently attend to distinct parts, producing semantic correspondences across images and a structured decomposition of visual cues. To quantify this diversity, we define a \emph{complementarity} metric measuring how differently attention predictors distribute mass over patch tokens. For standard backbones, we extract the \cls$\!\!\to$patch attention from the internal last-block MHSA heads, and for \OUR the learned queries, with matched counts for fairness. We $L2$-normalize each distribution, compute pairwise cosine similarities, and define complementarity as one minus the average off-diagonal similarity. Higher values indicate more diverse (less redundant) attention. As shown in \autoref{fig:complementarity}, \OUR achieves significantly higher complementarity than MHSA and outperforms other probing approaches. More experiments in~\autoref{subsec:more_exp_analysis}.

\section{Conclusion}
\label{sec:conclusion}
\vspace{-4pt}
We revisit evaluation protocols for pre-training methods and introduce efficient probing (EP), a scalable alternative to fine-tuning. Lightweight yet expressive, EP delivers interpretable attention, strong generalization across paradigms, and consistent gains over linear probing—up to +24.3\% on ImageNet-1K. By comparing EP to standard parameter-efficient fine-tuning (PEFT) baselines, we show that EP is not only competitive on its own but also complementary: hybrid configurations achieve the best accuracy–parameter trade-offs, hinting at new research directions.

%Beyond evaluation, our analysis reveals emerging properties, hinting at new research directions.

%\OUR is a lightweight attentive probing method that eliminates redundant projections and leverages multiple learnable queries for efficient and expressive feature aggregation. It produces interpretable attention maps with strong localization, generalizes well across models and pretraining paradigms, and consistently outperforms linear probing, achieving up to +13.6\% on ImageNet-1k.

\textbf{Acknowledgments.} This work was supported by the EU Horizon Europe programme MSCA PF RAVIOLI (No. 101205297) and the Junior Star GACR GM 21-28830M. We acknowledge VSB–Technical University of Ostrava, IT4Innovations National Supercomputing Center, Czech Republic, for awarding this project (OPEN-33-67) access to the LUMI supercomputer, owned by the EuroHPC Joint Undertaking, hosted by CSC (Finland) and the LUMI consortium, through the Ministry of Education, Youth and Sports of the Czech Republic via the e-INFRA CZ project (ID: 90254). AWS resources were provided by the National Infrastructures for Research and Technology GRNET and funded by the EU Recovery and Resiliency Facility.

\bibliographystyle{iclr2026_conference}
\bibliography{iclr2026_conference}

\clearpage
\appendix

\section*{Appendix}
\addcontentsline{toc}{section}{Appendix}
\startcontents[supp] 

\begingroup
  \setlength{\parskip}{0pt}%
  \linespread{1.4}\selectfont
  \setcounter{tocdepth}{4}
  \printcontents[supp]{l}{1}{}%
\endgroup

\clearpage

\section{Additional Related Works}
\label{sec:more_related}

\subsection{Pooling}
\label{subsec:pooling}

Pooling reduces spatial resolution while retaining semantic information. In CNNs, fixed pooling (\eg, global average pooling~\citep{nin, resnet}) is standard; in vision transformers (ViTs)~\citep{vit}, the \cls token aggregates features via self-attention.

Recent work proposes attention-based pooling to enhance representation quality. \simpool~\citep{psomas2023simpool} replaces global average pooling using trainable attention in both CNNs and ViTs. Vision-language models such as \clip~\citep{radford2021clip}, \siglip~\citep{timm}, and \coca~\citep{yu2022coca} use attentive pooling or cross-attention to fuse modalities. \vjepa~\citep{v-jepa} applies cross-attention pooling for probing pretrained representations. In image retrieval, \delf~\citep{delf} and \dolg~\citep{yang2021dolg} use spatial attention to focus on salient regions. \cait~\citep{cait} improves class-token attention, \abmilp~\citep{rymarczyk2021abmilp} uses single-query pooling for multiple-instance learning, and \cbam~\citep{woo2018cbam} combines channel and spatial attention to recalibrate features. Although these poolings are originally introduced in diverse contexts, we repurpose them for probing of frozen pre-trained models, enabling a fair and comprehensive benchmark.

\subsection{Parameter-efficient Fine-tuning}
\label{subsec:peft}

Parameter-efficient fine-tuning (PEFT) adapts large pre-trained models to downstream tasks without updating all model parameters. PEFT techniques broadly include \emph{additive}, \emph{selective}, and \emph{low-rank adaptation} methods.

\paragraph{Additive.} Additive methods introduce small, task-specific modules into the frozen backbone, leaving the pre-trained weights untouched. These modules often reside within the transformer blocks and are trained to specialize the model for a new domain. Notable examples include AdapterFusion~\citep{pfeiffer2020adapterfusion}, LeTS~\citep{pmlr-v139-fu21a}, and TADA~\citep{hung2023tada} in natural language processing (NLP), VPT{~\citep{jia2022visualprompt}, AdaptFormer~\citep{chen2022adaptformer}, and Adapter-X~\citep{adapterX} in computer vision (CV), and FMA~\citep{fma}, AiRs~\citep{airs}, and DEFLECT~\citep{deflect} in remote sensing (RS).

\paragraph{Selective.} Selective methods fine-tune only specific subsets of parameters, typically chosen based on their functional role or estimated importance. Examples include BitFit~\citep{bitfit}, which adjusts only the bias terms, and norm-tuning approaches~\citep{layernorm} which update only the normalization layers. These techniques avoid introducing new components, making them lightweight, though sometimes at the expense of performance.

\paragraph{Low-rank adaptation.} Low-rank adaptation methods like LoRA~\citep{lora} in NLP assume that parameter updates lie in a low-dimensional subspace. They inject trainable low-rank matrices into existing layers, yielding strong performance with minimal parameter growth. In the vision domain, LoRa and its variants have been effectively adapted to vision transformers (ViTs), often rethinking where and how low-rank modules are inserted to align with the spatial and hierarchical nature of visual representations. Notable examples include structure-aware methods like Serial LoRA~\citep{zhong2025} and Flat-LoRA~\citep{li2025flatlora}, layer-wise extensions such as AdaptFormer~\citep{chen2022adaptformer}, and task specific designs like PETAH~\citep{augustin2024petah} and MeLo~\citep{zhu2024melo}, which adapt LoRa to mobile inference and medical imaging, respectively. Continued pretraining approaches such as ExPLoRA~\citep{explora} further extend low-rank adaptation to domain-shifted self-supervised settings.  

\vspace{1\baselineskip}

\OUR naturally fits the additive PEFT family. It introduces a compact learnable query set interacting with frozen tokens via multi-head cross-attention. Unlike typical prompt-based methods, it avoids backbone modifications and focuses training on minimal parameters. Thus, \OUR efficiently combines additive PEFT simplicity with task-specific attentive pooling.

\section{Additional Methods}
\label{sec:more_methods}

We provide here additional attentive pooling/probing methods evaluated in~\autoref{sec:results} but not described in detail in~\autoref{sec:var}. These approaches represent variations of the MHCA framework, highlighting only their key deviations from the default design.

\paragraph{CLIP.} CLIP~\citep{radford2021clip} differs from MHCA by employing self-attention rather than cross-attention. Specifically, CLIP prepends a global average pooled (GAP) token to the layer-normalized input features, treating this token as a global representation. All tokens, including the GAP token, are augmented with learnable positional encodings and processed through a single self-attention block (which includes a query projection matrix $W_Q$). The global representation is extracted from the output corresponding to the GAP token. Additionally, CLIP includes a linear projection matrix $W_P$ after attention aggregation. These modifications enable interactions across all tokens but increase parameter count and computational complexity.

\paragraph{CAiT.} CAiT~\citep{cait} adapts the MHCA-with-learnable-query formulation by concatenating the learnable query token with the input features and applying self-attention rather than cross-attention. It retains the query projection matrix $W_Q$ and includes a linear projection matrix $W_P$ after attention aggregation, followed by an MLP block with GELU activations, residual connections, and LayerScale parameters, similar to V-JEPA. The global representation is obtained from the updated query token after these operations, thus increasing complexity and parameter count relative to the default MHCA variant.

%\paragraph{CLIP} CLIP~\citep{radford2021clip} deviates from the MHCA framework by performing self-attention rather than cross-attention. It layer-normalizes the input features and prepends a global average pooled (GAP) token to them, treating this as a surrogate for a classification token. The GAP token and input features are jointly processed with learnable positional encodings and a single self-attention block. Attention is computed among all $(N+1)$ tokens, and the output corresponding to the GAP token is used as the final global image representation. CLIP also includes a learnable output projection matrix $W_{proj}$ applied after the attention aggregation. These modifications introduce additional parameters and higher complexity compared to standard MHCA, but allow for interaction among all tokens, including the global one.

%\paragraph{SigLIP} SigLIP~\citep{siglip} aligns with the MHCA framework using a learnable query, but retains a query projection matrix $W_Q$, instead of absorbing it into the token. After attention aggregation, SigLIP applies an output projection matrix $W_{proj}$ followed by an MLP block with GELU activation and residual connection, forming a transformer-like structure. Optional layer normalization is also supported before the MLP. These architectural choices introduce additional parameters and computational cost compared to the default variant of the framework.

\paragraph{SigLIP.} SigLIP~\citep{siglip, timm} remains close to the MHCA-with-learnable-query formulation but retains the query projection matrix $W_Q$. After the attention aggregation, SigLIP incorporates an output projection $W_P$, followed by a transformer-style MLP block with GELU activation and residual connections, similar to V-JEPA and CAiT. Optional layer normalization can also be applied before the MLP. These changes add further parameters and computational overhead compared to the baseline MHCA design.

%\paragraph{CAiT} CAiT~\citep{cait} follows the MHCA framework with a learnable query, but modifies it by first concatenating the query token with the input features and applying self-attention instead of cross-attention. It includes an output projection after attention, followed by a MLP with GELU activation, residual connections, and LayerScale parameters. The output corresponding to the updated learnable query token is used as the final global image representation. These additions increase both parameter count and computational cost compared to the default framework.

\paragraph{CAE.} CAE~\citep{chen2023cae} follows the MHCA-with-learnable-query template closely but retains the query projection matrix $W_Q$ and applies separate layer normalization to both input features and the query token prior to attention. After attention aggregation, it employs an additional output projection matrix $W_P$. These modifications introduce additional parameters and computational complexity.

%\paragraph{CAE} CAE~\citep{chen2023cae} follows the MHCA framework with a learnable query, but retains the query projection matrix $W_Q$ and applies separate layer normalization to the input features and the query token before computing attention. After attention aggregation, it includes an output projection $W_{proj}$. These modifications introduce additional parameters and computational cost compared to the default framework.

%\paragraph{CoCa} Contrastive Captioner~\citep{yu2022coca} stays within the MHCA framework with a learnable query, but keeps the query projection matrix $W_Q$ and layer-normalises the query token before attention. Both the attention computation and the value aggregation are carried out in a reduced sub-space of dimension $D_a=D_o< D_i$. A final linear layer $W_{proj}$ up-projects the result back to $D_i$.

\paragraph{CoCa} CoCa~\citep{yu2022coca} is aligned with the MHCA-with-learnable-query framework but retains the query projection matrix $W_Q$ and layer-normalizes the query token before computing attention. Attention and value aggregation both occur in a reduced-dimensional space, with dimension $D_a = D_o < D_i$. A final linear projection matrix $W_{\text{proj}}$ is then applied to restore the feature dimension to the original backbone dimension $D_i$. These choices introduce a controlled amount of additional complexity and parameters.

\vspace{1\baselineskip}

%------------------------------------------------------------------------------
\begin{figure*}[ht]
\vspace{-10pt}
    \centering
    \includegraphics[width=\textwidth]{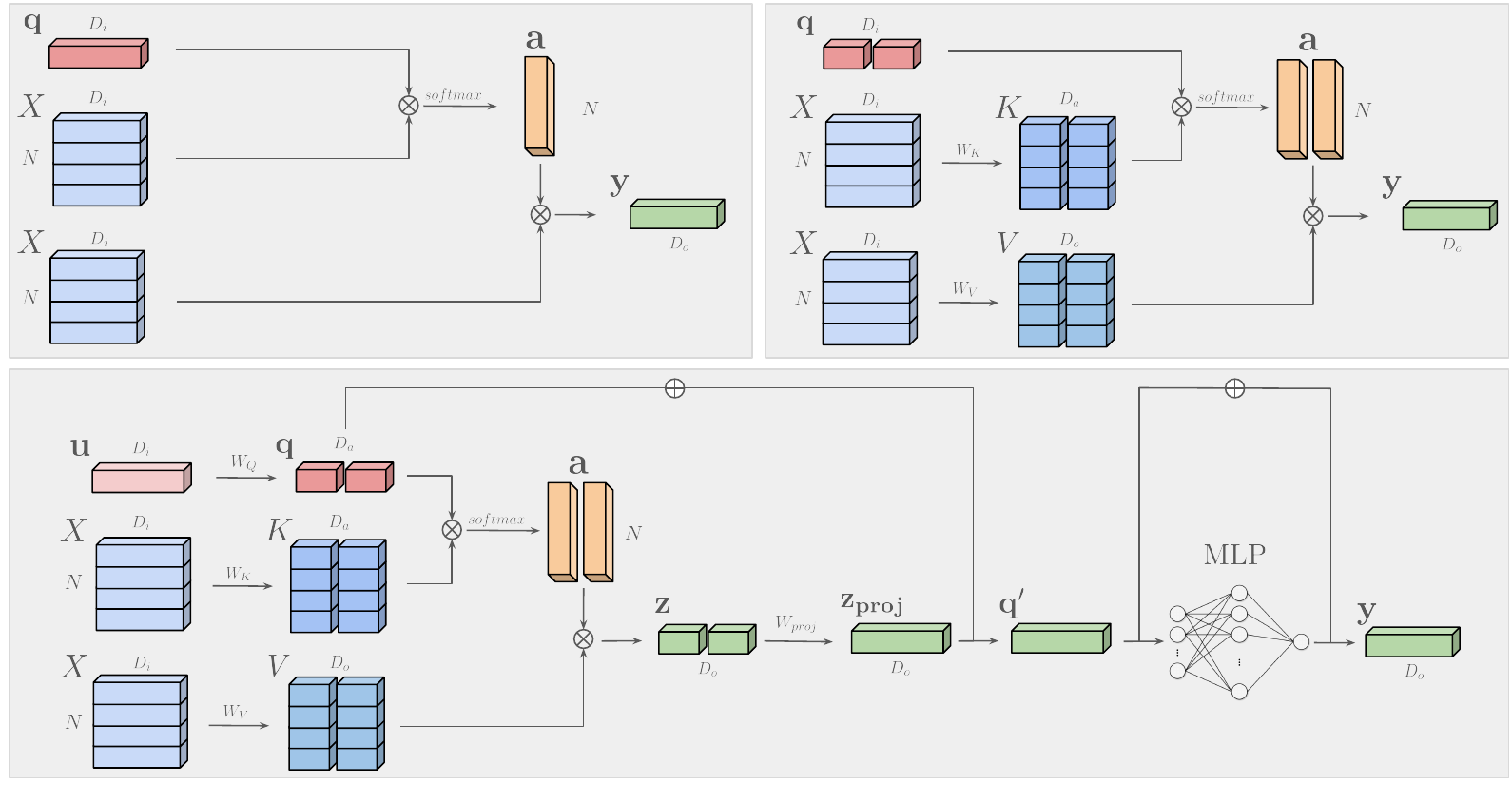}
    \vspace{-14pt}
    \caption{\emph{Visual comparison of three attentive pooling/probing methods}. AbMILP (top-left) employs a single-head, learnable query without linear projections, minimizing complexity. AIM (top-right) extends the approach by introducing multi-head attention, operating in multiple subspaces, and applies linear projections to keys and values. V-JEPA (bottom) offers a more comprehensive architecture by integrating multi-head attention with extensive linear projections and an additional MLP block with a residual connection, increasing representational capacity.}
    \label{fig:variants}
    \vspace{-6pt}
\end{figure*}
%------------------------------------------------------------------------------

In~\autoref{fig:variants} we present a visual comparison of three selected attentive pooling/probing techniques: AbMILP~\citep{rymarczyk2021abmilp}, AIM~\citep{aim}, and V-JEPA~\citep{v-jepa}. AbMILP (top-left) serves as a lightweight method, employing a single-head learnable query without additional linear projection matrices, thus requiring only $D_i$ parameters. AIM (top-right) extends this by adopting multi-head cross-attention, operating within multiple subspaces. This approach introduces linear projection matrices for keys and values, increasing the number of parameters, yet allowing more expressive query-key interactions. V-JEPA (bottom) represents a significantly more complex and computationally intensive architecture. Beyond multi-head attention and multiple linear projections for queries, keys, and values, it integrates an additional projection step, followed by a multi-layer perceptron (MLP) featuring GeLU activation and residual connections.

\section{Additional  Experiments}
\label{sec:more_exp}

\subsection{Experimental Setup}
\label{subsec:more_exp_setup}

\paragraph{Datasets.} We evaluate attentive probing across diverse image classification benchmarks. As a large-scale dataset, ImageNet-1K~\citep{imagenet} serves as the primary benchmark, containing 1.28M images across 1,000 categories. CIFAR-100~\citep{cifar} provides a smaller yet challenging 100-class task with 60K images. To assess scene understanding, we use Places365~\citep{places365}, comprising 1.8M images spanning 365 scene types. For fine-grained classification, we evaluate on CUB-200~\citep{cub} (11,788 images, 200 bird species), FGVC Aircraft~\citep{fgvc} (10K images, 100 aircraft models), Stanford Cars~\citep{cars} (16K images, 196 car types), and Food-101~\citep{food} (101K images, 101 food categories). Together, these datasets span a wide spectrum of scales and challenges—large-scale \vs small-scale, generic \vs fine-grained, and object- \vs scene-centric—providing a comprehensive testbed for probing methods.

\begin{wraptable}{r}{0.48\textwidth}
\vspace{-13pt}
% \begin{table}[h!]
\caption{\emph{AbMILP architecture ablation} increasing MLP depth. \textsc{\# Par.}: number of parameters.}
\centering
\footnotesize
\vspace{-5pt}
\begin{tabular}{lcc}
\toprule
\textsc{Method} & \textsc{\# Par.} & \textsc{Accuracy} \\
\midrule
AbMILP (depth 1) & 769{,}769 & 71.74 \\
AbMILP (depth 2) & 1{,}360{,}361 & 72.25 \\
AbMILP (depth 3) & 1{,}950{,}953 & 72.84 \\
\bottomrule
\end{tabular}
\label{tab:abmilp_depth}
% \end{table}
\vspace{-5pt}
\end{wraptable}

\paragraph{Pooling/probing methods.} We adopt AbMILP with \texttt{depth=1}, which reduces AbMILP to the formulation described in~\autoref{tab:pooling-variants} and~\autoref{sec:var}. In~\autoref{tab:abmilp_depth}, we evaluate AbMILP on ImageNet-1K with larger depths (\texttt{depth=2,3}), which correspond to the MLP-based variants and scale as~$\mathcal{O}(D_i^2)$, exploring their parameter–accuracy tradeoffs. Although deeper MLPs introduce substantially more parameters (\(+\!\mathcal{O}(D_i^2)\)), the gains in accuracy are marginal. To ensure a fair comparison in the \emph{accuracy-\vs-parameter-efficiency} setting, we consider the most competitive AbMILP variant (\texttt{depth=1}) as default, which successfully lies on the Pareto front.

\paragraph{Implementation details.} We evaluate 15 models spanning five pre-training paradigms: four masked image modeling (MAE~\citep{mae}, BEiTv2~\citep{beitv2}, SimMIM~\citep{simMIM}, CAPI~\citep{capi}), two joint-embedding (BYOL~\citep{byol}, DINO~\citep{caron2021emerging}), two hybrid (iBOT~\citep{ibot}, DINOv2~\citep{dinov2}), three vision-language (CLIP~\citep{radford2021clip}, SigLIP~\citep{siglip}, SigLIP2~\citep{siglip2}), and two generative—DiT (diffusion)~\citep{dit} and AIMv2 (autoregressive)~\citep{aimv2}. Architectures range from small (\eg, ViT-S for MAE) to extra-large (\eg, DiT-XL for DiT). 

To ensure a fair comparison of pooling/probing methods, we mostly adopt the LARS optimizer~\cite{you2017large} and conduct a learning rate search in the range [0.1, 5.0] with a step size of 0.1 for each model. For large-scale datasets such as ImageNet-1K and Places365, we fix the learning rate to 0.1 due to the computational cost of an exhaustive search. All models are trained for 90 epochs with 10 warm-up epochs, ensuring consistent training schedules even though many models converge much earlier (\eg, SigLIP within 15 epochs). The effective batch size is set to 4096 for all datasets except FGVC Aircraft, where it is reduced to 512 due to the smaller dataset size. Data augmentation follows standard PyTorch~\citep{paszke2019pytorch} image pre-processing: RandomResizedCrop, horizontal flipping, and normalization. For vision-language models, we adopt their official preprocessing pipelines (\eg, OpenCLIP~\citep{openclip} transforms for CLIP, SigLIP, and SigLIP2) to ensure alignment with pre-training distributions. Experiments are conducted on a cluster of 8 NVIDIA A100 GPUs (40 GB VRAM each) and on the LUMI supercomputer, on clusters of 8 AMD Instinct MI250X GPUs (128 GB HBM2e memory each).

\subsection{Experimental Results}
\label{subsec:more_exp_results}
%------------------------------------------------------------------------------
\begin{figure*}[ht]
    \centering
    \begin{subfigure}[t]{0.48\textwidth}
        \centering
        \caption{MAE ViT-S ImageNet-1k}
        % Scatter Plot
%\begin{figure}[ht]
%\centering
\begin{tikzpicture}
    \begin{axis}[
        width=7.3cm, % Adjust the width as needed
        height=6cm,  % Adjust the height as needed
        ylabel={top-1 accuracy (\%)},
        grid=major,
        enlargelimits=true,
        xmode=log,
%         log ticks with fixed point,
        legend pos=south east,
        legend style={font=\tiny}
    ]

%-----------------------------------------------------------------
% 1) Baseline Methods (Black)
%-----------------------------------------------------------------
\addplot[
    only marks,
    mark=*,
    draw=black,
    fill=black,
    mark size=1.5pt,
    nodes near coords,
    every node near coord/.append style={font=\tiny,anchor=south, yshift=-1pt, xshift=-8pt},
    point meta=explicit symbolic, % Set meta data source
]
table [
    meta=Method,
    x=Parameters,
    y=Accuracy,
    col sep=space,
] {
    Method Accuracy Parameters
    [CLS] 47.42 385000
    %GAP 47.14 385000
    %CBAM 47.05 403532
    %CLIP 63.46 1052776
    %SigLIP 66.54 2158312
    %SimPool 54.34 533608
    %AIM 63.60 680296
    %CoCa 62.75 903016
    CaiT 62.38 2161384
    ViT 68.93 2158312
    %V-Jepa 66.71 2159848
    %DELF 56.30 533993
    AbMILP 54.43 385384
    %CAE 62.84 830056
    %EP$_2$ 53.67 533224
    EP$_4$ 61.20 533992
    EP$_8$ 62.83 535528
    EP$_{64}$ 64.60 557032
};
\addlegendentry{$D_o = D_i$}

% 3) D_i/2 (blue)
%-----------------------------------------------------------------
\addplot[
    only marks,
    mark=*,
    draw=blue,
    fill=blue,
    mark size=1.5pt,
    nodes near coords,
    every node near coord/.append style={font=\tiny,anchor=south, yshift=-1pt, xshift=-8pt},
    point meta=explicit symbolic,
]
table [
    meta=Method,
    x=Parameters,
    y=Accuracy,
    col sep=space,
] {
    Method Accuracy Parameters
    EP$_2$ 56.55 267496
    EP$_4$ 58.50 268264
    EP$_{16}$ 60.15 272872
    EP$_{96}$ 61.21 303592
};
\addlegendentry{$D_o = D_i / 2$}

%-----------------------------------------------------------------
% 4) D_i/4 (orange)
%-----------------------------------------------------------------
\addplot[
    only marks,
    mark=*,
    draw=orange,
    fill=orange,
    mark size=1.5pt,
    nodes near coords,
    every node near coord/.append style={font=\tiny,anchor=south, yshift=-1pt, xshift=-8pt},
    point meta=explicit symbolic,
]
table [
    meta=Method,
    x=Parameters,
    y=Accuracy,
    col sep=space,
] {
    Method Accuracy Parameters
    EP$_2$ 52.42 134632
    EP$_4$ 53.69 135400
    EP$_{16}$ 55.02 140008
    EP$_{48}$ 55.96 152296
};
\addlegendentry{$D_o = D_i / 4$}

%-----------------------------------------------------------------
% 5) D_o/8 (purple)
%-----------------------------------------------------------------
\addplot[
    only marks,
    mark=*,
    draw=purple,
    fill=purple,
    mark size=1.5pt,
    nodes near coords,
    every node near coord/.append style={font=\tiny,anchor=south, yshift=-1pt, xshift=-8pt},
    point meta=explicit symbolic,
]
table [
    meta=Method,
    x=Parameters,
    y=Accuracy,
    col sep=space,
] {
    Method Accuracy Parameters
    EP$_2$ 46.50 68200
    EP$_8$ 47.60 70504
    EP$_{24}$ 48.60 76648
    %EP$_{48}$ 70.27 207592
};
\addlegendentry{$D_o = D_i / 8$}

%-----------------------------------------------------------------
% Manual positions
%-----------------------------------------------------------------
\addplot[
    only marks,
    mark=*,
    draw=black,
    fill=black,
    mark size=1.5pt,
]
coordinates {
    (385000, 47.14) % GAP
    (403532, 47.05) % CBAM
    (533608, 54.34) % SimPool
    (1052776, 63.46) % CLIP
    (903016, 62.75) % CoCa
    (533993, 56.30) % DELF
    (680296, 63.60) % AIM
    (830056, 62.84) % CAE
    (2158312, 66.54) % SigLIP
    (2159848, 66.71) % V-JEPA
};

\node[anchor=north, font=\tiny, yshift=1pt, xshift=8pt] at (403532, 47.05) {CBAM};
\node[anchor=north, font=\tiny, yshift=1pt, xshift=-8pt] at (403532, 47.14) {GAP};
\node[anchor=north, font=\tiny, yshift=1pt, xshift=10pt] at (533608, 54.34) {SimPool};
\node[anchor=south, font=\tiny, yshift=0pt, xshift=6pt] at (1052776, 63.46) {CLIP};
\node[anchor=north, font=\tiny, yshift=1pt, xshift=7pt] at (903016, 62.75) {CoCa};
\node[anchor=north, font=\tiny, yshift=1pt, xshift=7pt] at (533993, 56.30) {DELF};
\node[anchor=south, font=\tiny, yshift=0pt, xshift=0pt] at (680296, 63.60) {AIM};
\node[anchor=north, font=\tiny, yshift=1pt, xshift=-6pt] at (830056, 62.84) {CAE};
\node[anchor=north, font=\tiny, yshift=0pt, xshift=2pt] at (2158312, 66.54) {SigLIP};
\node[anchor=south, font=\tiny, yshift=0pt, xshift=2pt] at (2159848, 66.71) {V-JEPA};
%-----------------------------------------------------------------
% Pareto Front
%-----------------------------------------------------------------
\path [name path=pareto]
    (385000, 40)
    (385000, 47.42)
    -- (385384, 54.43)
    -- (680296, 63.60)
    -- (2158312, 68.93);

% Upper boundary (extend down to lowest y-value)
\path [name path=top]
    (12000000, 68.93) -- (12000000, 40);

% Fill the area to the right of the Pareto front
\addplot [fill=blue!30, opacity=0.3]
    fill between [of=pareto and top];

% --- Pareto front line ---
\addplot[thick, dashed, blue, mark=none] coordinates {
    (385000, 47.42)
    (385384, 54.43)
    (680296, 63.60)
    (2158312, 68.93)
};

    \end{axis}
\end{tikzpicture}
%\caption{Scatter plot of Top-1 Accuracy (\%) vs. Number of Parameters on ImageNet-1k.}
%\label{fig:accuracy_parameters_scatter}
%\end{figure}
        \label{fig:bench-mae-vitb}
    \end{subfigure}
    \vspace{-12pt}
    \begin{subfigure}[t]{0.48\textwidth}
        \centering
        \caption{MAE ViT-L ImageNet-1k}
        % Scatter Plot
%\begin{figure}[ht]
%\centering
\begin{tikzpicture}
    \begin{axis}[
        width=7.3cm, % Adjust the width as needed
        height=6cm,  % Adjust the height as needed
        grid=major,
        enlargelimits=true,
        xmode=log,
%         log ticks with fixed point,
        legend pos=south east,
        legend style={font=\tiny}
    ]

%-----------------------------------------------------------------
% 1) Baseline Methods (Black)
%-----------------------------------------------------------------
\addplot[
    only marks,
    mark=*,
    draw=black,
    fill=black,
    mark size=1.5pt,
    nodes near coords,
    every node near coord/.append style={font=\tiny,anchor=south, yshift=-1pt, xshift=-8pt},
    point meta=explicit symbolic, % Set meta data source
]
table [
    meta=Method,
    x=Parameters,
    y=Accuracy,
    col sep=space,
] {
    Method Accuracy Parameters
    [CLS] 76.01 1025000
    GAP 73.24 1025000
    %CBAM 73.21 1156172
    CLIP 79.18 5427176
    %SigLIP 79.34 13618152
    %SimPool 78.09 3124200
    %AIM 79.09 3123176
    %CoCa 78.43 2406376
    CaiT 77.91 13626344
    ViT 79.31 13618152
    V-JEPA 78.29 13622248
    DELF 78.19 2077673
    AbMILP 77.80 1026025
    CAE 78.71 5229544
    %EP$_2$ 78.60 2075624
    EP$_4$ 78.80 2077672
    EP$_8$ 79.11 2081768
    %EP$_{64}$ 79.25 2139112
    EP$_{128}$ 79.38 2204648
};
\addlegendentry{$D_o = D_i$}

% 3) D_i/2 (blue)
%-----------------------------------------------------------------
\addplot[
    only marks,
    mark=*,
    draw=blue,
    fill=blue,
    mark size=1.5pt,
    nodes near coords,
    every node near coord/.append style={font=\tiny,anchor=south, yshift=-1pt, xshift=-8pt},
    point meta=explicit symbolic,
]
table [
    meta=Method,
    x=Parameters,
    y=Accuracy,
    col sep=space,
] {
    Method Accuracy Parameters
    EP$_2$ 78.36 1039336
    EP$_4$ 78.63 1041384
    EP$_{16}$ 79.14 1053672
    %EP$_{96}$ 61.21 303592
};
\addlegendentry{$D_o = D_i / 2$}

%-----------------------------------------------------------------
% 4) D_i/4 (orange)
%-----------------------------------------------------------------
\addplot[
    only marks,
    mark=*,
    draw=orange,
    fill=orange,
    mark size=1.5pt,
    nodes near coords,
    every node near coord/.append style={font=\tiny,anchor=south, yshift=-1pt, xshift=-8pt},
    point meta=explicit symbolic,
]
table [
    meta=Method,
    x=Parameters,
    y=Accuracy,
    col sep=space,
] {
    Method Accuracy Parameters
    EP$_2$ 77.65 521192
    EP$_8$ 78.24 528360
    EP$_{16}$ 78.62 535528
    EP$_{64}$ 78.81 584680
};
\addlegendentry{$D_o = D_i / 4$}

%-----------------------------------------------------------------
% 5) D_o/8 (purple)
%-----------------------------------------------------------------
\addplot[
    only marks,
    mark=*,
    draw=purple,
    fill=purple,
    mark size=1.5pt,
    nodes near coords,
    every node near coord/.append style={font=\tiny,anchor=south, yshift=-1pt, xshift=-8pt},
    point meta=explicit symbolic,
]
table [
    meta=Method,
    x=Parameters,
    y=Accuracy,
    col sep=space,
] {
    Method Accuracy Parameters
    EP$_2$ 76.61 262120
    EP$_8$ 76.91 268264
    EP$_{32}$ 77.25 292840
    %EP$_{48}$ 70.27 207592
};
\addlegendentry{$D_o = D_i / 8$}

%-----------------------------------------------------------------
% Manual positions
%-----------------------------------------------------------------
\addplot[
    only marks,
    mark=*,
    draw=black,
    fill=black,
    mark size=1.5pt,
]
coordinates {
    (3123176, 79.09) % AIM
    (3124200, 78.09) % SimPool
    (2406376, 78.43) % CoCa
    (13618152, 79.34) % SigLIP
};
\node[anchor=south, font=\tiny, yshift=0pt, xshift=-2pt] at (3123176, 79.09) {AIM};
\node[anchor=south, font=\tiny, yshift=-5pt, xshift=12pt] at (3124200, 78.09) {SimPool};
\node[anchor=south, font=\tiny, yshift=-5pt, xshift=8pt] at (2406376, 78.43) {CoCa};
\node[anchor=south, font=\tiny, yshift=-1pt, xshift=6pt] at (13618152, 79.34) {SigLIP};
%-----------------------------------------------------------------
% Pareto Front
%-----------------------------------------------------------------
\path [name path=pareto]
    (1025000, 70)
    (1025000, 76.01)
    -- (1026025, 77.80)
    -- (2077673, 78.19)
    -- (2406376, 78.43)
    -- (3123176, 79.09)
    -- (13618152, 79.34);

% Upper boundary (extend down to lowest y-value)
\path [name path=top]
    (20000000, 79.34) -- (20000000, 40);

% Fill the area to the right of the Pareto front
\addplot [fill=blue!30, opacity=0.3]
    fill between [of=pareto and top];

% --- Pareto front line ---
\addplot[thick, dashed, blue, mark=none] coordinates {
    (1025000, 76.01)
    (1026025, 77.80)
    (2077673, 78.19)
    (2406376, 78.43)
    (3123176, 79.09)
    (13618152, 79.34)
};

    \end{axis}
\end{tikzpicture}
%\caption{Scatter plot of Top-1 Accuracy (\%) vs. Number of Parameters on ImageNet-1k.}
%\label{fig:accuracy_parameters_scatter}
%\end{figure}
        \label{fig:bench-mae-vitl}
    \end{subfigure}
    \begin{subfigure}[t]{0.48\textwidth}
        \centering
        \caption{SimMIM ViT-B ImageNet-1k}
        % Scatter Plot
%\begin{figure*}[ht]
%\centering
\begin{tikzpicture}
    \begin{axis}[
        width=7.3cm, % Adjust the width as needed
        height=6cm,  % Adjust the height as needed
        ylabel={top-1 accuracy (\%)},
        grid=major,
        enlargelimits=true,
        xmode=log,
%         log ticks with fixed point,
        legend pos=south east,
        legend style={font=\tiny}
    ]

%-----------------------------------------------------------------
% 1) Do POINTS (black)
%-----------------------------------------------------------------
\addplot[
only marks,
mark=*,         % marker shape
draw=black,
fill=black,
mark size=1.5pt,
nodes near coords,
every node near coord/.append style={font=\tiny,anchor=south, yshift=-1pt, xshift=-8pt},
point meta=explicit symbolic, % Set meta data source
  % row predicate: keep only the rows with Color = 'black'
  % If you have trouble with quotes, you can do { \thisrow{Color} == "black" }
  % or for older pgfplots versions:
  % restrict expr to domain={\thisrow{Color}=="black":true}{false},
  %
  % or do a quick numerical test if you place e.g. 1 for black, 2 for red in a separate column
]
table[x=Parameters, y=Accuracy, col sep=space, meta=Method]
{
Method Accuracy Parameters
[CLS] 51.50 769000
GAP 54.30 769000
%CLIP 62.10 3284200
%SigLIP 65.07 7854568
%SimPool 54.54 1950184
%AIM 63.98 1949416
CBAM 45.90 842828
%CoCa 60.80 1805032
%CaiT 63.71 7860712
ViT 67.75 7854568
%V-JEPA 65.47 7857640
DELF 55.64 1361897
%AbMILP 53.19 769769
%CAE 62.98 3135976
EP$_{2}$ 58.80 1360360
EP$_{4}$ 60.95 1361896
EP$_{8}$ 63.01 1364968
%EP$_{16}$ 64.14 1371112
%EP$_{32}$ 64.92 1383400
EP$_{64}$ 65.12 1407976
};
\addlegendentry{$D_o = D_i$}

%-----------------------------------------------------------------
% 2) Do/2 POINTS (blue)
%-----------------------------------------------------------------
\addplot[
only marks,
mark=*,         % marker shape
draw=blue,
fill=blue,
mark size=1.5pt,
nodes near coords,
every node near coord/.append style={font=\tiny,anchor=south, yshift=-1pt, xshift=-8pt},
point meta=explicit symbolic, % Set meta data source
]
table[x=Parameters, y=Accuracy, col sep=space, meta=Method]
{
Method Accuracy Parameters
EP$_2$ 56.65 681448
EP$_4$ 58.65 683752
EP$_8$ 60.56 686056
%EP$_16$ 61.88 692200
EP$_{64}$ 62.73 729064
};
\addlegendentry{$D_o = D_i / 2$}

%-----------------------------------------------------------------
% 2) Do/4 POINTS (red)
%-----------------------------------------------------------------
\addplot[
only marks,
mark=*,         % marker shape
draw=orange,
fill=orange,
mark size=1.5pt,
nodes near coords,
every node near coord/.append style={font=\tiny,anchor=south, yshift=-1pt, xshift=-8pt},
point meta=explicit symbolic, % Set meta data source
]
table[x=Parameters, y=Accuracy, col sep=space, meta=Method]
{
Method Accuracy Parameters
EP$_2$ 54.73 341992
EP$_8$ 57.74 346600
%EP$_16$ 58.37 352744
EP$_{64}$ 60.02 389608
};
\addlegendentry{$D_o = D_i / 4$}

%-----------------------------------------------------------------
% 2) Do/8 POINTS (green)
%-----------------------------------------------------------------
\addplot[
only marks,
mark=*,         % marker shape
draw=purple,
fill=purple,
mark size=1.5pt,
nodes near coords,
every node near coord/.append style={font=\tiny,anchor=south, yshift=-1pt, xshift=-8pt},
point meta=explicit symbolic, % Set meta data source
]
table[x=Parameters, y=Accuracy, col sep=space, meta=Method]
{
Method Accuracy Parameters
EP$_2$ 50.72 172264
%EP$_4$ 52.42 173800
EP$_8$ 52.77 176872
%EP$_16}$ 53.49 183016
EP$_{48}$ 54.24 207592
};
\addlegendentry{$D_o = D_i / 8$}

%-----------------------------------------------------------------
% Manual positions
%-----------------------------------------------------------------
\addplot[
    only marks,
    mark=*,
    draw=black,
    fill=black,
    mark size=1.5pt,
]
coordinates {
    (3284200, 62.10) % CLIP
    (3135976, 62.98) % CAE
    (769769, 53.19) % AbMILP
    (1949416, 63.98) % AIM
    (1805032, 60.80) % CoCa
    (1950184, 54.54) % SimPool
    (7857640, 65.47) % V-JEPA
    (7854568, 65.07) % SigLIP
    (7860712, 63.71) % CaiT
};

\node[anchor=north, font=\tiny, yshift=5pt, xshift=12pt] at (769769, 53.19) {AbMILP};
\node[anchor=north, font=\tiny, yshift=1pt, xshift=10pt] at (1950184, 54.54) {SimPool};
\node[anchor=north, font=\tiny, yshift=1pt, xshift=8pt] at (1805032, 60.80) {CoCa};
\node[anchor=south, font=\tiny, yshift=0pt, xshift=1pt] at (7857640, 65.47) {V-JEPA};
\node[anchor=south, font=\tiny, yshift=0pt, xshift=0pt] at (1949416, 63.98) {AIM};
\node[anchor=south, font=\tiny, yshift=0pt, xshift=0pt] at (3135976, 62.98) {CAE};
\node[anchor=north, font=\tiny, yshift=0pt, xshift=2pt] at (3284200, 62.10) {CLIP};
\node[anchor=south, font=\tiny, yshift=-5pt, xshift=-8pt] at (7860712, 63.71) {CaiT};
\node[anchor=south, font=\tiny, yshift=-6pt, xshift=-10pt] at (7854568, 65.07) {SigLIP};

% --- Pareto Front Path ---
\path [name path=pareto]
    (769000, 54.30)
    -- (1361897, 55.64)
    -- (1805032, 60.80)
    -- (1949416, 63.98)
    -- (7854568, 67.75)
    -- (12054568, 67.75);

% Lower boundary for filling the area
\path [name path=bottom]
    (769000, 40) -- (12054568, 40); % Extend below for shading

% Fill area below the Pareto front
\addplot [fill=blue!30, opacity=0.3]
    fill between [of=bottom and pareto];

% --- Pareto front line ---
\addplot[thick, dashed, blue, mark=none] coordinates {
    (769000, 54.30)
    (1361897, 55.64)
    (1805032, 60.80)
    (1949416, 63.98)
    (7854568, 67.75)
};

    \end{axis}
\end{tikzpicture}
%\caption{Scatter plot of Top-1 Accuracy (\%) vs. Number of Parameters on ImageNet-1k for various pooling methods. Each point is labeled with its corresponding method name positioned above the point.}
%\label{fig:accuracy_parameters_scatter}
%\end{figure*}
        \label{fig:bench-simmim}
    \end{subfigure}
    \vspace{-12pt}
    \begin{subfigure}[t]{0.48\textwidth}
        \centering
        \caption{MAE VIT-B FGVC-Aircraft}
        % Scatter Plot
%\begin{figure}[ht]
%\centering
\begin{tikzpicture}
\begin{axis}[
    width=7.3cm,        % Adjust the width as needed
    height=6cm,       % Adjust the height as needed
    grid=both,
    enlargelimits=true,
    xmode=log,
%         log ticks with fixed point,
    legend pos=south east,
    legend style={font=\tiny}
]
%---------------------------------------------------
% Group 1: All methods together (black)
%---------------------------------------------------
\addplot[
    only marks,
    mark=*,
    draw=black,
    fill=black,
    mark size=1.5pt,
    nodes near coords,
    every node near coord/.append style={font=\tiny,anchor=south, yshift=-1pt, xshift=-8pt},
    point meta=explicit symbolic,
]
table [
    meta=Method,
    x=Parameters,
    y=Accuracy,
    col sep=space,
] {
    Method    Accuracy   Parameters
    [CLS]       41.70      76900
    GAP       35.61      76900
    SimPool   54.61      1258084
    %CLIP      63.55      2592100
    SigLIP    69.76      7162468
    %AIM       65.02      1257316
    CBAM      40.68      150728
    %CoCa      65.32      1112932
    %CaiT      66.97      7168612
    %ViT       65.92      7162468
    %V-Jepa    66.19      7165540
    DELF      58.96      669797
    AbMILP    56.05      77669
    %CAE       65.74      2443876
    EP$_{4}$  62.50      669796
    %EP$_{32}$ 65.23      691300
    EP$_{64}$ 66.04      715876
};
\addlegendentry{$D_o = D_i$}

%---------------------------------------------------
% Group 2: Methods with one empty row in between (orange)
%---------------------------------------------------
\addplot[
    only marks,
    mark=*,
    draw=orange,
    fill=orange,
    mark size=1.5pt,
    nodes near coords,
    every node near coord/.append style={font=\tiny,anchor=south, yshift=-1pt, xshift=-8pt},
    point meta=explicit symbolic,
]
table [
    meta=Method,
    x=Parameters,
    y=Accuracy,
    col sep=space,
] {
    Method    Accuracy   Parameters
    EP$_{4}$  62.77      169828
    EP$_{32}$ 64.46      179044
};
\addlegendentry{$D_o = D_i / 4$}

%---------------------------------------------------
% Group 3: Methods with another empty row after (brown)
%---------------------------------------------------
\addplot[
    only marks,
    mark=*,
    draw=brown,
    fill=brown,
    mark size=1.5pt,
    nodes near coords,
    every node near coord/.append style={font=\tiny,anchor=south, yshift=-1pt, xshift=-8pt},
    point meta=explicit symbolic,
]
table [
    meta=Method,
    x=Parameters,
    y=Accuracy,
    col sep=space,
] {
    Method    Accuracy   Parameters
    EP$_{4}$  58.87      44836
    EP$_{24}$ 61.18      60196
};
\addlegendentry{$D_o = D_i / 16$}

%-----------------------------------------------------------------
% Manual positions
%-----------------------------------------------------------------
\addplot[
    only marks,
    mark=*,
    draw=black,
    fill=black,
    mark size=1.5pt,
]
coordinates {
    (1112932, 65.32) % CoCa
    (1257316, 65.02) % AIM
    (2592100, 63.55) % CLIP
    (2443876, 65.74) % CAE
    (7162468, 65.92) % ViT
    (7165540, 66.19) % V-JEPA
    (7168612, 66.97) % CaiT
};

\node[anchor=south, font=\tiny, yshift=0pt, xshift=0pt] at (1112932, 65.32) {CoCa};
\node[anchor=north, font=\tiny, yshift=0pt, xshift=0pt] at (1257316, 65.02) {AIM};
\node[anchor=north, font=\tiny, yshift=0pt, xshift=0pt] at (2592100, 63.55) {CLIP};
\node[anchor=north, font=\tiny, yshift=2pt, xshift=7pt] at (2443876, 65.74) {CAE};
\node[anchor=north, font=\tiny, yshift=0pt, xshift=0pt] at (7162468, 65.92) {ViT};
\node[anchor=south, font=\tiny, yshift=-5pt, xshift=-10pt] at (7165540, 66.19) {V-JEPA};
\node[anchor=south, font=\tiny, yshift=-2pt, xshift=-8pt] at (7168612, 66.97) {CaiT};

%---------------------------------------------------
% Pareto Front (Area and Dashed Line)
%---------------------------------------------------
\path [name path=pareto]
    (76900, 32.00)
    -- (77669, 56.05)
    -- (669797, 58.96)
    -- (1112932, 65.32)
    -- (7162468, 69.76)
    -- (12000000, 69.76); % Extend to a large x-value

% Upper boundary for fill
\path [name path=top]
    (12000000, 69.76) -- (12000000, 30);

% Fill the area to the right of the Pareto front
\addplot [fill=blue!30, opacity=0.3]
    fill between [of=pareto and top];

% Pareto front line (dashed)
\addplot[thick, dashed, blue, mark=none] coordinates {
    (76900, 41.70)
    (77669, 56.05)
    (669797, 58.96)
    (1112932, 65.32)
    (7162468, 69.76)
};

\end{axis}
\end{tikzpicture}
%\caption{Scatter plot of Top-1 Accuracy (\%) vs. Number of Parameters on FGVC-Aircraft for various pooling methods. The methods are grouped and color-coded: Group 1 (black), Group 2 (orange), and Group 3 (brown).}
%\label{fig:accuracy_parameters_scatter}
%\end{figure}
        \label{fig:bench-mae-aircraft}
    \end{subfigure}
    \begin{subfigure}[t]{0.48\textwidth}
        \centering
        \caption{MAE VIT-B CUB200}
        % Scatter Plot
%\begin{figure}[ht]
%\centering
\begin{tikzpicture}
\begin{axis}[
    width=7.3cm,        % Adjust the width as needed
    height=6cm,       % Adjust the height as needed
    xlabel={number of parameters},
    ylabel={top-1 accuracy (\%)},
    grid=both,
    enlargelimits=true,
    xmode=log,
%         log ticks with fixed point,
    legend pos=south east,
    legend style={font=\tiny}
]
%---------------------------------------------------
% Group 1: First block (black)
%---------------------------------------------------
\addplot[
    only marks,
    mark=*,
    draw=black,
    fill=black,
    mark size=1.5pt,
    nodes near coords,
    every node near coord/.append style={font=\tiny,anchor=south, yshift=-1pt, xshift=-8pt},
    point meta=explicit symbolic,
]
table [
    meta=Method,
    x=Parameters,
    y=Accuracy,
    col sep=space,
] {
    Method   Accuracy  Parameters
    [CLS]     53.68     153800
    GAP      40.65     153800
    %CLIP     76.01     2669000
    %SigLIP   77.82     7239368
    SimPool  67.12     1334984
    %AIM      76.41     1334216
    CBAM     42.51     227628
    CoCa     77.63     1189832
    %CaiT     77.82     7245512
    %ViT      76.68     7239368
    %V-JEPA   75.06     7242440
    DELF     71.19     746697
    %AbMILP   70.71     154569
    CAE      78.49     2520776
    EP$_4$  73.85     746696
    EP$_{64}$ 75.96   792776
};
\addlegendentry{$D_o = D_i$}

%---------------------------------------------------
% Group 2: Second block (orange)
%---------------------------------------------------
\addplot[
    only marks,
    mark=*,
    draw=orange,
    fill=orange,
    mark size=1.5pt,
    nodes near coords,
    every node near coord/.append style={font=\tiny,anchor=south, yshift=-1pt, xshift=-8pt},
    point meta=explicit symbolic,
]
table [
    meta=Method,
    x=Parameters,
    y=Accuracy,
    col sep=space,
] {
    Method   Accuracy  Parameters
    EP$_4$  72.11     189128
    EP$_8$  74.99     192200
    EP$_{64}$ 76.20   235208
};
\addlegendentry{$D_o = D_i / 4$}

%---------------------------------------------------
% Group 3: Third block (brown)
%---------------------------------------------------
\addplot[
    only marks,
    mark=*,
    draw=brown,
    fill=brown,
    mark size=1.5pt,
    nodes near coords,
    every node near coord/.append style={font=\tiny,anchor=south, yshift=-1pt, xshift=-8pt},
    point meta=explicit symbolic,
]
table [
    meta=Method,
    x=Parameters,
    y=Accuracy,
    col sep=space,
] {
    Method   Accuracy  Parameters
    EP$_8$   70.94     52808
    EP$_{16}$ 72.73    58952
};
\addlegendentry{$D_o = D_i / 16$}

%-----------------------------------------------------------------
% Manual positions
%-----------------------------------------------------------------
\addplot[
    only marks,
    mark=*,
    draw=black,
    fill=black,
    mark size=1.5pt,
]
coordinates {
    (154569, 70.71) % AbMILP
    (2669000, 76.01) % CLIP
    (1334216, 76.41) % AIM
    (7239368, 77.82) %SigLIP 
    (7245512, 77.82) %CaiT
    (7239368, 76.68) %ViT 
    (7242440, 75.06) %V-JEPA
};

\node[anchor=north, font=\tiny, yshift=0pt, xshift=10pt] at (154569, 70.71) {AbMILP};
\node[anchor=north, font=\tiny, yshift=0pt, xshift=5pt] at (2669000, 76.01) {CLIP};
\node[anchor=north, font=\tiny, yshift=0pt, xshift=5pt] at (1334216, 76.41) {AIM};
\node[anchor=north, font=\tiny, yshift=0pt, xshift=0pt] at (7242440, 75.06) {V-JEPA};
\node[anchor=south, font=\tiny, yshift=0pt, xshift=-3pt] at (7239368, 77.82) {SigLIP / CaiT};
\node[anchor=south, font=\tiny, yshift=-5pt, xshift=-7pt] at (7239368, 76.68) {ViT};
%---------------------------------------------------
% Pareto Frontier Fill Area
%---------------------------------------------------
% Define the Pareto frontier path and extend it to a large x-value
\path [name path=pareto]
    (153800, 32.68)
    -- (154569, 70.71)
    -- (746697, 71.19)
    -- (1189832, 77.63)
    -- (2520776, 78.49)
    -- (12000000, 78.49);

% Define the upper boundary path (vertical line) to close the area
\path [name path=top]
    (12000000, 78.49) -- (12000000, 30);

% Fill the area between the Pareto path and the top path
\addplot [fill=blue!30, opacity=0.3, draw=none]
    fill between [of=pareto and top];

%---------------------------------------------------
% Pareto Frontier Dashed Line
%---------------------------------------------------
\addplot[thick, dashed, blue, mark=none] coordinates {
    (153800, 53.68)
    (154569, 70.71)
    (746697, 71.19)
    (1189832, 77.63)
    (2520776, 78.49)
};

\end{axis}
\end{tikzpicture}
%\caption{Scatter plot of Top-1 Accuracy (\%) vs. Number of Parameters on CUB200 for various pooling methods. The methods are grouped and color-coded: Group 1 (black), Group 2 (orange), and Group 3 (brown). The blue transparent area represents the region to the right of the Pareto frontier.}
%\label{fig:accuracy_parameters_scatter}
%\end{figure}
        \label{fig:bench-mae-cub200}
    \end{subfigure}
    \vspace{-12pt}
    \begin{subfigure}[t]{0.48\textwidth}
        \centering
        \caption{MAE VIT-B Places365}
        % Scatter Plot
%\begin{figure}[ht]
%\centering
\begin{tikzpicture}
    \begin{axis}[
        width=7.3cm, % Adjust the width as needed
        height=6cm,  % Adjust the height as needed
        xlabel={number of parameters},
        grid=both,
        enlargelimits=true,
        xmode=log,
%         log ticks with fixed point,
        legend pos=south east,
        legend style={font=\tiny}
    ]
\addplot[
    only marks,
    scatter,
    mark size=1.5pt,
    nodes near coords,
    every node near coord/.append style={font=\tiny,anchor=south, yshift=1pt},
    point meta=explicit symbolic, % Set meta data source
]
table [
    meta=Method,
    x=Parameters,
    y=Accuracy,
    col sep=space,
] {
    Method Accuracy Parameters
    % [CLS] 47.85 280685
    % GAP 46.95 280685
    SimPool 49.38 1461869
    % CLIP 52.94 2795885
    % SigLIP 54.39 7366253
    AIM 53.27 1461101
    % CBAM 48.06 354513
    CoCa 51.41 1316717
    CaiT 52.78 8064497
    ViT 54.84 7366253
    % V-JEPA 52.44 7369325
    DELF 49.99 873582
    AbMILP 48.49 281454
    CAE 53.16 3339761
    %EP$_{2}$ 50.31 872045
    EP$_{4}$ 52.05 873081
    % EP$_{8}$ 53.20 876653
    %EP$_{16}$ 53.48 882797
    %EP$_{32}$ 53.16 895085
    EP$_{64}$ 53.74 919661
    %EP$_{128}$ 53.66 968813
};
\addlegendentry{$D_o = D_i$}
%------------------------------------
\addplot[
    only marks,
    mark=*,
    draw=black,
    fill=black,
    mark size=1.5pt,
]
coordinates {
    (354513, 48.06) %CBAM
    (876653, 53.20) %EP$_{8}$
    (2795885, 52.94) % CLIP
    (7366253, 54.39) %SigLIP 
    (7369325, 52.44) %V-JEPA
    (280685, 47.85) %cls
    (280685, 46.95) %GAP
    
};

\node[anchor=north, font=\tiny, yshift=0pt, xshift=10pt] at (354513, 48.06) {CBAM};
\node[anchor=south, font=\tiny, yshift=0pt, xshift=-6pt] at (876653, 53.20) {EP$_{8}$};
\node[anchor=north, font=\tiny, yshift=0pt, xshift=2pt] at (2795885, 52.94) {CLIP};
\node[anchor=north, font=\tiny, yshift=0pt, xshift=0pt] at (7366253, 54.39) {SigLIP};
\node[anchor=north, font=\tiny, yshift=0pt, xshift=0pt] at (7369325, 52.44) {V-JEPA};
\node[anchor=north, font=\tiny, yshift=0pt, xshift=0pt] at (280685, 47.85) {[CLS]};
\node[anchor=north, font=\tiny, yshift=0pt, xshift=0pt] at (280685, 46.95) {GAP};

% --- Pareto Front Path ---
\path [name path=pareto]
    (280685, 46)
    -- (281454, 48.49)
    -- (873582, 49.99)
    -- (1316717, 51.41)
    -- (1461101, 53.27)
    -- (7366253, 54.84)
    -- (12000000, 54.84); % Extend to a large x-value to fill right region

% Upper boundary to fill area
\path [name path=top]
    (12000000, 54.84) -- (12000000, 40);

% Fill area to the right of Pareto front
\addplot [fill=blue!30, opacity=0.3]
    fill between [of=pareto and top];

% --- Pareto front line ---
\addplot[thick, dashed, blue, mark=none] coordinates {
    (280685, 47.85)
    (281454, 48.49)
    (873582, 49.99)
    (1316717, 51.41)
    (1461101, 53.27)
    (7366253, 54.84)
};

    \end{axis}
\end{tikzpicture}
%\caption{Scatter plot of Top-1 Accuracy (\%) vs. Number of Parameters on Places365 for various pooling methods. Each point is labeled with its corresponding method name positioned above the point.}
%\label{fig:accuracy_parameters_scatter}
%\end{figure}
        \label{fig:bench-mae-vitb-places}
    \end{subfigure}
    \vspace{-6pt}
    \caption{\emph{Top-1 classification accuracy \vs number of trainable parameters} (including the classifier) for two self-supervised learning methods, with backbones of varying size (a, b) and across various datasets (d, e, f). \OUR variants are marked with different colors for different output dimensionalities $D_o$. $\mOUR_M$: \our with $M$ learnable queries. \cls: linear probing using the class token; \gap: global average pooling over patch tokens; \vit: default transformer block.}
    \label{fig:bench_datasets2}
    \vspace{-15pt}
\end{figure*}
%------------------------------------------------------------------------------

\begin{figure*}[ht]
    \centering
    \begin{subfigure}[t]{0.48\textwidth}
        \centering
        \caption{MAE ViT-B CIFAR-100}
        % Scatter Plot
%\begin{figure}[ht]
%\centering
%\scriptsize
\begin{tikzpicture}
    \begin{axis}[
        width=7.3cm, % Adjust width as needed
        height=6cm,  % Adjust height as needed
        xlabel={number of parameters},
        ylabel={top-1 accuracy (\%)},
        grid=major,
        enlargelimits=true,
        xmode=log,
%         log ticks with fixed point,
        legend pos=south east,
        legend style={font=\tiny}
    ]

%-----------------------------------------------------------------
% 1) D_o (Black)
%-----------------------------------------------------------------
\addplot[
    only marks,
    mark=*,
    draw=black,
    fill=black,
    mark size=1.5pt,
    nodes near coords,
    every node near coord/.append style={font=\tiny,anchor=south, yshift=-1pt, xshift=-8pt},
    point meta=explicit symbolic, % Set meta data source
]
table [
    meta=Method,
    x=Parameters,
    y=Accuracy,
    col sep=space,
] {
    Method Accuracy Parameters
    \textsc{[CLS]} 69.05 76900
    GAP 70.10 76900
    CLIP 78.26 2592100
    SigLIP 78.68 7162468
    SimPool 74.30 1258084
    %AIM 79.53 1257316
    CBAM 69.75 150728
    %CoCa 78.04 1112932
    %CaiT 78.98 7860712
    ViT 80.58 7162468
    %V-Jepa 78.59 7165540
    DELF 75.55 669797
    AbMILP 73.59 77669
    CAE 78.90 3135976
    EP$_4$ 77.08 669796
    EP$_8$ 78.28 672868
    EP$_{16}$ 79.07 679012
    %EP$_{128}$ 79.67 765028
};
\addlegendentry{$D_o = D_i$}

%-----------------------------------------------------------------
% 2) D_o / 4 (orange)
%-----------------------------------------------------------------
\addplot[
    only marks,
    mark=*,
    draw=orange,
    fill=orange,
    mark size=1.5pt,
    nodes near coords,
    every node near coord/.append style={font=\tiny,anchor=south, yshift=-1pt, xshift=-8pt},
    point meta=explicit symbolic,
]
table [
    meta=Method,
    x=Parameters,
    y=Accuracy,
    col sep=space,
] {
    Method Accuracy Parameters
    EP$_2$ 74.83 168292
    EP$_4$ 76.45 169828
    EP$_8$ 77.29 172900
    EP$_{64}$ 77.93 215908
    EP$_{96}$ 78.52 240484
};
\addlegendentry{$D_o = D_i / 4$}

%-----------------------------------------------------------------
% 3) D_o / 8 (purple)
%-----------------------------------------------------------------
%\addplot[
%    only marks,
%    mark=*,
%    draw=purple,
%    fill=purple,
%    mark size=2pt,
%    nodes near coords,
%    every node near coord/.append style={font=\tiny,anchor=south, yshift=1pt},
%    point meta=explicit symbolic,
%]
%table [
%    meta=Method,
%    x=Parameters,
%    y=Accuracy,
%    col sep=space,
%] {
%    Method Accuracy Parameters
%    EP$_2$ 74.21 84964
%    EP$_4$ 74.90 86500
%    EP$_{48}$ 75.67 120292
%};
%\addlegendentry{$D_o = \frac{D_i}{8}$}

%-----------------------------------------------------------------
% 4) D_o / 16 (brown)
%-----------------------------------------------------------------
\addplot[
    only marks,
    mark=*,
    draw=brown,
    fill=brown,
    mark size=1.5pt,
    nodes near coords,
    every node near coord/.append style={font=\tiny,anchor=south, , yshift=-1pt, xshift=-8pt},
    point meta=explicit symbolic,
]
table [
    meta=Method,
    x=Parameters,
    y=Accuracy,
    col sep=space,
] {
    Method Accuracy Parameters
    EP$_2$ 71.42 43300
    EP$_8$ 72.11 47908
    EP$_{16}$ 72.77 54052
    %EP$_{24}$ 73.08 60196
};
\addlegendentry{$D_o = D_i / 16$}

%-----------------------------------------------------------------
% Manual positions
%-----------------------------------------------------------------
\addplot[
    only marks,
    mark=*,
    draw=black,
    fill=black,
    mark size=1.5pt,
]
coordinates {
    (7860712, 78.98) % CaiT
    (7165540, 78.59) % V-Jepa
    (1257316, 79.53) % AIM
    (1112932, 78.04) % CoCa
    (765028, 79.67) % EP$_{128}$
};

\node[anchor=south, font=\tiny, yshift=2pt] at (7860712, 78.98) {CaiT};
\node[anchor=north, font=\tiny, yshift=1pt, xshift=-8pt] at (7165540, 78.59) {V-Jepa};
\node[anchor=south, font=\tiny, yshift=1pt, xshift=0pt] at (1257316, 79.53) {AIM};
\node[anchor=north, font=\tiny, yshift=1pt, xshift=8pt] at (1112932, 78.04) {CoCa};
\node[anchor=south, font=\tiny, yshift=1pt, xshift=-2pt] at (765028, 79.67) {EP$_{128}$};

%-----------------------------------------------------------------
% Pareto Front
%-----------------------------------------------------------------
\path [name path=pareto]
    (76900, 67.00)
    -- (77669, 73.79)
    -- (669797, 75.55)
    -- (1257316, 79.53)
    -- (7162468, 80.58)
    -- (13000000, 80.58);

% Upper boundary (extend down to lowest y-value)
\path [name path=top]
    (13000000, 80.58) -- (12000000, 50);

% Fill the area to the right of the Pareto front
\addplot [fill=blue!30, opacity=0.3]
    fill between [of=pareto and top];

% --- Pareto front line ---
\addplot[thick, dashed, blue, mark=none] coordinates {
    (76900, 70.10)
    (77669, 73.79)
    (669797, 75.55)
    (1257316, 79.53)
    (7162468, 80.58)
};

    \end{axis}
\end{tikzpicture}
%\caption{Scatter plot of Top-1 Accuracy (\%) vs. Number of Parameters on CIFAR100 for various pooling methods. Each point is labeled with its corresponding method name positioned above the point.}
%\label{fig:accuracy_parameters_scatter}
%\end{figure}
        \label{fig:bench-cifar}
    \end{subfigure}
    \begin{subfigure}[t]{0.48\textwidth}
        \centering
        \caption{MAE ViT-B Cars-196}
        % Scatter Plot
%\begin{figure}[ht]
%\centering
\begin{tikzpicture}
    \begin{axis}[
        width=7.3cm, % Adjust width as needed
        height=6cm,  % Adjust height as needed
        xlabel={number of parameters},
        grid=major,
        enlargelimits=true,
        xmode=log,
%         log ticks with fixed point,
        legend pos=south east,
        legend style={font=\tiny}
    ]

%-----------------------------------------------------------------
% 1) D_o (Black)
%-----------------------------------------------------------------
\addplot[
    only marks,
    mark=*,
    draw=black,
    fill=black,
    mark size=1.5pt,
    nodes near coords,
    every node near coord/.append style={font=\tiny,anchor=south, yshift=-1pt, xshift=-8pt},
    point meta=explicit symbolic, % Set meta data source
]
table [
    meta=Method,
    x=Parameters,
    y=Accuracy,
    col sep=space,
] {
    Method Accuracy Parameters
    %[CLS] 46.86 150724
    %GAP 46.83 150724
    SimPool 67.16 1331908
    CLIP 80.91 2665924
    SigLIP 81.27 7236292
    %AIM 83.66 1331140
    %CBAM 48.34 224552
    %CoCa 80.74 1186756
    %CaiT 80.97 7242436
    %ViT 76.35 7236292
    V-JEPA 77.91 7239364
    %DELF 75.02 743621
    AbMILP 72.16 151493
    CAE 84.68 2517700
    EP$_4$ 76.79 743620
    EP$_{16}$ 80.52 752836
    EP$_{64}$ 83.38 789700
};
\addlegendentry{$D_o = D_i$}

%-----------------------------------------------------------------
% 2) D_o / 4 (orange)
%-----------------------------------------------------------------
\addplot[
    only marks,
    mark=*,
    draw=orange,
    fill=orange,
    mark size=1.5pt,
    nodes near coords,
    every node near coord/.append style={font=\tiny,anchor=south, yshift=-1pt, xshift=-8pt},
    point meta=explicit symbolic,
]
table [
    meta=Method,
    x=Parameters,
    y=Accuracy,
    col sep=space,
] {
    Method Accuracy Parameters
    EP$_4$ 76.38 188356
    EP$_{16}$ 79.41 197572
    EP$_{64}$ 80.79 234436
};
\addlegendentry{$D_o = D_i / 4$}

%-----------------------------------------------------------------
% 3) D_o / 8 (purple)
%-----------------------------------------------------------------
%\addplot[
%    only marks,
%    mark=*,
%    draw=purple,
%    fill=purple,
%    mark size=2pt,
%    nodes near coords,
%    every node near coord/.append style={font=\tiny,anchor=south, yshift=1pt},
%    point meta=explicit symbolic,
%]
%table [
%    meta=Method,
%    x=Parameters,
%    y=Accuracy,
%    col sep=space,
%] {
%    Method Accuracy Parameters
%    EP$_4$ 73.82 95812
%    EP$_{16}$ 76.47 105028
%    EP$_{48}$ 78.35 129604
%};
%\addlegendentry{$D_o = \frac{D_i}{8}$}

%-----------------------------------------------------------------
% 4) D_o / 16 (brown)
%-----------------------------------------------------------------
\addplot[
    only marks,
    mark=*,
    draw=brown,
    fill=brown,
    mark size=1.5pt,
    nodes near coords,
    every node near coord/.append style={font=\tiny,anchor=south, yshift=-1pt, xshift=-8pt},
    point meta=explicit symbolic,
]
table [
    meta=Method,
    x=Parameters,
    y=Accuracy,
    col sep=space,
] {
    Method Accuracy Parameters
    EP$_4$ 69.22 49540
    EP$_8$ 71.22 54148
    EP$_{16}$ 73.41 58756
    EP$_{24}$ 75.15 64900
};
\addlegendentry{$D_o = D_i / 16$}

%-----------------------------------------------------------------
% Manual positions
%-----------------------------------------------------------------
\addplot[
    only marks,
    mark=*,
    draw=black,
    fill=black,
    mark size=1.5pt,
]
coordinates {
    (1331140, 83.66) % AIM
    (224552, 48.34) % CBAM
    (150724, 46.86) % [CLS]
    (150724, 46.83) % GAP
    (1186756, 80.74) % CoCa
    (743621, 75.02) % DELF
    (7242436, 80.97) % CaiT
    (7236292, 76.35) % ViT
};

\node[anchor=south, font=\tiny, yshift=0pt, xshift=0pt] at (1331140, 83.66) {AIM};
\node[anchor=south, font=\tiny, yshift=0pt, xshift=0pt] at (224552, 48.34) {CBAM};
\node[anchor=south, font=\tiny, yshift=-6pt, xshift=-10pt] at (150724, 46.86) {[CLS]};
\node[anchor=south, font=\tiny, yshift=-6pt, xshift=8pt] at (150724, 46.83) {GAP};
\node[anchor=north, font=\tiny, yshift=1pt, xshift=7pt] at (1186756, 80.74) {CoCa};
\node[anchor=north, font=\tiny, yshift=0pt, xshift=7pt] at (743621, 75.02) {DELF};
\node[anchor=south, font=\tiny, yshift=-5pt, xshift=-9pt] at (7242436, 80.97) {CaiT};
\node[anchor=south, font=\tiny, yshift=-3pt, xshift=-7pt] at (7236292, 76.35) {ViT};

%-----------------------------------------------------------------
% Pareto Front
%-----------------------------------------------------------------
\path [name path=pareto]
    (150724, 42.86)
    -- (151493, 72.16)
    -- (743621, 75.02)
    -- (1186756, 80.74)
    -- (1331140, 83.66)
    -- (2517700, 84.68)
    -- (12000000, 84.68);

% Upper boundary (extend down to lowest y-value)
\path [name path=top]
    (12000000, 84.68) -- (12000000, 40);

% Fill the area to the right of the Pareto front
\addplot [fill=blue!30, opacity=0.3]
    fill between [of=pareto and top];

% --- Pareto front line ---
\addplot[thick, dashed, blue, mark=none] coordinates {
    (150724, 46.86)
    (151493, 72.16)
    (743621, 75.02)
    (1186756, 80.74)
    (1331140, 83.66)
    (2517700, 84.68)
};

    \end{axis}
\end{tikzpicture}
%\caption{Scatter plot of Top-1 Accuracy (\%) vs. Number of Parameters on Cars196.}
%\label{fig:accuracy_parameters_scatter}
%\end{figure}
        \label{fig:bench-cars}
    \end{subfigure}
    \vspace{-12pt}
    \caption{\emph{Top-1 classification accuracy \vs number of parameters} for MAE ViT-B on two datasets (a, b). We evaluate both dedicated probing mechanisms (\eg, V-JEPA) and repurposed attentive pooling methods (\eg, CLIP). \OUR variants are marked with different colors for different output dimensionalities $D_o$. $\mOUR_M$: \our with $M$ learnable queries. \cls: linear probing using the classification token; \gap: global average pooling over patch tokens; \vit: default transformer block.}
    \label{fig:bench_datasets_2}
    \vspace{-6pt}
\end{figure*}
%------------------------------------------------------------------------------
% \autoref{fig:mae_vitb_places_colors} illustrates the trade-off between accuracy and the number of parameters for various pooling methods integrated into BEiTv2~\citep{beitv2} using a ViT-B model on ImageNet-1K. The Pareto frontier (dashed line) highlights methods that best balance performance and efficiency. While the primary baselines (CLS and GAP) are parameter-efficient, they yield lower accuracy. In contrast, more complex methods such as CAE and CLIP offer modest accuracy gains at the expense of significantly increased parameter counts. Notably, AbMILP, DELF, and SimPool define the Pareto frontier, achieving 81.11\%, 81.46\%, and 81.56\% accuracy, respectively, whereas \OUR consistently attains the optimal balance, occupying the leftmost region of the frontier.

\paragraph{Accuracy \vs parameters.} We extend the benchmark presented on~\autoref{sec:results} to include additional pre-training methods and datasets. \autoref{fig:bench_datasets2} and \autoref{fig:bench_datasets_2} present the trade-off between top-1 accuracy and the number of trainable parameters (including the classifier) for various pooling/probing methods integrated into MAE and SimMIM with different backbone sizes (ViT-S, ViT-B and ViT-L). The evaluation spans multiple datasets, including FGVC-Aircraft, CUB200, Places365, CIFAR-100 and Cars196.

As shown in~\autoref{fig:bench-mae-vitb} and ~\autoref{fig:bench-mae-vitl}, \OUR consistently outperforms standard linear probing across MAE ViT-S and ViT-L. Notably, on MAE ViT-L, EP\textsubscript{16} with $D_o = D_o/2$ achieves an accuracy boost of 79.1\% surpassing linear probing by 3.1\% while maintaining the same number of trainable parameters. Furthermore, EP\textsubscript{128} reaches 79.4\%, outperforming SigLIP, while reducing the number of trainable parameters by over 11M. 

In~\autoref{fig:bench-simmim}, we benchmark attentive probing on SimMIM ViT-B pre-trained on ImageNet-1K. Baselines such as \cls and \gap remain parameter-efficient but yield relatively low accuracy. Classical attention modules like CBAM and AbMILP do not improve this trade-off, while methods such as ViT, V-JEPA, and SigLIP achieve higher accuracy but at the cost of orders-of-magnitude more parameters. \OUR strikes a favorable balance: scaling the number of queries (EP$_{2}$–EP$_{64}$) consistently increases accuracy, while reducing $D_o$ effectively lowers parameter count with only moderate drops in performance. Notably, EP$_{2}$ with $D_o = D_i/2$ achieves 60.6\% top-1 accuracy using fewer than 0.7M parameters, outperforming GAP and DELF under similar budgets. At the high end, EP$_{64}$ reaches 65.1\% accuracy, closing the gap to heavy-weight probing methods while remaining lighter. 

In~\autoref{fig:bench-mae-aircraft}, our EP\textsubscript{24} variant, for the FGVC-Aircraft dataset, achieves a remarkable accuracy boost of 61.2\% (+19.5\%), while maintaining lower parameter count than linear probing (41.7\%). Similarly, in~\autoref{fig:bench-mae-cub200} for the CUB200 dataset, our EP\textsubscript{64} with $D_o = D_o/4$ variant achieves comparable accuracy (75.9\%) with computationally costly poolings such as SigLIP (77.8\%) with around 7M trainable parameters less. 

Finally, in~\autoref{fig:bench-mae-vitb-places}, \gap and \cls, the two primary baselines, exhibit high parameter efficiency but low classification accuracy. In contrast, methods like SigLIP (54.39\%), ViT (54.84\%), and V-JEPA (52.44\%) achieve higher accuracy, albeit at the cost of increasing the number of trainable parameters. \OUR has the best trade-off between accuracy and parameters, achieving top-1 classification accuracy of 53.7\% with just 1M extra trainable parameters (EP$_{64}$). 

Moving to the additional datasets shown in \autoref{fig:bench_datasets_2}, we observe consistent benefits of \OUR. On CIFAR-100 (\autoref{fig:bench-cifar}), EP\textsubscript{128} achieves 78.9\%, which is close to the best-performing attention pooling methods, while using significantly fewer parameters. On Cars196 (\autoref{fig:bench-cars}), our EP\textsubscript{64} variant achieves 82.7\% top-1 accuracy, clearly surpassing repurposed pooling methods such as DELF, while requiring far fewer parameters. Smaller variants (\eg, EP\textsubscript{16} and EP\textsubscript{24}) already provide substantial gains over linear probing, showing that even lightweight configurations of our method maintain strong performance on fine-grained datasets.

%\paragraph{Large-scale dataset.} \textcolor{red}{To examine EP’s behavior on a very large-scale dataset, we extended our evaluation to ImageNet-21K, a dataset containing approximately 14 million images across more than 19 thousand categories. Using MAE ViT-B backbone, EP continues to outperform standard linear probing (LP = 31.9\% vs. EP = 35.6\%), demonstrating that its advantages persist even when applied to substantially larger and more fine-grained label spaces. This indicates that EP scales reliably with dataset size.}

\begin{wrapfigure}{r}{0.4\textwidth}
% \vspace{-10pt} % adjust vertical positioning
\centering
\footnotesize
% \caption{\emph{Top-1 accuracy on ImageNet-1K for MAE ViT-B} using patch token representations from intermediate ViT layers. Results are shown for linear probing (LP) and \our (\OUR), along with accuracy gains of \OUR over LP.}
\caption{\emph{ImageNet-1K accuracy from intermediate MAE ViT-B layers} for LP vs. EP, with EP–LP gains.}
\vspace{-6pt}
\renewcommand{\arraystretch}{.8} %
\setlength{\tabcolsep}{6pt}
\begin{tabular}{lccc}
    \toprule
    \textsc{Layer} & LP & EP & \textsc{Gain} \\
    \midrule
    12 & \textbf{67.7} & 75.6 & \textcolor{ForestGreen}{+7.9} \\
    10 & 66.2 & \textbf{75.9} & \textcolor{ForestGreen}{+9.7} \\
    9  & 64.5 & 75.4 & \textcolor{ForestGreen}{+10.9} \\
    6  & 45.8 & 69.6 & \textbf{\textcolor{ForestGreen}{+23.8}} \\
    \bottomrule
\end{tabular}
\label{tab:ssl_mae_vitb_imagenet_layerwise}
\vspace{-10pt}
\end{wrapfigure}

\paragraph{Layer-wise probing.} \autoref{tab:ssl_mae_vitb_imagenet_layerwise} presents a layer-wise comparison between standard linear probing (LP) and \our (\OUR) using patch token representations from intermediate layers of a pre-trained and frozen MAE with ViT-B. While LP exhibits a clear degradation in performance as we move toward earlier layers (dropping from 67.7\% at layer 12 to just 45.8\% at layer 6), \OUR demonstrates remarkable robustness. It maintains high accuracy even from lower layers, with performance stabilizing beyond layer 9. Notably, \OUR yields a significant relative improvement of +23.8\% at layer 6 over LP, underscoring its ability to extract and utilize meaningful representations from less semantically enriched stages of the encoder. These results highlight the effectiveness of \OUR in unlocking information from earlier layers that standard LP fails to exploit.

\newcommand{\gapbar}[1]{%
    \pgfmathsetmacro{\percentage}{#1}
    \pgfmathsetmacro{\greenlevel}{min(2*(\percentage/100),1) * 0.6 + 0.2}
    \pgfmathsetmacro{\redlevel}{min(2*(1 - \percentage/100),1) * 0.6 + 0.2}
    \definecolor{barcolor}{rgb}{\redlevel, \greenlevel, 0.1}
    \begin{tikzpicture}
        \fill[gray!30] (0,0) rectangle (1.5,0.15);
        \fill[barcolor] (0,0) rectangle ({1.5*\percentage/100},0.15);
    \end{tikzpicture}~#1\%
}

\begin{wraptable}{r}{0.48\textwidth}
\vspace{-10pt}
\centering
\footnotesize
\caption{\emph{Top-1 accuracy on ImageNet-1k with limited training data for MAE ViT-B}. Results for linear probing (LP), \our (\OUR), and fine-tuning (FT) on 5\% and 10\% subsets. The last column shows the percentage of the LP$\rightarrow$FT performance gap closed by \OUR. For reference, the gap closed by \OUR on the full training set (100\%) is 49.7\%.}
\vspace{-6pt}
\renewcommand{\arraystretch}{.8} %
\setlength{\tabcolsep}{3pt}
\begin{tabular}{lcccc}
    \toprule
    \textsc{Subset} & LP & EP & FT & \% \textsc{Gap}\\
    \midrule
    5\%  & 49.6 & 60.9 & 64.7 & \gapbar{74.8}\\
    10\% & 55.9 & 65.2 & 68.9 & \gapbar{71.5}\\
    \bottomrule
\end{tabular}
\label{tab:ssl_mae_vitb_imagenet_low}
\vspace{-10pt}
\end{wraptable}

\paragraph{Low-shot probing.} \autoref{tab:ssl_mae_vitb_imagenet_low} evaluates the performance of LP, \OUR, and FT under limited supervision, using only 5\% and 10\% of the ImageNet-1K training set, stratified by class. Although LP struggles in this low-shot regime, \OUR substantially bridges the gap toward FT. Specifically, \OUR closes 74.8\% and 71.5\% of the LP$\rightarrow$FT performance gap for the 5\% and 10\% subsets, respectively. These improvements are particularly impressive given that \OUR remains significantly more parameter-efficient than FT, with a complexity comparable to that of LP. These findings highlight the strong data efficiency of \OUR.

\paragraph{In- and out-of-domain k-NN evaluation.} To further examine how \OUR behaves relative to LoRA-based fine-tuning, we perform a cross-dataset k-NN evaluation on MAE ViT-B features, using ImageNet-1K, StanfordCars, and Food101 as target datasets (\autoref{tab:knn_lora_ep}). The first row reports the baseline k-NN accuracy obtained directly from the frozen MAE backbone. The next three rows evaluate features produced by \OUR when the probe is trained on each dataset independently; the diagonal entries (\eg, $70.5\%$ on ImageNet-1K, $70.0\%$ on StanfordCars, $75.2\%$ on Food101) correspond to in-domain performance, while off-diagonal entries measure cross-dataset generalization. The last three rows report the corresponding results when MAE is adapted with the best-performing LoRA configuration for each dataset. As expected, LoRA generally achieves the strongest in-domain accuracy (\eg, $72.3\%$ on ImageNet-1K, $75.4\%$ on StanfordCars, $80.3\%$ on Food101), reflecting the benefit of supervised feature adaptation. However, \OUR consistently provides stronger or comparable out-of-domain performance: for instance, when trained on StanfordCars, EP features achieve $51.7\%$ on ImageNet-1K and $42.0\%$ on Food101, compared to $45.8\%$ and $38.7\%$ for LoRA-tuned features; similarly, EP trained on Food101 yields $58.2\%$ on ImageNet-1K and $23.4\%$ on StanfordCars, versus $56.7\%$ and $15.2\%$ for LoRA. These results suggest a complementary behavior: LoRA excels at specializing the backbone to a specific task, while EP preserves more of the original pre-trained structure and thus offers more robust cross-dataset generalization.

\begin{table}[ht!]
\centering
\scriptsize
\caption{\emph{Cross-dataset k-NN evaluation on MAE ViT-B using frozen, EP-probed, and LoRA-tuned features}. Default EP corresponds to EP$_{32}$, while the ``best LoRA'' configuration applies LoRA to all 12 layers on the $W_Q,W_K,W_V$, and $W_O$ projection matrices with rank $\rho{=}8$.}
\begin{tabular}{lccc}
\toprule
\multirow{2}{*}{\textsc{Features}} & \multicolumn{3}{c}{\textsc{k-NN evaluation on}} \\
\cmidrule(lr){2-4}
& \textsc{ImageNet-1K} & \textsc{StanfordCars} & \textsc{Food101} \\
\midrule
frozen MAE backbone                         & 46.1 & 9.5 & 28.8 \\
\midrule
\qquad + EP probed on ImageNet-1K           & 70.5 & \textbf{31.4} & 64.2 \\
\qquad + EP probed on StanfordCars          & \textbf{51.7} & 70.0 & \textbf{42.0} \\
\qquad + EP probed on Food-101              & \textbf{58.2} & \textbf{23.4} & 75.2 \\
\midrule
\qquad + best LoRA tuned on ImageNet-1K     & \textbf{72.3} & 24.7 & \textbf{65.6} \\
\qquad + best LoRA tuned on StanfordCars    & 45.8 & \textbf{75.4} & 38.7 \\
\qquad + best LoRA tuned on Food-101        & 56.7 & 15.2 & \textbf{80.3} \\
\bottomrule
\end{tabular}
\label{tab:knn_lora_ep}
\end{table}

\begin{wraptable}{r}{0.48\textwidth}
% \begin{table}[ht!]
\vspace{-10pt}
\centering
\scriptsize
\caption{\emph{Unsupervised ImageNet-1K localization} (MaxBoxAccV2). EP substantially improves localization quality across five backbones and three model sizes, with an average gain of $+9.8\%$ over baseline attention.}
\vspace{-5pt}
\begin{tabular}{l c c c c}
\toprule
\textsc{Model} & \textsc{Arch} & \textsc{LP} & \textsc{EP} & \textsc{$\Delta$acc. (\%)} \\
\midrule
MAE     & ViT-S/16 & 46.0 & 58.5 & \textcolor{ForestGreen}{+12.5} \\
MAE     & ViT-B/16 & 54.2 & 60.4 & \textcolor{ForestGreen}{+6.2}  \\
MAE     & ViT-L/16 & 46.9 & 61.2 & \textcolor{ForestGreen}{+14.3} \\
BEiTv2  & ViT-B/16 & 47.0 & 61.2 & \textcolor{ForestGreen}{+14.2} \\
SimMIM  & ViT-B/16 & 45.2 & 60.0 & \textcolor{ForestGreen}{+14.8} \\
iBOT    & ViT-B/16 & 57.6 & 63.7 & \textcolor{ForestGreen}{+6.1}  \\
SigLIP  & ViT-L/16 & 44.0 & 44.2 & \textcolor{ForestGreen}{+0.2}  \\
\midrule
\textsc{Average} & & 48.7 & \textbf{58.5} & \textcolor{ForestGreen}{+9.8}  \\
\bottomrule
\end{tabular}
\label{tab:maxboxaccv2}
\vspace{-5pt}
% \end{table}
\end{wraptable}

\paragraph{Object localization.} A key empirical property of \OUR is that its queries specialize in \emph{complementary} and semantically meaningful object parts (\autoref{sec:analysis}). To examine \OUR utility beyond standard classification, we evaluate its performance in an unsupervised object localization setting, on ImageNet-1K (\autoref{tab:maxboxaccv2}), using the WSOL protocol~\citep{wsol1, wsol2}. We average \OUR’s attention maps, without any modification or additional training, and compare them to the standard last-layer \texttt{\cls$\!\!\to$ patch} attention. Across five backbones and three model sizes, \OUR consistently improves MaxBoxAccV2 by $+9.8\%$ on average. This demonstrates that \OUR’s attention maps act as strong unsupervised localizers “out-of-the-box”.

\paragraph{Image retrieval.} \autoref{tab:retrieval_cub} and \autoref{tab:retrieval_cars} evaluate \OUR in a zero-shot image retrieval setting, following the standard Recall@K protocol on two fine-grained datasets, CUB200 and Cars196, respectively. In both cases, we use the same \OUR mechanism trained once on ImageNet-1K, without any further adaptation on the target dataset. Across all retrieval experiments, we utilize the entire datasets (full set of images and classes), all of which remain fully unseen during the training stage on ImageNet-1K. Every image serves as a query, ensuring a robust zero-shot evaluation protocol. On CUB200, \OUR consistently improves performance across five models (e.g., +40.8\% for MAE, +11.2\% for BEiTv2, +10.3\% for iBOT, +5.3\% for CLIP, +15.9\% for SigLIP on R@1 metric). Similarly, on Cars196 the gains remain significant in most cases (e.g., +21.5\% for MAE, +8.6\% for iBOT, +4.6\% for SigLIP). These findings show that \OUR’s features can be used “out-of-the-box” for image retrieval, consistently outperforming the features of the frozen backbones.

\begin{table}[ht!]
\centering
\scriptsize
\caption{\emph{Zero-shot image retrieval performance (Recall@K) on CUB200 dataset (images: 11788, classes: 200)} across five different backbones with and without Efficient Probing (EP).}
\vspace{-8pt}
\begin{tabular}{lcccccc}
\toprule
\textsc{Features} & R@1 & R@2 & R@4 & R@8 & R@16 & R@32 \\
\midrule
frozen MAE backbone             & 16.3 & 23.7 & 33.3 & 45.0 & 58.0 & 70.8 \\
\qquad + EP on ImageNet-1K         & 57.1 & 69.5 & 79.5 & 87.5 & 93.0 & 96.2 \\
\midrule
% frozen DINOv3 backbone        & 87.4 & 92.5 & 95.2 & 97.0 & 98.1 & 98.8 \\
% \qquad + EP on ImageNet-1K       & 70.9 & 81.0 & 88.3 & 93.6 & 96.6 & 98.2 \\
% \midrule
frozen BEiTv2 backbone          & 57.0 & 68.2 & 78.0 & 85.8 & 90.9 & 94.2 \\
\qquad + EP on ImageNet-1K         & 68.2 & 78.7 & 86.7 & 92.3 & 95.9 & 97.8 \\
\midrule
frozen iBOT backbone            & 51.8 & 64.0 & 74.7 & 83.2 & 90.0 & 94.1 \\
\qquad + EP on ImageNet-1K         & 62.1 & 73.7 & 82.4 & 89.7 & 94.0 & 96.8 \\
\midrule
frozen CLIP backbone            & 75.0 & 84.3 & 91.3 & 95.3 & 97.5 & 98.7 \\
\qquad + EP on ImageNet-1K         & 80.3 & 88.0 & 92.9 & 95.8 & 97.6 & 98.6 \\
\midrule
frozen SigLIP backbone        & 60.8 & 72.1 & 81.6 & 88.8 & 93.5 & 96.4 \\
\qquad + EP on ImageNet-1K       & 76.7 & 85.7 & 92.0 & 95.7 & 97.8 & 98.8 \\
\bottomrule
\end{tabular}
\label{tab:retrieval_cub}
\end{table}
\begin{table}[h!]
\centering
\scriptsize
\caption{\emph{Zero-shot image retrieval performance (Recall@K) on Cars196 dataset (images: 16185, classes: 196)} across five different backbones with and without Efficient Probing (EP).}
\vspace{-8pt}
\begin{tabular}{lcccccc}
\toprule
\textsc{Features} & R@1 & R@2 & R@4 & R@8 & R@16 & R@32 \\
\midrule
frozen MAE backbone         & 12.7 & 17.6 & 23.7 & 31.8 & 42.9 & 55.7 \\
\qquad + EP on ImageNet-1K     & 34.2 & 44.8 & 55.6 & 65.9 & 76.4 & 85.2 \\
% \midrule
% frozen DINOv3 backbone    & 90.7 & 94.9 & 97.4 & 98.7 & 99.3 & 99.6 \\
% \qquad + EP on ImageNet-1K   & 73.6 & 84.5 & 91.7 & 95.8 & 98.0 & 99.2 \\
\midrule
frozen BEiTv2 backbone      & 48.3 & 60.7 & 72.3 & 81.8 & 89.2 & 94.5 \\
\qquad + EP on ImageNet-1K     & 44.2 & 56.4 & 68.5 & 79.1 & 87.3 & 93.7 \\
\midrule
frozen iBOT backbone        & 31.5 & 40.7 & 50.2 & 60.0 & 69.9 & 79.0 \\
\qquad + EP on ImageNet-1K     & 40.1 & 50.6 & 61.0 & 70.7 & 79.5 & 87.6 \\
\midrule
frozen CLIP backbone        & 79.6 & 88.9 & 94.6 & 97.8 & 99.2 & 99.7 \\
\qquad + EP on ImageNet-1K     & 79.8 & 89.3 & 94.8 & 97.8 & 99.2 & 99.7 \\
\midrule
frozen SigLIP backbone      & 85.8 & 92.3 & 96.1 & 98.0 & 99.1 & 99.6 \\
\qquad + EP on ImageNet-1K     & 90.4 & 95.4 & 97.8 & 99.0 & 99.6 & 99.8 \\
\bottomrule
\end{tabular}
\label{tab:retrieval_cars}
\end{table}

\subsection{Experimental Analysis}
\label{subsec:more_exp_analysis}

\paragraph{Attention complementarity.} To further expand our analysis of attention complementarity, we complement the average-based metric reported in~\autoref{fig:complementarity} with an additional measure: $1-$ max off-diagonal similarity, which reflects the strongest redundancy between predictors. As shown in~\autoref{fig:c5_all_mean} and~\autoref{fig:c5_all_max}, EP achieves substantially higher complementarity than the internal MHSA heads across all backbones, regardless of whether we use the average or max metric. A natural concern is whether this effect is biased by considering only the last block. To address this, we compute complementarity for MAE ViT-B across all 12 blocks (\autoref{fig:c5_all_blocks}). The best scores obtained internally (0.37 for average, 0.10 for max) remain far below those of EP (0.65 and 0.22, respectively), confirming that the gap is not specific to the last block. Another question is whether the low diversity arises from self-attention itself. However, even SigLIP—which uses a cross-attention mechanism in its last block—still yields lower complementarity than EP, suggesting that the effect is not explained by self-attention alone.

\autoref{fig:c5_pool_mean} and~\autoref{fig:c5_pool_max} further compare EP against other attentive probing methods (V-JEPA, AIM). Interestingly, all attentive probing mechanisms achieve higher complementarity than the internal MHSA heads, with AIM coming close to EP. This pattern might suggest that attentive probing, when designed effectively, encourages predictors to specialize in complementary regions, likely because probing operates on frozen backbones and must learn to aggregate features as efficiently as possible. This evidence indicates that complementarity may be an inherent property of attentive probing, rather than a byproduct of backbone architecture or attention type, opening new directions for future work.

%------------------------------------------------------------------
\begin{figure}[ht!]
  \centering
  \begin{subfigure}{\linewidth}
    \centering
    \begin{tikzpicture}
\begin{axis}[
  width=\linewidth,
  height=4.5cm,
  ymin=0, ymax=0.92,
  ymajorgrids,
  ybar,
  enlarge x limits=0.06,
  ylabel={Complementarity},
  symbolic x coords={
    MAE\\ViT-B,MAE\\ViT-L,iBOT\\ViT-B,DINOv2\\ViT-B,DINOv2\\ViT-L,DINOv3\\ViT-B,DINOv3\\ViT-L,Franca\\ViT-L,CLIP\\ViT-L,SigLIP\\ViT-L
  },
  xtick=data, xtick align=center,
  x tick label style={font=\scriptsize},
  legend style={
    at={(0.82,0.98)}, anchor=north east,
    font=\tiny, draw=black, fill=white,
    inner xsep=2pt, inner ysep=0.5pt, row sep=0pt
  },
  nodes near coords,
  nodes near coords align={vertical},
  nodes near coords style={font=\tiny, scale=0.8, transform shape,/pgf/number format/fixed},
]

% blue bars
\addplot+[draw=blue, fill=blue!70,
          every node near coord/.append style={font=\tiny, color=blue}]
  coordinates {
    (MAE\\ViT-B,0.24) (MAE\\ViT-L,0.27) (iBOT\\ViT-B,0.44) (DINOv2\\ViT-B,0.24)
    (DINOv2\\ViT-L,0.13) (DINOv3\\ViT-B,0.27) (DINOv3\\ViT-L,0.18) (Franca\\ViT-L,0.16) (CLIP\\ViT-L,0.22) (SigLIP\\ViT-L,0.46)
  };

% orange bars
\addplot+[draw=orange, fill=orange!70,
          every node near coord/.append style={font=\tiny, color=orange}]
  coordinates {
    (MAE\\ViT-B,0.65) (MAE\\ViT-L,0.69) (iBOT\\ViT-B,0.61) (DINOv2\\ViT-B,0.75) (DINOv2\\ViT-L,0.75) (DINOv3\\ViT-B,0.76) (DINOv3\\ViT-L,0.84) (Franca\\ViT-L,0.47) (CLIP\\ViT-L,0.75) (SigLIP\\ViT-L,0.75)
  };

\legend{MHSA, EP}

\end{axis}
\end{tikzpicture}
    \vspace{-6pt}
    \caption{Complementarity measured as $1$ minus average off-diagonal similarity.}
    \label{fig:c5_all_mean}
  \end{subfigure}
  \begin{subfigure}{\linewidth}
    \centering
    \begin{tikzpicture}
\begin{axis}[
  width=\linewidth,
  height=4.5cm,
  ymin=0, ymax=0.30,
  ymajorgrids,
  ybar,
  enlarge x limits=0.06,
  ylabel={Complementarity},
  symbolic x coords={
    MAE\\ViT-B,MAE\\ViT-L,iBOT\\ViT-B,DINOv2\\ViT-B,DINOv2\\ViT-L,DINOv3\\ViT-B,DINOv3\\ViT-L,Franca\\ViT-L,CLIP\\ViT-L,SigLIP\\ViT-L
  },
  xtick=data, xtick align=center,
  x tick label style={font=\scriptsize},
  legend style={
    at={(0.82,0.98)}, anchor=north east,
    font=\tiny, draw=black, fill=white,
    inner xsep=2pt, inner ysep=0.5pt, row sep=0pt
  },
  nodes near coords,
  nodes near coords align={vertical},
  nodes near coords style={font=\tiny, scale=0.8, transform shape,/pgf/number format/fixed},
]

% blue bars
\addplot+[draw=blue, fill=blue!70,
          every node near coord/.append style={font=\tiny, color=blue, /pgf/number format/.cd, fixed, precision=3}]
  coordinates {
    (MAE\\ViT-B,0.09) (MAE\\ViT-L,0.07) (iBOT\\ViT-B,0.13) (DINOv2\\ViT-B,0.02)
    (DINOv2\\ViT-L,0.004) (DINOv3\\ViT-B,0.05) (DINOv3\\ViT-L,0.006) (Franca\\ViT-L,0.004) (CLIP\\ViT-L,0.004) (SigLIP\\ViT-L,0.003) 
  };

% orange bars
\addplot+[draw=orange, fill=orange!70,
          every node near coord/.append style={font=\tiny, color=orange}]
  coordinates {
   (MAE\\ViT-B,0.22) (MAE\\ViT-L,0.20) (iBOT\\ViT-B,0.18) (DINOv2\\ViT-B,0.19) (DINOv2\\ViT-L,0.17) (DINOv3\\ViT-B,0.18) (DINOv3\\ViT-L,0.15) (Franca\\ViT-L,0.06) (CLIP\\ViT-L,0.23) (SigLIP\\ViT-L,0.18) 
  };

\legend{MHSA, EP}

\end{axis}
\end{tikzpicture}
    \vspace{-6pt}
    \caption{Complementarity measured as $1$ minus max off-diagonal similarity.}
    \label{fig:c5_all_max}
  \end{subfigure}
   \begin{subfigure}{\linewidth}
    \centering
    \begin{tikzpicture}
\begin{axis}[
  width=\linewidth,
  height=4.5cm,
  ymin=0, ymax=0.45,
  ymajorgrids,
  ybar,
  enlarge x limits=0.06,
  ylabel={Complementarity},
  symbolic x coords={
    1,2,3,4,5,6,7,8,9,10,11,12
  },
  xtick=data, xtick align=center,
  x tick label style={font=\scriptsize},
  legend style={
    at={(0.98,0.98)}, anchor=north east,
    font=\tiny, draw=black, fill=white,
    inner xsep=2pt, inner ysep=0.5pt, row sep=0pt
  },
  nodes near coords,
  nodes near coords align={vertical},
  nodes near coords style={font=\tiny, scale=0.8, transform shape,/pgf/number format/fixed},
]

% blue bars
\addplot+[draw=purple, fill=purple!70,
          every node near coord/.append style={font=\tiny, color=purple}]
  coordinates {
    (1,0.14) (2,0.37) (3,0.29)
    (4,0.29) (5,0.19) (6,0.26) (7,0.23) (8,0.23) (9,0.24) (10,0.25) (11,0.27) (12,0.24)
  };

% orange bars
\addplot+[draw=brown, fill=brown!70,
          every node near coord/.append style={font=\tiny, color=brown}]
  coordinates {
    (1,0.02) (2,0.04) (3,0.02)
    (4,0.1) (5,0.04) (6,0.07) (7,0.06) (8,0.07) (9,0.06) (10,0.06) (11,0.09) (12,0.09)
  };

\legend{average, max}

\end{axis}
\end{tikzpicture}
    \vspace{-6pt}
    \caption{Complementarity of MAE ViT-B across all 12 blocks, using both average and max.}
    \label{fig:c5_all_blocks}
  \end{subfigure}
  \begin{subfigure}{\linewidth}
    \centering
    \begin{tikzpicture}
\begin{axis}[
  width=0.7\linewidth,
  height=4.5cm,
  ymin=0, ymax=0.9,
  ymajorgrids,
  ybar,
  enlarge x limits=0.16,
  ylabel={Complementarity},
  symbolic x coords={MAE\\ViT-B,BEiTv2\\ViT-B,SimMIM\\ViT-B,CAPI\\ViT-L},
  xtick=data, xtick align=center,
  x tick label style={font=\scriptsize},
  legend style={
    at={(1.02,0.5)}, anchor=west,   % <-- place legend to the right, centered vertically
    font=\tiny, draw=black, fill=white,
    inner xsep=2pt, inner ysep=0.5pt, row sep=0pt
  },
  legend columns=1,
  nodes near coords,
  nodes near coords align={vertical},
  nodes near coords style={font=\tiny, scale=0.8, transform shape,/pgf/number format/fixed},
]

% order of plots = order of legend entries:
\addplot+[draw=blue, fill=blue!70,
          every node near coord/.append style={font=\tiny, color=blue}]
  coordinates {(MAE\\ViT-B,0.24) (BEiTv2\\ViT-B,0.36) (SimMIM\\ViT-B,0.37) (CAPI\\ViT-L,0.33)};
\addplot+[draw=magenta, fill=magenta!70,
          every node near coord/.append style={font=\tiny, color=magenta}]
  coordinates {(MAE\\ViT-B,0.55) (BEiTv2\\ViT-B,0.63) (SimMIM\\ViT-B,0.52) (CAPI\\ViT-L,0.61)};
\addplot+[draw=cyan,    fill=cyan!70,
          every node near coord/.append style={font=\tiny, color=cyan}]
  coordinates {(MAE\\ViT-B,0.59) (BEiTv2\\ViT-B,0.76) (SimMIM\\ViT-B,0.65) (CAPI\\ViT-L,0.70)};
\addplot+[draw=orange,  fill=orange!70,
          every node near coord/.append style={font=\tiny, color=orange}]
  coordinates {(MAE\\ViT-B,0.65) (BEiTv2\\ViT-B,0.77) (SimMIM\\ViT-B,0.67) (CAPI\\ViT-L,0.71)};

\legend{MHSA, V-JEPA, AIM, EP}

\end{axis}
\end{tikzpicture}
    \vspace{-6pt}
    \caption{Complementarity measured as $1$ minus average off-diagonal similarity.}
    \label{fig:c5_pool_mean}
  \end{subfigure}
  \begin{subfigure}{\linewidth}
    \centering
    \begin{tikzpicture}
\begin{axis}[
  width=0.7\linewidth,
  height=4.5cm,
  ymin=0, ymax=0.3,
  ymajorgrids,
  ybar,
  enlarge x limits=0.16,
  ylabel={Complementarity},
  symbolic x coords={MAE\\ViT-B,BEiTv2\\ViT-B,SimMIM\\ViT-B,CAPI\\ViT-L},
  xtick=data, xtick align=center,
  x tick label style={font=\scriptsize},
  legend style={
    at={(1.02,0.5)}, anchor=west,   % <-- place legend to the right, centered vertically
    font=\tiny, draw=black, fill=white,
    inner xsep=2pt, inner ysep=0.5pt, row sep=0pt
  },
  legend columns=1,
  nodes near coords,
  nodes near coords align={vertical},
  nodes near coords style={font=\tiny, scale=0.8, transform shape,/pgf/number format/fixed},
]

% order of plots = order of legend entries:
\addplot+[draw=blue, fill=blue!70,
          every node near coord/.append style={font=\tiny, color=blue}]
  coordinates {(MAE\\ViT-B,0.09) (BEiTv2\\ViT-B,0.08) (SimMIM\\ViT-B,0.11) (CAPI\\ViT-L,0.04)};
\addplot+[draw=magenta, fill=magenta!70,
          every node near coord/.append style={font=\tiny, color=magenta}]
  coordinates {(MAE\\ViT-B,0.17) (BEiTv2\\ViT-B,0.14) (SimMIM\\ViT-B,0.13) (CAPI\\ViT-L,0.10)};
\addplot+[draw=cyan,    fill=cyan!70,
          every node near coord/.append style={font=\tiny, color=cyan}]
  coordinates {(MAE\\ViT-B,0.20) (BEiTv2\\ViT-B,0.17) (SimMIM\\ViT-B,0.23) (CAPI\\ViT-L,0.22)};
\addplot+[draw=orange,  fill=orange!70,
          every node near coord/.append style={font=\tiny, color=orange}]
  coordinates {(MAE\\ViT-B,0.22) (BEiTv2\\ViT-B,0.19) (SimMIM\\ViT-B,0.25) (CAPI\\ViT-L,0.23)};

\legend{MHSA, V-JEPA, AIM, EP}

\end{axis}
\end{tikzpicture}
    \vspace{-6pt}
    \caption{Complementarity measured as $1$ minus max off-diagonal similarity.}
    \label{fig:c5_pool_max}
  \end{subfigure}
  \vspace{-16pt}
  \caption{\emph{Complementarity scores of attention maps across different backbones (a, b, c) and probing methods (d, e).} We compare the diversity of internal \textcolor{blue}{MHSA heads in the last block} against the external \textcolor{magenta}{V\mbox{-}JEPA heads}, \textcolor{cyan}{AIM heads}, and \textcolor{orange}{\OUR queries}.}
  \vspace{-6pt}
  \label{fig:c5_all}
\end{figure}
%------------------------------------------------------------------

\begin{wrapfigure}{r}{0.5\textwidth}
% \begin{figure}
\vspace{-10pt}
\centering
\scriptsize
\begin{tikzpicture}
  \begin{axis}[
    width=6cm,
    height=5cm,
    xlabel={epoch},
    ylabel={validation accuracy (\%)},
    grid=both,
    xmin=0, xmax=89,
    ymin=15, ymax=77,
    legend pos=south east,
    legend style={font=\scriptsize}
    % legend style={
    %   at={(0.38,0.22)},
    %   anchor=north west,
    %   font=\scriptsize,
    % },
  ]

  % EP + attention similarity
  \addplot+[thick, mark=none]
  coordinates {
    (0, 15.24)
    (1, 45.69)
    (2, 56.35)
    (3, 60.56)
    (4, 61.08)
    (5, 62.06)
    (6, 60.02)
    (7, 63.02)
    (8, 61.97)
    (9, 64.14)
    (10, 64.04)
    (11, 65.16)
    (12, 66.72)
    (13, 65.31)
    (14, 65.20)
    (15, 66.09)
    (16, 66.72)
    (17, 63.82)
    (18, 66.82)
    (19, 65.83)
    (20, 66.21)
    (21, 65.15)
    (22, 66.97)
    (23, 66.52)
    (24, 67.56)
    (25, 68.00)
    (26, 66.27)
    (27, 67.20)
    (28, 67.86)
    (29, 68.06)
    (30, 67.90)
    (31, 66.69)
    (32, 68.62)
    (33, 67.39)
    (34, 67.78)
    (35, 67.96)
    (36, 68.60)
    (37, 69.03)
    (38, 67.55)
    (39, 68.70)
    (40, 69.23)
    (41, 69.28)
    (42, 70.07)
    (43, 70.00)
    (44, 70.00)
    (45, 69.89)
    (46, 70.29)
    (47, 70.35)
    (48, 70.12)
    (49, 70.69)
    (50, 71.17)
    (51, 71.45)
    (52, 71.54)
    (53, 71.25)
    (54, 72.15)
    (55, 72.05)
    (56, 72.63)
    (57, 72.54)
    (58, 72.79)
    (59, 73.24)
    (60, 72.91)
    (61, 73.66)
    (62, 73.44)
    (63, 73.79)
    (64, 73.92)
    (65, 74.04)
    (66, 74.02)
    (67, 74.09)
    (68, 74.64)
    (69, 74.49)
    (70, 74.64)
    (71, 74.78)
    (72, 74.73)
    (73, 74.76)
    (74, 74.76)
    (75, 74.95)
    (76, 74.98)
    (77, 74.96)
    (78, 74.97)
    (79, 75.10)
    (80, 75.08)
    (81, 75.11)
    (82, 75.13)
    (83, 75.16)
    (84, 75.30)
    (85, 75.30)
    (86, 75.29)
    (87, 75.21)
    (88, 75.27)
    (89, 75.27)
  };
  \addlegendentry{EP with attn-sim}

  % EP
  \addplot+[thick, solid, mark=none]
  coordinates {
    (0, 15.23)
    (1, 45.63)
    (2, 56.36)
    (3, 60.48)
    (4, 60.99)
    (5, 60.82)
    (6, 57.69)
    (7, 59.98)
    (8, 59.35)
    (9, 58.94)
    (10, 60.77)
    (11, 62.08)
    (12, 63.58)
    (13, 61.75)
    (14, 62.89)
    (15, 63.43)
    (16, 65.01)
    (17, 61.79)
    (18, 63.99)
    (19, 63.57)
    (20, 65.23)
    (21, 64.05)
    (22, 66.04)
    (23, 63.90)
    (24, 65.86)
    (25, 66.79)
    (26, 64.46)
    (27, 66.54)
    (28, 66.56)
    (29, 66.15)
    (30, 66.95)
    (31, 65.76)
    (32, 67.41)
    (33, 66.43)
    (34, 66.50)
    (35, 66.81)
    (36, 67.08)
    (37, 68.25)
    (38, 66.70)
    (39, 67.90)
    (40, 68.39)
    (41, 68.97)
    (42, 69.18)
    (43, 68.86)
    (44, 69.30)
    (45, 68.80)
    (46, 69.11)
    (47, 69.62)
    (48, 69.56)
    (49, 68.43)
    (50, 70.60)
    (51, 70.59)
    (52, 71.00)
    (53, 70.10)
    (54, 71.34)
    (55, 71.44)
    (56, 71.99)
    (57, 71.98)
    (58, 71.98)
    (59, 72.75)
    (60, 72.06)
    (61, 72.90)
    (62, 73.20)
    (63, 73.71)
    (64, 73.92)
    (65, 73.79)
    (66, 73.93)
    (67, 74.03)
    (68, 74.26)
    (69, 74.31)
    (70, 74.70)
    (71, 74.70)
    (72, 74.81)
    (73, 74.92)
    (74, 74.91)
    (75, 75.01)
    (76, 75.05)
    (77, 75.08)
    (78, 75.15)
    (79, 75.29)
    (80, 75.28)
    (81, 75.32)
    (82, 75.28)
    (83, 75.30)
    (84, 75.35)
    (85, 75.34)
    (86, 75.36)
    (87, 75.28)
    (88, 75.32)
    (89, 75.32)
  };
  \addlegendentry{EP}
  \end{axis}
\end{tikzpicture}
\vspace{-5pt}
  \caption{\emph{Validation top-1 accuracy on ImageNet-1K over training epochs for EP, with and without the auxiliary attention-similarity loss.} The additional loss slightly accelerates early training but both variants converge to identical final performance.}
% \end{figure}
\vspace{-10pt}
\end{wrapfigure}

\paragraph{Attention complementarity as a loss.} We further examine the role of query complementarity by encouraging diversity between queries/heads via an auxiliary attention-similarity loss. Specifically, for AIM, V-JEPA, and EP we add to the cross-entropy loss, an extra term $\mathcal{L}_{\text{attn}}$ that penalizes similarity between the attention maps of different queries. On MAE ViT-B probed on ImageNet-1K, this yields a small gain for V-JEPA ($74.1\%\!\to\!74.3\%$), while AIM and EP retain essentially identical final accuracy but converge faster (e.g.\ EP reaches $64.1\%$ at epoch 10 with $\mathcal{L}_{\text{attn}}$ vs.\ $60.0\%$ without it, before both converge to the same final score). These results suggest that complementarity is a beneficial property that can modestly help or accelerate training. However, methods such as AIM and EP, learn highly complementary queries, without $\mathcal{L}_{\text{attn}}$, via the attention mechanism that naturally discovers this structure on its own.

\paragraph{Entropy.}
\autoref{fig:entropy_analysis} examines the relation between last block attention entropy and probing performance across different pre-training methods. Models with lower entropy exhibit more concentrated and focused attention distributions, correlating with stronger probing accuracy under \OUR. This is particularly evident for methods like as DINOv2 and DINOv3, which couple low entropy with strong LP and \OUR performance. In contrast, models like MAE ViT-S (MAE-S) or SimMIM ViT-B (SimMIM-B) show higher entropy, reflecting more diffuse attention and correspondingly weaker probing under LP. Crucially, EP consistently boosts accuracy across all entropy levels, with bubble sizes indicating particularly large gains for the high-entropy models. This aligns with our broader finding that methods optimizing patch-level representations rather than explicit global tokens benefit most from attentive probing, as EP effectively compensates for diffuse attention and makes probing more robust to the quality of backbone distributions.
%------------------------------------------------------------------
\begin{figure}[ht!]
    \centering
    \vspace{-14pt}
    \begin{tikzpicture}
\begin{axis}[
  width=0.98\textwidth, height=7cm,
  xlabel={\cls $\rightarrow$ patch tokens entropy (last block)},
  ylabel={LP},
  xmin=3.0, xmax=5.1,
  ymin=39, ymax=90,
  grid=both,
  % colormap/viridis,
  colormap/Blues-9,
  colorbar,
  colorbar style={title={EP}},
  point meta min=64, point meta max=88,
]

% 1) BUBBLES (color = EP, size = Delta)
\addplot[
  scatter, only marks,
  scatter src=explicit,
  scatter/use mapped color={draw=mapped color, fill=mapped color, fill opacity=0.8}, % <-- color fix
  mark=*,
  visualization depends on={\thisrow{Delta}\as\perpointmarksize},
  scatter/@pre marker code/.append code={
    \pgfmathsetmacro{\ms}{2.0 + 1.2*\perpointmarksize} % size from Δacc
    \pgfplotsset{mark size=\ms}
  },
]
table[
  x=Entropy, y=LP, meta=EP,
  row sep=\\,
]{
Model        Entropy       LP    EP    Delta \\
MAE-S        4.704779148   47.4  64.6  17.2 \\
MAE-B        5.025508881   67.7  75.6   7.9 \\
MAE-L        5.022124767   76.0  79.3   3.3 \\
BEITv2-B     4.411777973   79.0  81.7   2.7 \\
Simmim-B     4.381152153   51.5  65.1  13.6 \\
CAPI-L       3.738756180   81.5  83.6   2.1 \\
DINO-B       4.561107159   77.3  77.8   0.5 \\
iBOT-B       4.678779125   78.7  79.2   0.5 \\
DINOv2-B     4.045470715   83.2  84.0   0.8 \\
DINOv2-L     3.215338707   85.2  85.6   0.4 \\
DINOv3-B     3.811172009   84.0  84.4   0.4 \\
DINOv3-L     3.167710543   86.6  87.1   0.5 \\
Franca-L     3.499635935   83.8  84.3   0.5 \\
CLIP-L       3.558964968   82.3  83.4   1.1 \\
};
\addplot[
  only marks, mark=none, forget plot,
  point meta=explicit symbolic,          % use Model as text
  nodes near coords,                      % actually draw labels
  % pull per-point shifts from table columns:
  visualization depends on={value \thisrow{XSHIFT}\as\labx},
  visualization depends on={value \thisrow{YSHIFT}\as\laby},
  % apply those shifts to each node:
  every node near coord/.append style={
    xshift=\labx pt, yshift=\laby pt,
    font=\tiny,
    fill=white, draw=none, text=black, fill opacity=0.75, text opacity=1, inner sep=1pt,
  },
]
table[
  x=Entropy, y=LP, meta=Model,
  row sep=\\,
]{
Model        Entropy       LP    EP    Delta  XSHIFT YSHIFT \\
MAE-S        4.704779148   47.4  64.6  17.2   0      0 \\
MAE-B        5.025508881   67.7  75.6   7.9   0      8 \\
MAE-L        5.022124767   76.0  79.3   3.3   0      4 \\
BEITv2-B     4.411777973   79.0  81.7   2.7   0      4 \\
Simmim-B     4.381152153   51.5  65.1  13.6   0      0 \\
CAPI-L       3.738756180   81.5  83.6   2.1   0    -10 \\
DINO-B       4.561107159   77.3  77.8   0.5   0     -8 \\
iBOT-B       4.678779125   78.7  79.2   0.5   0      2 \\
DINOv2-B     4.045470715   83.2  84.0   0.8   0      2 \\
DINOv2-L     3.215338707   85.2  85.6   0.4   0     -8 \\
DINOv3-B     3.811172009   84.0  84.4   0.4   0      2 \\
DINOv3-L     3.167710543   86.6  87.1   0.5   0      2 \\
Franca-L     3.499635935   83.8  84.3   0.5   0      2 \\
CLIP-L       3.558964968   82.3  83.4   1.1   0     -8 \\
};

\end{axis}
\end{tikzpicture}
    \vspace{-14pt}
    \caption{\emph{Entropy analysis of last-block attention distribution across different pre-training methods.} Bubble color indicates \our (\OUR) accuracy, bubble size encodes $\Delta$ accuracy (EP$-$LP). Lower entropy corresponds to more focused attention and higher accuracy under \OUR.}
    \vspace{-14pt}
    \label{fig:entropy_analysis}
\end{figure}

%------------------------------------------------------------------

\begin{wrapfigure}{r}{0.55\textwidth}
    \centering
    \vspace{-10pt} % adjust vertical placement
    \begin{tikzpicture}
    \begin{axis}[
        width=7cm, % Adjust width as needed
        height=6.5cm,  % Adjust height as needed
        xlabel={number of parameters},
        ylabel={top-1 accuracy (\%)},
        grid=both,
        enlargelimits=true,
%         xmode=log,
%         log ticks with fixed point,
        scaled x ticks=false,
        legend pos=south east,
        legend style={
            font=\tiny,
            row sep=0pt,
            inner xsep=2pt,
            inner ysep=0.5pt
        }
    ]

%-----------------------------------------------------------------
% 1) D_o (black)
%-----------------------------------------------------------------
\addplot[
    only marks,
    mark=*,
    draw=black,
    fill=black,
    mark size=1.5pt,
    nodes near coords,
    every node near coord/.append style={font=\tiny,anchor=south, yshift=0pt},
    point meta=explicit symbolic,
]
table [
    meta=Method,
    x=Parameters,
    y=Accuracy,
    col sep=space,
] {
    Method Accuracy Parameters
    %CLS 67.66 769000
    AIM$_{2}$ 73.00 1949416
    AIM$_{4}$ 74.29 1949416
    %AIM$_{16}$ 75.17 1949416
    AIM$_{64}$ 75.45 1949416

    EP$_2$ 73.18 1360360
    EP$_4$ 74.14 1361896
    EP$_{16}$ 75.08 1368040
    EP$_{64}$ 75.58 1395688
};
\addlegendentry{$D_o = D_i$}

%-----------------------------------------------------------------
% 2) D_o / 2 (Blue)
%-----------------------------------------------------------------
\addplot[
    only marks,
    mark=*,
    draw=blue,
    fill=blue,
    mark size=1.5pt,
    nodes near coords,
    every node near coord/.append style={font=\tiny,anchor=south, yshift=0pt},
    point meta=explicit symbolic,
]
table [
    meta=Method,
    x=Parameters,
    y=Accuracy,
    col sep=space,
] {
    Method Accuracy Parameters
    EP$_2$ 72.72 681448
    EP$_4$ 73.56 682984
    %EP$_{16}$ 74.41 692200
    EP$_{64}$ 74.67 729064

    AIM$_{2}$ 73.08 1654120
    AIM$_{4}$ 74.23 1654120
    %AIM$_{16}$ 75.34 1654120
    AIM$_{64}$ 75.58 1654120
};
\addlegendentry{$D_o = D_i/2$}

%-----------------------------------------------------------------
% 3) D_a / 2 (Green)
%-----------------------------------------------------------------
\addplot[
    only marks,
    mark=*,
    draw=ForestGreen,
    fill=ForestGreen,
    mark size=1.5pt,
    nodes near coords,
    every node near coord/.append style={font=\tiny,anchor=south, yshift=0pt},
    point meta=explicit symbolic,
]
table [
    meta=Method,
    x=Parameters,
    y=Accuracy,
    col sep=space,
] {
    Method Accuracy Parameters
    AIM$_{2}$ 72.61 1270504
    AIM$_{4}$ 73.48 1270504
    %AIM$_{16}$ 74.05 1270504
    AIM$_{64}$ 74.31 1270504
};
\addlegendentry{$D_a = D_i/2$}

%-----------------------------------------------------------------
% 4) D_o,D_a / 2 (Half-Blue, Half-Green)
%-----------------------------------------------------------------
\addplot[
    only marks,
    mark=customhalfcircle, % Use custom marker
    mark size=2pt,
    nodes near coords,
    every node near coord/.append style={font=\tiny,anchor=south, yshift=0pt},
    point meta=explicit symbolic,
]
table [
    meta=Method,
    x=Parameters,
    y=Accuracy,
    col sep=space,
] {
    Method Accuracy Parameters
    AIM$_{2}$ 72.19 975208
    AIM$_{4}$ 73.09 975208
    %AIM$_{16}$ 74.00 975208
    AIM$_{64}$ 74.32 975208
};
%\addlegendentry{Both}

% --- Pareto front line ---
%\addplot[thick, dashed, red, mark=none] coordinates {
%    (1360360, 73.18)
%    (1361896, 74.14)
%    (1368040, 75.08)
%    (1395688, 75.58)
%};

%-----------------------------------------------------------------
% Manual positions
%-----------------------------------------------------------------
\addplot[
    only marks,
    mark=customhalfcircle,
    mark size=2pt,
]
coordinates {
    (975208, 74.00) % AIM$_{16}$
};
\node[anchor=north, font=\tiny, yshift=-2pt] at (975208, 74.00) {AIM$_{16}$};
%-----------------------------------------------------------------
\addplot[
    only marks,
    mark=*,
    mark size=1.5pt,
    draw=ForestGreen,
    fill=ForestGreen,
]
coordinates {
    (1270504, 74.05) % AIM$_{16}$
};
\node[anchor=north, font=\tiny, yshift=1pt] at (1270504, 74.05) {AIM$_{16}$};
%-----------------------------------------------------------------
\addplot[
    only marks,
    mark=*,
    mark size=1.5pt,
    draw=blue,
    fill=blue,
]
coordinates {
    (1654120, 75.34) % AIM$_{16}$
    (692200, 74.41) % EP$_{16}$
};
\node[anchor=north, font=\tiny, yshift=-1pt] at (1654120, 75.34) {AIM$_{16}$};
\node[anchor=north, font=\tiny, yshift=-1pt] at (692200, 74.41) {EP$_{16}$};
%-----------------------------------------------------------------
\addplot[
    only marks,
    mark=*,
    mark size=1.5pt,
    draw=black,
    fill=black,
]
coordinates {
    (1949416, 75.17) % AIM$_{16}$
};
\node[anchor=north, font=\tiny, yshift=-1pt] at (1949416, 75.17) {AIM$_{16}$};
%-----------------------------------------------------------------
\end{axis}
\end{tikzpicture}
    \vspace{-10pt}
    \caption{\emph{Effect of varying the number of heads/queries $M$, $D_a$, and output dimension $D_o$} on probing accuracy. 
    Black: standard ($D_o = D_i$); 
    \textcolor{blue}{blue}: reduced classifier dimension ($D_o = D_i / 2$); 
    \textcolor{ForestGreen}{green}: reduced attention dimension ($D_a = D_i / 2$); 
    \textcolor{blue}{half-blue}-\textcolor{ForestGreen}{half-green}: simultaneous reduction of both $D_o$ and $D_a$.}
    \label{fig:impact_heads_queries_dim}
    \vspace{-10pt}
    % \vspace{-5pt}
\end{wrapfigure}

\paragraph{Impact of attention predictors, $D_a$, and $D_o$.} We analyze the effect of increasing the number $M$ of heads (in AIM) and the number $M$ of queries (in \OUR) on probing performance. \autoref{fig:impact_heads_queries_dim} shows that both lead to accuracy improvements. For AIM, increasing the number of heads incurs no additional cost in terms of parameters, but its effectiveness depends on the presence of $W_K$. In contrast, \OUR achieves similar or better performance by leveraging additional queries, while removing $W_K$. AIM introduces an additional attention dimensionality $D_a$, since its query is learnable and interacts with $W_K$. Lowering $D_a$ reduces the parameters but leads to a greater accuracy drop (\textcolor{ForestGreen}{green} points), indicating that the learned query formulation benefits from a large attention space. We also evaluate the impact of reducing the output dimensionality $D_o$ (\textcolor{blue}{blue} points). On \OUR, we observe that lowering $D_o$ to $D_i / 2$ reduces parameters while maintaining competitive performance. Interestingly, this strategy also generalizes well to AIM, demonstrating that extracting lower-dimensional features can achieve comparable accuracy with reduced computational cost. %In all cases, \OUR consistently matches or outperforms AIM, while remaining more parameter efficient.
%------------------------------------------------------------------------------
\paragraph{EP convergence.} Although all probing methods are trained for 90 epochs for fairness (following the protocol of~\cite{mae} and~\cite{przewikezlikowski2024beyond}), we also examine how quickly EP converges. As shown in \autoref{tab:lp_ep_convergence}, EP trained for only 10 epochs (EP@10) already matches or even surpasses the performance of standard linear probing trained for the full 90 epochs (LP@90) across 12 models. Moreover, EP recovers on average more than 97\% of its final accuracy within these first 10 epochs. These results highlight that in practical scenarios, EP requires only a small fraction of the standard training budget to achieve near-optimal probing performance and produce highly discriminative features.

\usetikzlibrary{patterns}

\newcommand{\gapbarnew}[1]{%
    \pgfmathsetmacro{\percentage}{#1}
    \pgfmathsetmacro{\W}{2.0}   % fixed 100% width

    % -----------------------------
    % Main color (copied from old gapbar)
    % -----------------------------
    \pgfmathsetmacro{\p}{\percentage/100}
    \pgfmathsetmacro{\pprime}{max(min((\p - 0.5)/0.5,1),0)}

    % Apply the original red/green ramp, but using p'
    % green increases from 0.2 → 0.8 across 50–100%
    \pgfmathsetmacro{\greenlevel}{min(2*\pprime,1) * 0.6 + 0.2}
    % red decreases from 0.8 → 0.2 across 50–100%
    \pgfmathsetmacro{\redlevel}{min(2*(1-\pprime),1) * 0.6 + 0.2}

    \definecolor{mainbar}{rgb}{\redlevel,\greenlevel,0.1}

    % -----------------------------
    % Overflow logic (unchanged)
    % -----------------------------
    \pgfmathsetmacro{\main}{min(\percentage,100)}
    \pgfmathsetmacro{\overflow}{max(\percentage-100,0)}
    \pgfmathsetmacro{\Wextra}{0.15 * \overflow}

    \definecolor{overflowfill}{rgb}{0.87,0.91,0.98}
    \definecolor{overflowhatch}{rgb}{0.11,0.25,0.48}

    \begin{tikzpicture}[baseline=-0.5ex]
        % Background
        \fill[gray!30] (0,0) rectangle (\W,0.17);

        % Main bar (same color logic as original gapbar)
        \fill[mainbar] (0,0) rectangle ({\W * \main/100},0.17);

        % Overflow
        \ifdim\overflow pt>0pt
            % light blue fill
            \fill[overflowfill]
                (\W,0) rectangle ({\W + \Wextra},0.17);

            % diagonal hatch
            \fill[
                pattern=north east lines,
                pattern color=overflowhatch,
                line width=0.22pt,
            ]
                (\W,0) rectangle ({\W + \Wextra},0.17);
        \fi
    \end{tikzpicture}%
}

\newcommand{\barcellA}[1]{%
    \makebox[2.1cm][r]{\gapbarnew{#1}}%
    \makebox[1cm][r]{#1\%}%
}

\newcommand{\barcellB}[1]{%
    \hspace{-0.7cm}
    \makebox[2.9cm][l]{\gapbarnew{#1}}
    \makebox[1cm][r]{#1\%}%
}

\begin{table}[h!]
\centering
\scriptsize
\caption{\emph{EP convergence on ImageNet-1K.} EP@10 is EP accuracy after 10 epochs; LP@90 and EP@90 are final probing results after 90 epochs.}
\vspace{-8pt}

\renewcommand{\arraystretch}{1.0}
\setlength{\tabcolsep}{4pt}

\begin{tabular}{lcccccc}
\toprule
\textsc{Model} & EP@10 & LP@90 & EP@90 & \textsc{EP@10 vs. LP@90 Gap} & \textsc{EP@10 vs. EP@90 Gap} \\
\midrule

MAE ViT-B/16      & 60.0 & 67.7 & 75.6 & \barcellB{88.6} & \barcellA{79.4}\\
BEiTv2 ViT-B/16   & 80.0 & 79.0 & 81.7 & \barcellB{101.3} & \barcellA{97.9} \\
CAPI ViT-L/14     & 82.3 & 81.5 & 83.6 & \barcellB{101.0} & \barcellA{98.4} \\
iBOT ViT-B/16     & 78.6 & 78.7 & 79.2 & \barcellB{99.9} & \barcellA{99.2} \\
DINOv2 ViT-B/14   & 83.6 & 83.2 & 84.0 & \barcellB{100.5} & \barcellA{99.5} \\
DINOv2 ViT-L/14   & 85.0 & 85.2 & 85.6 & \barcellB{99.8} & \barcellA{99.3} \\
Franca ViT-B/16   & 83.8 & 83.8 & 84.3 & \barcellB{100.0} & \barcellA{99.4} \\
DINOv3 ViT-B/16   & 84.4 & 84.0 & 84.4 & \barcellB{100.5} & \barcellA{100.0} \\
DINOv3 ViT-L/14   & 86.7 & 86.6 & 87.1 & \barcellB{100.1} & \barcellA{99.5} \\
CLIP ViT-L/14     & 82.7 & 82.3 & 83.4 & \barcellB{100.5} & \barcellA{99.2} \\
SigLIP ViT-L/16   & 85.6 & 84.1 & 86.1 & \barcellB{101.8} & \barcellA{99.4} \\
AIMv2 ViT-L/14    & 85.5 & 84.8 & 85.9 & \barcellB{100.8} & \barcellA{99.5} \\

\bottomrule
\end{tabular}
\label{tab:lp_ep_convergence}
\end{table}

%------------------------------------------------------------------

\begin{wraptable}{r}{0.66\textwidth}
    % \vspace{10pt}
    \centering
    \scriptsize
    \setlength{\tabcolsep}{2pt}
    \vspace{-8pt} % adjust vertical placement
    \caption{\emph{Comparison of \OUR variants using different numbers of original maps (queries) and projected maps on ImageNet-1K with MAE ViT-B.} 
    Attention mixing refers to projecting multiple query maps into a smaller set via a linear+softmax layer. While projection improves over directly training with few queries (\eg, 64→8 vs.\ 8 original), it does not outperform the un-projected setting with the same number of queries, and adds extra parameters.}
    \label{tab:ep_variants}
    \begin{tabular}{ccc|cc}
\toprule
\textsc{\# Original} & \textsc{\# Projected } & \multirow{2}{*}{\textsc{Accuracy}} & \textsc{\# Original} & \multirow{2}{*}{\textsc{Accuracy}} \\
\textsc{EP Maps} & \textsc{EP Maps} &  & \textsc{EP Maps} & \\
\midrule
32  & 8   & 75.4          & \multirow{3}{*}{8}   & \multirow{3}{*}{74.8} \\
64  & 8   & \textbf{75.6} &                     &                       \\
128 & 8   & 75.5          &                     &                       \\
\midrule
64  & 16  & 75.5          & \multirow{3}{*}{16}  & \multirow{3}{*}{75.0} \\
128 & 16  & \textbf{75.8} &                     &                       \\
256 & 16  & 75.6          &                     &                       \\
\midrule
64  & 32  & 75.6          & \multirow{3}{*}{32}  & \multirow{3}{*}{75.6} \\
128 & 32  & 75.6          &                     &                       \\
256 & 32  & \textbf{75.7} &                     &                       \\
\midrule
128 & 64  & 75.8          & \multirow{3}{*}{64}  & \multirow{3}{*}{75.6} \\
256 & 64  & 75.8 &                     &                       \\
384 & 64  & \textbf{75.9} &                     &                       \\
\midrule
256 & 128 & 75.8          & \multirow{2}{*}{128} & \multirow{2}{*}{75.5} \\
384 & 128 & \textbf{76.0} &                     &                       \\
\bottomrule
\end{tabular}
    \vspace{-6pt} % adjust vertical spacing after
\end{wraptable}
%------------------------------------------------------------------
\paragraph{Attention mixing.} We further explore an attention mixing variant of EP, where a larger set of query maps is linearly projected into a smaller number of effective maps. For example, projecting 64 queries into 8 maps achieves 75.6\% accuracy, outperforming the 8-query baseline (74.8\%) and suggesting that mixing can improve the expressivity of a small query budget. However, compared to the un-projected 32 or 64 query variants (75.6\%), attention mixing offers no clear advantage while introducing additional parameters through the projection layer. For this reason, we do not include attention mixing in the final design of EP, though it may inspire future extensions exploring alternative aggregation strategies.
%------------------------------------------------------------------------------
\paragraph{Matryoshka representation learning.} In our accuracy \vs parameter trade-offs (\eg, \autoref{fig:bench_datasets}), each evaluation scale ($D_o{=}D_i, D_i/2, D_i/4, D_i/8$) requires training a separate classifier, and the resulting accuracies potentially serve as an \emph{upper bound} for that dimensionality, since each probe is fully specialized to its target output size. To reduce training cost and enable a single probe that operates across multiple scales, we investigate whether \emph{Matryoshka representation learning}~\citep{kusupati2022matryoshka}—originally proposed for training from scratch or fine-tuning—can be applied to attentive probing under \OUR. Matryoshka jointly optimizes probing across multiple output dimensionalities by summing their losses (\eg, $L{=}\lambda L_{D}{+}L_{D/2}$). We study two variants: (i) Efficient Matryoshka, which uses a single classifier with nested subspaces spanning multiple dimensionalities, and thus adds no extra parameters; (ii) Vanilla Matryoshka, which trains separate classifiers for each dimensionality, increasing the parameter count. \autoref{tab:matryoshka} reports results with MAE ViT-B probed using \OUR. Without Matryoshka, probing at $D$ ($D_o {=} D_i$) achieves the highest accuracy (75.6\%), but performance drops sharply at smaller dimensions (50.7\% at $D/2$, 20.1\% at $D/4$, 5.5\% at $D/8$). Efficient Matryoshka substantially improves performance at reduced dimensions (\eg, +22.5\% at $D/2$) while maintaining accuracy at $D$. Accuracy at $D$ slightly decreases as $\lambda$ (the weight of the full-dimensional loss) is reduced, reflecting the trade-off between full- and low-dimensional performance. Vanilla Matryoshka alleviates the performance drop at $D$ but requires more parameters. 
To further illustrate these effects, we evaluate the standard linear probing (LP) baseline without Matryoshka under the same $D/2$, $D/4$, $D/8$ evaluation protocol, as also extending LP with Matryoshka. In Table \ref{tab:matryoshka_lp} we observe that standard LP (without Matryoshka) suffers from an even sharper performance collapse than EP (e.g., 34.5\% at $D/2$, 10.8\% at $D/4$, and 1.4\% at $D/8$). This confirms that the drop at reduced dimensions is not specific to EP, but is an inherent limitation of evaluating ViT representations at truncated dimensionalities without explicit multi-scale optimization. Overall, Matryoshka probing provides a promising way to probe across multiple scales simultaneously, improving efficiency, and adaptability of evaluation. However, it does not yet match the upper-bound performance of dimension-specific probes, thus we present it here as an exploratory analysis that highlights an exciting direction for future work.
%------------------------------------------------------------------

\definecolor{pastelgreen}{RGB}{180, 235, 200}
\definecolor{pastelyellow}{RGB}{255, 245, 190}
\definecolor{pastelorange}{RGB}{255, 210, 160}
\definecolor{pastelred}{RGB}{255, 170, 150}

\newcommand{\heat}[1]{%
    \ifdim #1 pt > 73pt \cellcolor{pastelgreen}{#1}%
    \else\ifdim #1 pt > 70pt \cellcolor{pastelyellow}{#1}%
    \else\ifdim #1 pt > 67pt \cellcolor{pastelorange}{#1}%
    \else \cellcolor{pastelred}{#1}%
    \fi\fi\fi
}

\begin{table}[ht]
\centering
\caption{\emph{Comparison of \OUR with and without Matryoshka representation learning} on ImageNet-1K (MAE ViT-B). Without Matryoshka, probing at $D$ ($D_o{=}D_i$) achieves the best accuracy but collapses at smaller dimensions. 
Efficient Matryoshka enables multi-scale probing without extra parameters, improving performance at reduced dimensions with relatively small degradation at $D$. Vanilla Matryoshka partly restores performance at $D$ but at the cost of additional parameters.}
\vspace{-6pt}
\label{tab:matryoshka}
\scriptsize
\resizebox{\linewidth}{!}{%
{\renewcommand{\arraystretch}{1.2}%
\begin{tabular}{l|cccc|cccc|cc}
\toprule
& \multicolumn{4}{c|}{\textsc{Without Matryoshka}} 
& \multicolumn{4}{c|}{\textsc{Efficient Matryoshka}} 
& \multicolumn{2}{c}{\textsc{Vanilla Matryoshka}} \\
\cline{2-5}\cline{6-9}\cline{10-11}
& $D$ & $D/2$ & $D/4$ & $D/8$
& \multicolumn{4}{c|}{$D$, $D/2$, $D/4$, $D/8$}
& \multicolumn{2}{c}{$D$, $D/2$, $D/4$, $D/8$} \\
\cline{2-5}\cline{6-9}\cline{10-11}
\makecell{\textsc{Eval. on}} 
&  &  &  &
& $\lambda{=}0.8$ & $\lambda{=}0.6$ & $\lambda{=}0.4$ & $\lambda{=}0.25$
& \hspace{3mm}$\lambda{=}0.6$ & $\lambda{=}0.25$ \\
\midrule
$D$   & \textbf{75.6} & --    & --    & --    & \heat{75.0} & \heat{74.7} & \heat{74.3} & \heat{73.9} & \hspace{3mm}75.4 & 75.2 \\
$D/2$ & 50.7 & \textbf{74.4} & --    & --    & \heat{72.2} & \heat{73.1} & \heat{73.6} & \heat{73.8} & \hspace{3mm}73.8 & 74.4 \\
$D/4$ & 20.1 & -- & \textbf{72.4} & --    & \heat{69.5} & \heat{70.9} & \heat{71.4} & \heat{71.6} & \hspace{3mm}71.7 & 72.2 \\
$D/8$ & 5.5  & -- & -- & \textbf{69.8} & \heat{65.7} & \heat{67.3} & \heat{67.9} & \heat{68.2} & \hspace{3mm}68.6 & 68.9 \\
\midrule
% Params (two-line block)
\multirow{2}{*}{\textsc{\# Par.}}
& 1.4M & 1.4M & 1.4M & 1.4M
& \multicolumn{4}{c|}{\multirow{2}{*}{\centering 1.4M}}
& \multicolumn{2}{c}{\multirow{2}{*}{\centering 2.1M}} \\
\cline{2-5} % short rule only under the 'Without' block
& \multicolumn{4}{c|}{5.5M}
& \multicolumn{4}{c|}{}
& \multicolumn{2}{c}{} \\
\bottomrule
\end{tabular}
}}
\end{table}
\begin{table}[h]
\centering
\caption{\emph{Comparison of LP with and without Matryoshka representation learning} on ImageNet-1K (MAE ViT-B). Without Matryoshka, linear probing collapses sharply at reduced output dimensionalities. Matryoshka improves low-dimensional performance, without increasing parameters, though the accuracy at $D$ decreases as $\lambda$ (the weight of the full-dimensional loss) is reduced.}
\vspace{-6pt}
\label{tab:matryoshka_lp}
\scriptsize
\begin{tabular}{l|c|cccc}
\toprule
\multirow{2}{*}{\textsc{Eval. on}} & \multirow{2}{*}{\textsc{Without Matryoshka}} & \multicolumn{4}{c}{\textsc{Efficient Matryoshka}} \\
\cmidrule{3-6}
 &  & $\lambda=0.8$ & $\lambda=0.6$ & $\lambda=0.4$ & $\lambda=0.25$ \\
\midrule
$D$   & \textbf{66.4} & 66.1  & 65.8 & 65.4 & 65.0 \\
$D/2$ & 34.5 & 59.2 & 60.6 & 61.2  & 61.4 \\
$D/4$ & 10.8 & 52.5 & 54.5 & 55.3 & 55.7 \\
$D/8$ & 1.4  & 43.4 & 46.1 & 47.0 & 47.4 \\
\bottomrule
\end{tabular}
\end{table}

%------------------------------------------------------------------

\paragraph{Visualizations.}

%------------------------------------------------------------------------------
\definecolor{darkred}{rgb}{0.6, 0, 0}
\definecolor{purple}{rgb}{0.5, 0, 0.5}
\definecolor{brown}{rgb}{0.6, 0.3, 0}
\setlength{\fboxrule}{1.3pt}

\begin{figure*}
\scriptsize
\centering
\setlength{\tabcolsep}{1.2pt}
\newcommand{\sz}{1.36cm}
\begin{tabular}{cccccccc}

original &
EP\textsubscript{1} &
EP\textsubscript{2} &
EP\textsubscript{2} &
EP\textsubscript{4} &
EP\textsubscript{4} &
EP\textsubscript{4} &
EP\textsubscript{4} \\

image &
$\vq_1$&
$\vq_1$&
$\vq_2$&
$\vq_1$&
$\vq_2$&
$\vq_3$&
$\vq_4$\\

\includegraphics[width=\sz,height=\sz]{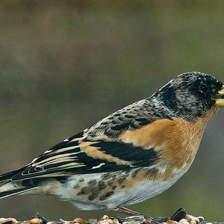} &
\fcolorbox{brown}{white}{\includegraphics[width=\sz,height=\sz]{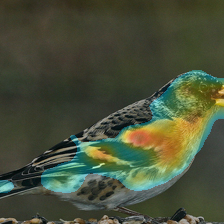}} &
\fcolorbox{purple}{white}{\includegraphics[width=\sz,height=\sz]{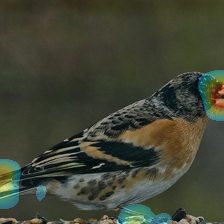}} &
\fcolorbox{purple}{white}{\includegraphics[width=\sz,height=\sz]{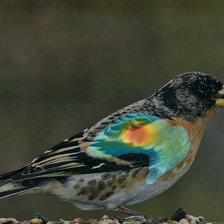}} &
\fcolorbox{orange}{white}{\includegraphics[width=\sz,height=\sz]{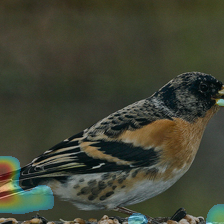}} &
\fcolorbox{orange}{white}{\includegraphics[width=\sz,height=\sz]{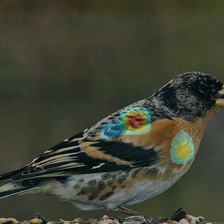}} &
\fcolorbox{orange}{white}{\includegraphics[width=\sz,height=\sz]{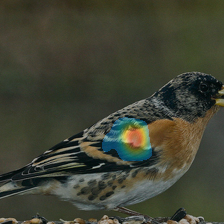}} &
\fcolorbox{orange}{white}{\includegraphics[width=\sz,height=\sz]{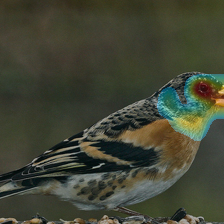}} \\

EP\textsubscript{8} &
EP\textsubscript{8} &
EP\textsubscript{8} &
EP\textsubscript{8} &
EP\textsubscript{8} &
EP\textsubscript{8} &
EP\textsubscript{8} &
EP\textsubscript{8} \\

$\vq_1$ &
$\vq_2$&
$\vq_3$&
$\vq_4$&
$\vq_5$&
$\vq_6$&
$\vq_7$&
$\vq_8$ \\

\fcolorbox{blue}{white}{\includegraphics[width=\sz,height=\sz]{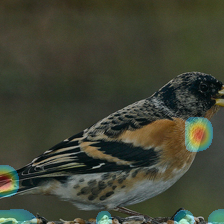}} &
\fcolorbox{blue}{white}{\includegraphics[width=\sz,height=\sz]{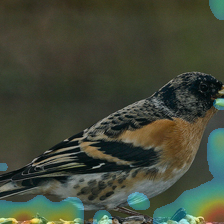}} &
\fcolorbox{blue}{white}{\includegraphics[width=\sz,height=\sz]{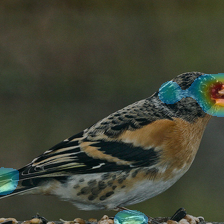}} &
\fcolorbox{blue}{white}{\includegraphics[width=\sz,height=\sz]{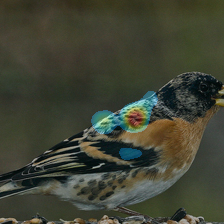}} &
\fcolorbox{blue}{white}{\includegraphics[width=\sz,height=\sz]{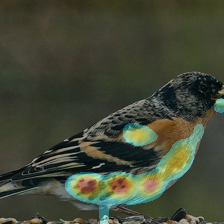}} &
\fcolorbox{blue}{white}{\includegraphics[width=\sz,height=\sz]{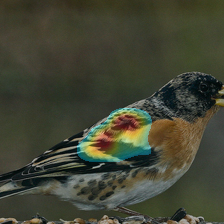}} &
\fcolorbox{blue}{white}{\includegraphics[width=\sz,height=\sz]{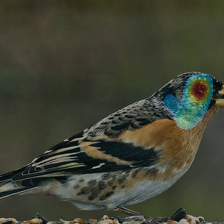}} &
\fcolorbox{blue}{white}{\includegraphics[width=\sz,height=\sz]{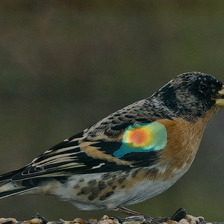}} \\

EP\textsubscript{16} &
EP\textsubscript{16} &
EP\textsubscript{16} &
EP\textsubscript{16} &
EP\textsubscript{16} &
EP\textsubscript{16} &
EP\textsubscript{16} &
EP\textsubscript{16} \\

$\vq_1$ &
$\vq_2$&
$\vq_3$&
$\vq_4$&
$\vq_5$&
$\vq_6$&
$\vq_7$&
$\vq_8$ \\

\fcolorbox{darkred}{white}{\includegraphics[width=\sz,height=\sz]{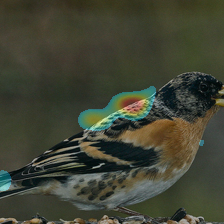}} &
\fcolorbox{darkred}{white}{\includegraphics[width=\sz,height=\sz]{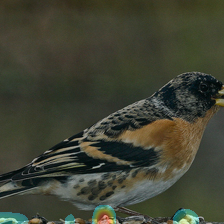}} &
\fcolorbox{darkred}{white}{\includegraphics[width=\sz,height=\sz]{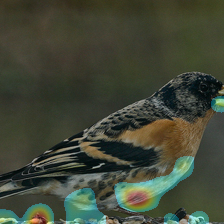}} &
\fcolorbox{darkred}{white}{\includegraphics[width=\sz,height=\sz]{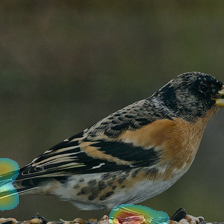}} &
\fcolorbox{darkred}{white}{\includegraphics[width=\sz,height=\sz]{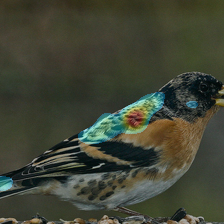}} &
\fcolorbox{darkred}{white}{\includegraphics[width=\sz,height=\sz]{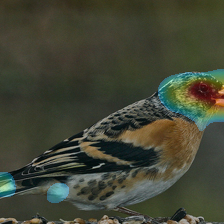}} &
\fcolorbox{darkred}{white}{\includegraphics[width=\sz,height=\sz]{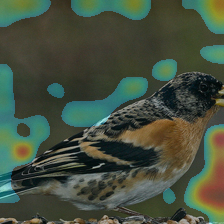}} &
\fcolorbox{darkred}{white}{\includegraphics[width=\sz,height=\sz]{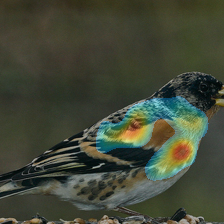}} \\

EP\textsubscript{16} &
EP\textsubscript{16} &
EP\textsubscript{16} &
EP\textsubscript{16} &
EP\textsubscript{16} &
EP\textsubscript{16} &
EP\textsubscript{16} &
EP\textsubscript{16} \\

$\vq_9$ &
$\vq_{10}$&
$\vq_{11}$&
$\vq_{12}$&
$\vq_{13}$&
$\vq_{14}$&
$\vq_{15}$&
$\vq_{16}$ \\

\fcolorbox{darkred}{white}{\includegraphics[width=\sz,height=\sz]{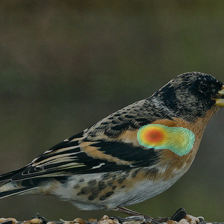}} &
\fcolorbox{darkred}{white}{\includegraphics[width=\sz,height=\sz]{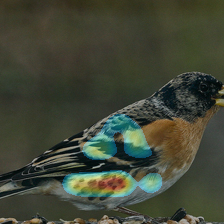}} &
\fcolorbox{darkred}{white}{\includegraphics[width=\sz,height=\sz]{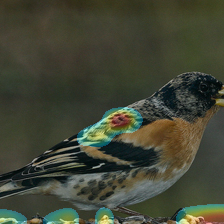}} &
\fcolorbox{darkred}{white}{\includegraphics[width=\sz,height=\sz]{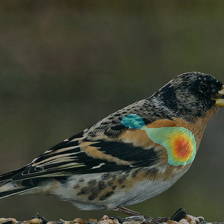}} &
\fcolorbox{darkred}{white}{\includegraphics[width=\sz,height=\sz]{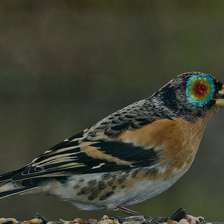}} &
\fcolorbox{darkred}{white}{\includegraphics[width=\sz,height=\sz]{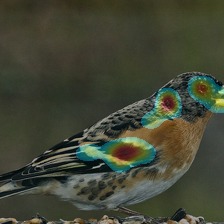}} &
\fcolorbox{darkred}{white}{\includegraphics[width=\sz,height=\sz]{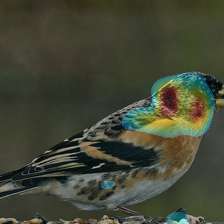}} &
\fcolorbox{darkred}{white}{\includegraphics[width=\sz,height=\sz]{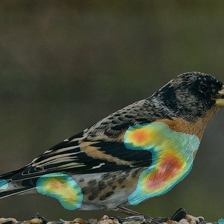}} \\

\end{tabular}
\caption{\emph{Attention maps of \our (\OUR)} variants grouped by the number of queries (1, 2, etc.). MAE ViT-B pre-trained on ImageNet-1K, probed with \OUR. Images: ImageNet-1k validation set.}
\label{fig:attention-maps-ours2}
\vspace{-6pt}
\end{figure*}

%------------------------------------------------------------------------------
\begin{figure*}
\scriptsize
\centering
\setlength{\tabcolsep}{1.2pt}
\newcommand{\sz}{2cm}
\begin{tabular}{ccccc}

\includegraphics[width=\sz,height=\sz]{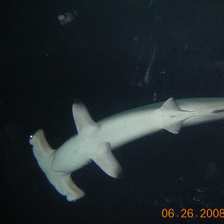} &
\includegraphics[width=\sz,height=\sz]{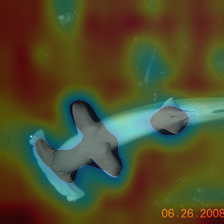} &
\includegraphics[width=\sz,height=\sz]{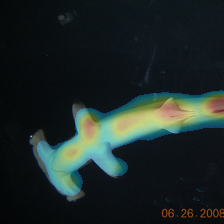} &
\includegraphics[width=\sz,height=\sz]{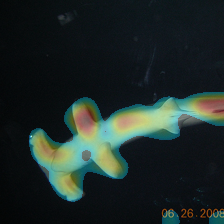} &
\includegraphics[width=\sz,height=\sz]{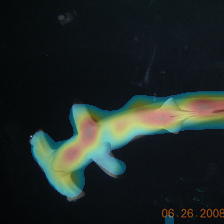} \\

\includegraphics[width=\sz,height=\sz]{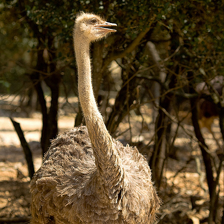} &
\includegraphics[width=\sz,height=\sz]{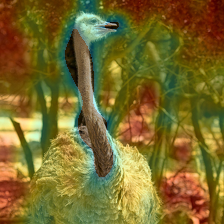} &
\includegraphics[width=\sz,height=\sz]{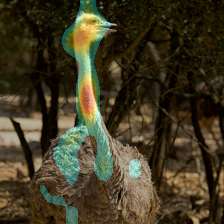} &
\includegraphics[width=\sz,height=\sz]{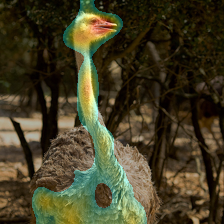} &
\includegraphics[width=\sz,height=\sz]{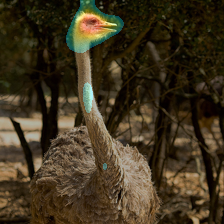} \\

\includegraphics[width=\sz,height=\sz]{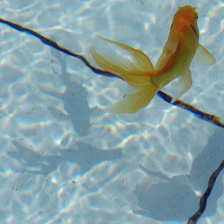} &
\includegraphics[width=\sz,height=\sz]{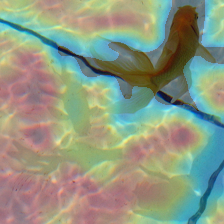} &
\includegraphics[width=\sz,height=\sz]{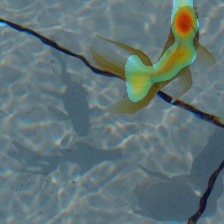} &
\includegraphics[width=\sz,height=\sz]{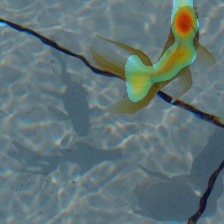} &
\includegraphics[width=\sz,height=\sz]{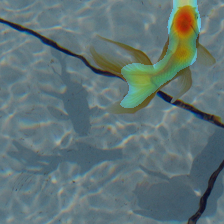} \\

\includegraphics[width=\sz,height=\sz]{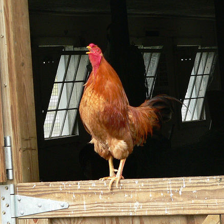} &
\includegraphics[width=\sz,height=\sz]{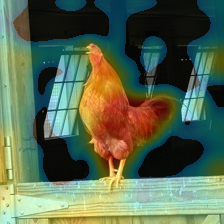} &
\includegraphics[width=\sz,height=\sz]{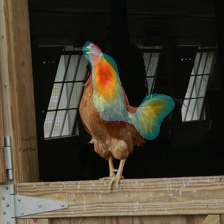} &
\includegraphics[width=\sz,height=\sz]{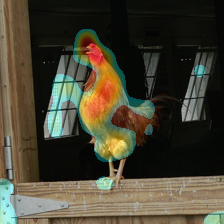} &
\includegraphics[width=\sz,height=\sz]{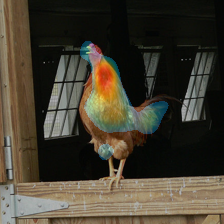} \\

\includegraphics[width=\sz,height=\sz]{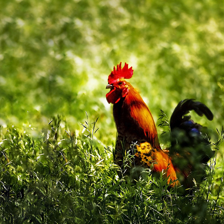} &
\includegraphics[width=\sz,height=\sz]{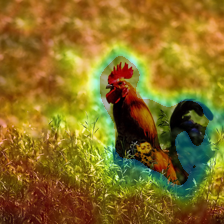} &
\includegraphics[width=\sz,height=\sz]{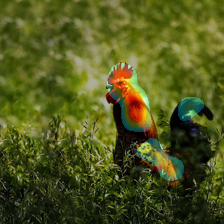} &
\includegraphics[width=\sz,height=\sz]{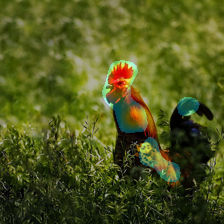} &
\includegraphics[width=\sz,height=\sz]{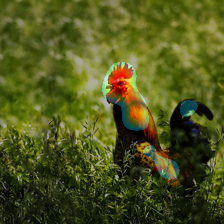} \\

\includegraphics[width=\sz,height=\sz]{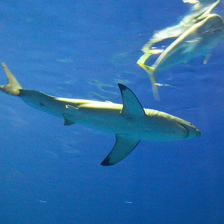} &
\includegraphics[width=\sz,height=\sz]{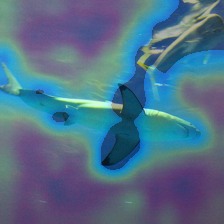} &
\includegraphics[width=\sz,height=\sz]{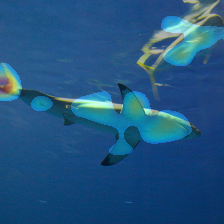} &
\includegraphics[width=\sz,height=\sz]{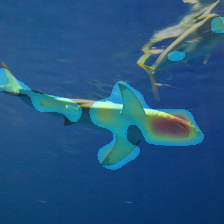} &
\includegraphics[width=\sz,height=\sz]{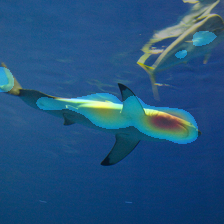} \\

\includegraphics[width=\sz,height=\sz]{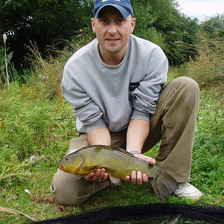} &
\includegraphics[width=\sz,height=\sz]{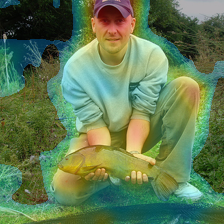} &
\includegraphics[width=\sz,height=\sz]{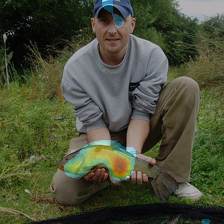} &
\includegraphics[width=\sz,height=\sz]{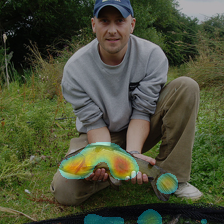} &
\includegraphics[width=\sz,height=\sz]{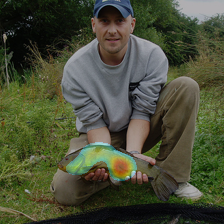} \\

\includegraphics[width=\sz,height=\sz]{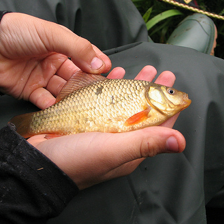} &
\includegraphics[width=\sz,height=\sz]{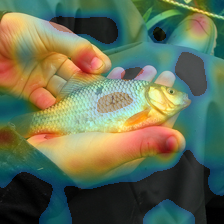} &
\includegraphics[width=\sz,height=\sz]{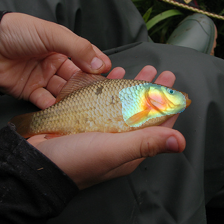} &
\includegraphics[width=\sz,height=\sz]{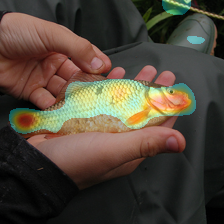} &
\includegraphics[width=\sz,height=\sz]{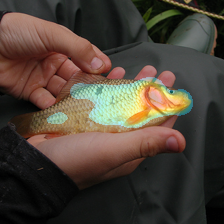} \\

\includegraphics[width=\sz,height=\sz]{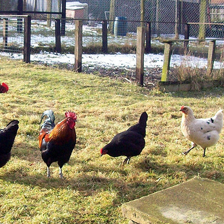} &
\includegraphics[width=\sz,height=\sz]{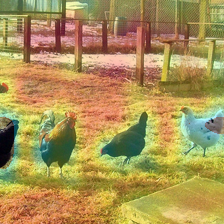} &
\includegraphics[width=\sz,height=\sz]{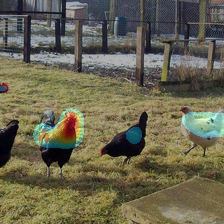} &
\includegraphics[width=\sz,height=\sz]{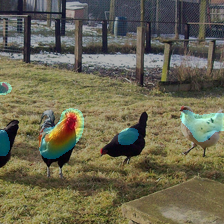} &
\includegraphics[width=\sz,height=\sz]{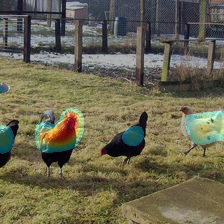} \\

\includegraphics[width=\sz,height=\sz]{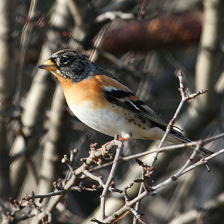} &
\includegraphics[width=\sz,height=\sz]{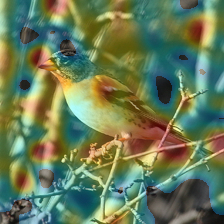} &
\includegraphics[width=\sz,height=\sz]{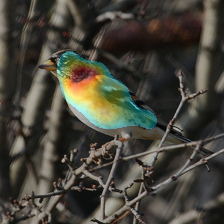} &
\includegraphics[width=\sz,height=\sz]{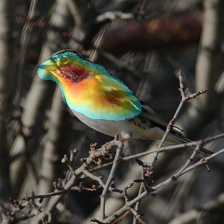} &
\includegraphics[width=\sz,height=\sz]{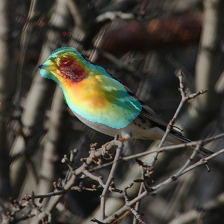} \\

input &
CBAM &
AbMILP &
DELF &
SimPool\\

image &
 &
 &
 &
\\

\end{tabular}
\caption{\emph{Attention maps of single-head attention pooling methods}. MAE ViT-B pre-trained on ImageNet-1k. Images: ImageNet-1k validation set.}
\label{fig:attention-maps-singlehead}
\end{figure*}
%------------------------------------------------------------------------------
%------------------------------------------------------------------------------
\begin{figure*}
\scriptsize
\centering
\setlength{\tabcolsep}{1.2pt}
\newcommand{\sz}{1.25cm}
\begin{tabular}{ccccccccccc}

\rotatebox{90}{mean} &
\includegraphics[width=\sz,height=\sz]{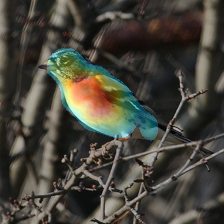} &
\includegraphics[width=\sz,height=\sz]{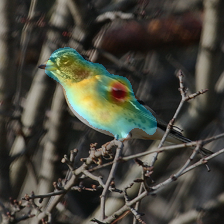} &
\includegraphics[width=\sz,height=\sz]{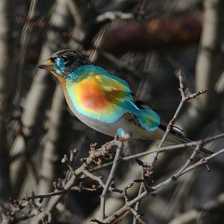} &
\includegraphics[width=\sz,height=\sz]{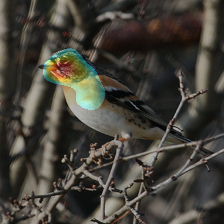} &
\includegraphics[width=\sz,height=\sz]{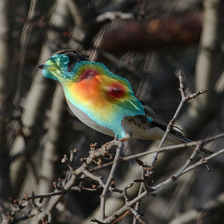} &
\includegraphics[width=\sz,height=\sz]{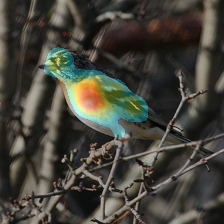} &
\includegraphics[width=\sz,height=\sz]{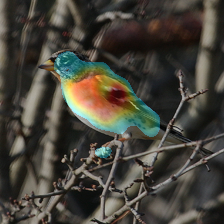} &
\includegraphics[width=\sz,height=\sz]{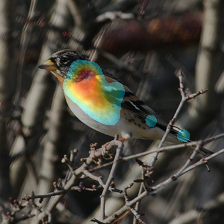} &
\includegraphics[width=\sz,height=\sz]{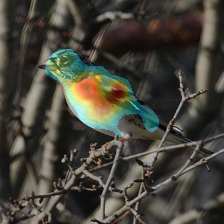} &
\includegraphics[width=\sz,height=\sz]{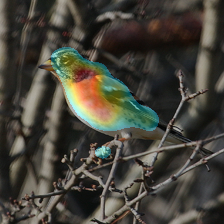}\\

\rotatebox{90}{std} &
\includegraphics[width=\sz,height=\sz]{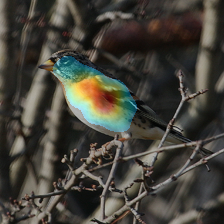} &
\includegraphics[width=\sz,height=\sz]{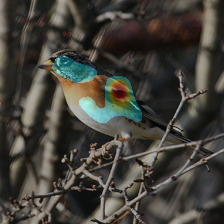} &
\includegraphics[width=\sz,height=\sz]{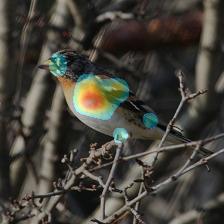} &
\includegraphics[width=\sz,height=\sz]{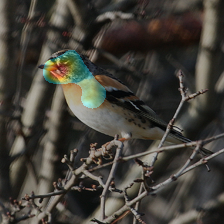} &
\includegraphics[width=\sz,height=\sz]{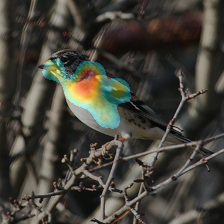} &
\includegraphics[width=\sz,height=\sz]{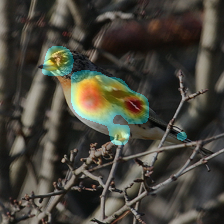} &
\includegraphics[width=\sz,height=\sz]{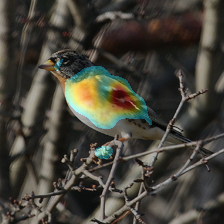} &
\includegraphics[width=\sz,height=\sz]{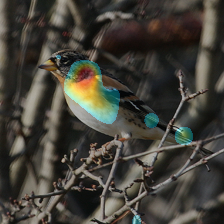} &
\includegraphics[width=\sz,height=\sz]{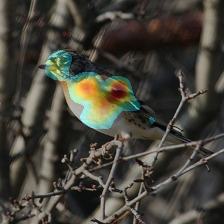}&
\includegraphics[width=\sz,height=\sz]{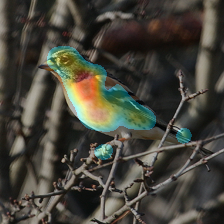} \\

\rotatebox{90}{min} &
\includegraphics[width=\sz,height=\sz]{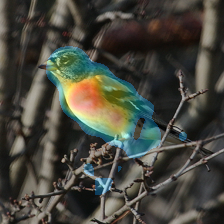} &
\includegraphics[width=\sz,height=\sz]{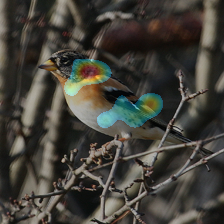} &
\includegraphics[width=\sz,height=\sz]{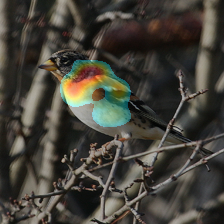} &
\includegraphics[width=\sz,height=\sz]{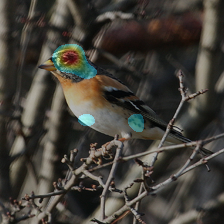} &
\includegraphics[width=\sz,height=\sz]{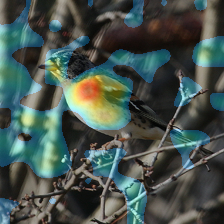} &
\includegraphics[width=\sz,height=\sz]{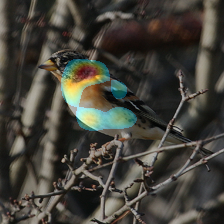} &
\includegraphics[width=\sz,height=\sz]{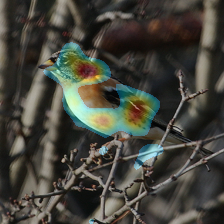} &
\includegraphics[width=\sz,height=\sz]{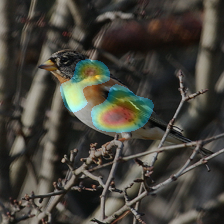} &
\includegraphics[width=\sz,height=\sz]{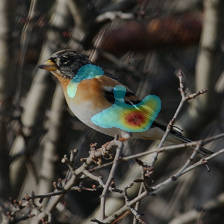}&
\includegraphics[width=\sz,height=\sz]{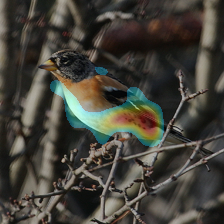} \\

\rotatebox{90}{max} &
\includegraphics[width=\sz,height=\sz]{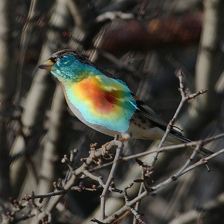} &
\includegraphics[width=\sz,height=\sz]{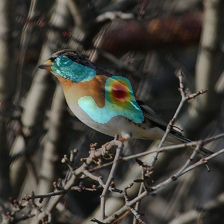} &
\includegraphics[width=\sz,height=\sz]{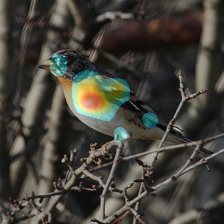} &
\includegraphics[width=\sz,height=\sz]{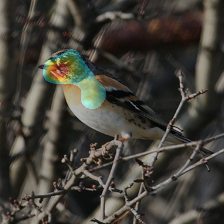} &
\includegraphics[width=\sz,height=\sz]{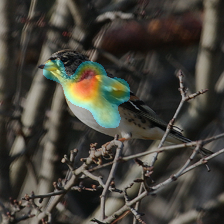} &
\includegraphics[width=\sz,height=\sz]{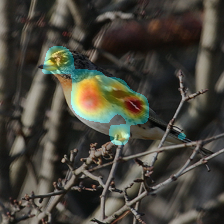} &
\includegraphics[width=\sz,height=\sz]{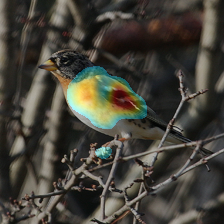} &
\includegraphics[width=\sz,height=\sz]{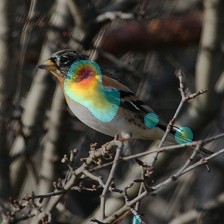} &
\includegraphics[width=\sz,height=\sz]{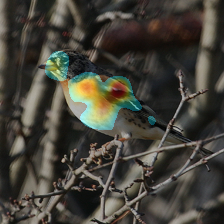}&
\includegraphics[width=\sz,height=\sz]{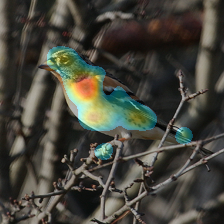}\vspace{3pt}\\

%---------------------------------------------------------------------------------------
\rotatebox{90}{mean} &
\includegraphics[width=\sz,height=\sz]{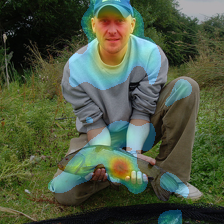} &
\includegraphics[width=\sz,height=\sz]{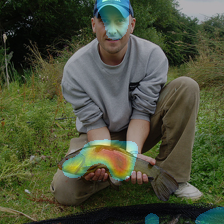} &
\includegraphics[width=\sz,height=\sz]{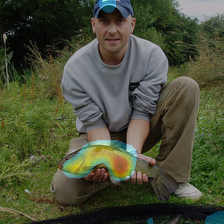} &
\includegraphics[width=\sz,height=\sz]{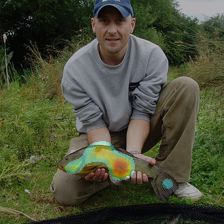} &
\includegraphics[width=\sz,height=\sz]{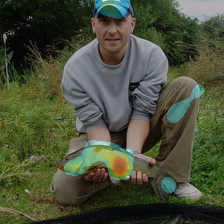} &
\includegraphics[width=\sz,height=\sz]{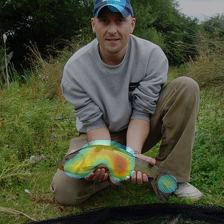} &
\includegraphics[width=\sz,height=\sz]{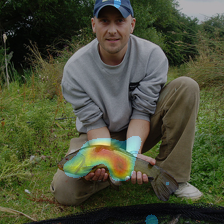} &
\includegraphics[width=\sz,height=\sz]{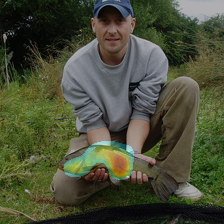} &
\includegraphics[width=\sz,height=\sz]{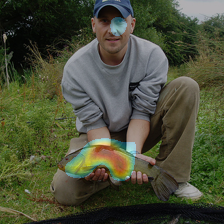}&
\includegraphics[width=\sz,height=\sz]{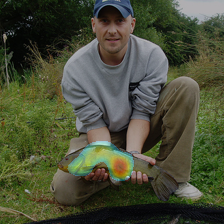} \\

\rotatebox{90}{std} &
\includegraphics[width=\sz,height=\sz]{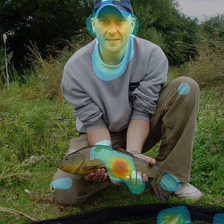} &
\includegraphics[width=\sz,height=\sz]{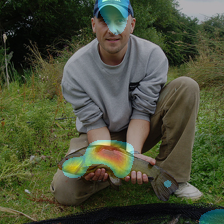} &
\includegraphics[width=\sz,height=\sz]{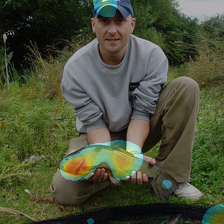} &
\includegraphics[width=\sz,height=\sz]{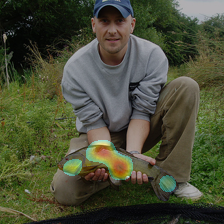} &
\includegraphics[width=\sz,height=\sz]{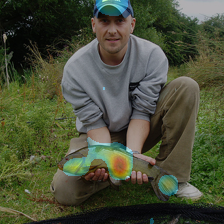} &
\includegraphics[width=\sz,height=\sz]{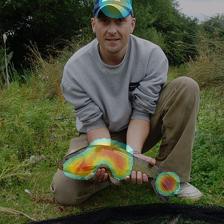} &
\includegraphics[width=\sz,height=\sz]{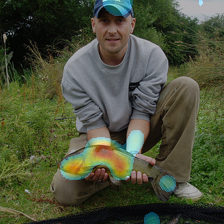} &
\includegraphics[width=\sz,height=\sz]{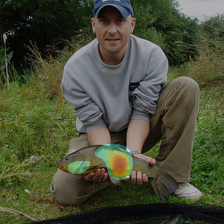} &
\includegraphics[width=\sz,height=\sz]{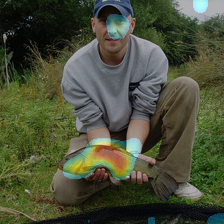}&
\includegraphics[width=\sz,height=\sz]{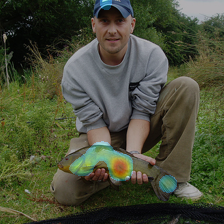} \\

\rotatebox{90}{min} &
\includegraphics[width=\sz,height=\sz]{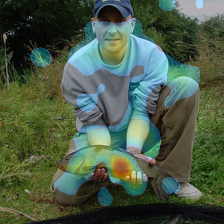} &
\includegraphics[width=\sz,height=\sz]{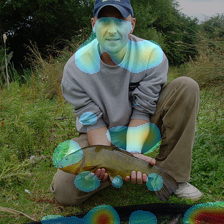} &
\includegraphics[width=\sz,height=\sz]{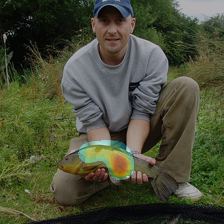} &
\includegraphics[width=\sz,height=\sz]{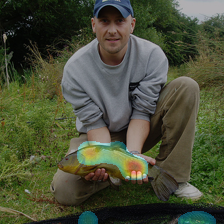} &
\includegraphics[width=\sz,height=\sz]{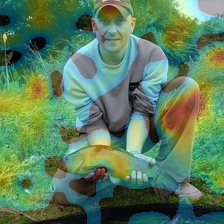} &
\includegraphics[width=\sz,height=\sz]{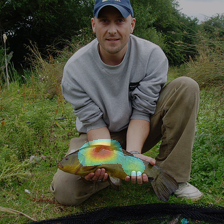} &
\includegraphics[width=\sz,height=\sz]{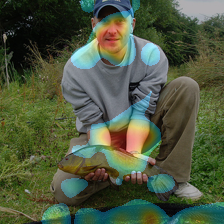} &
\includegraphics[width=\sz,height=\sz]{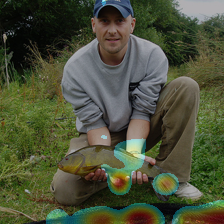} &
\includegraphics[width=\sz,height=\sz]{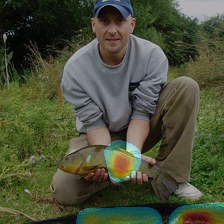}&
\includegraphics[width=\sz,height=\sz]{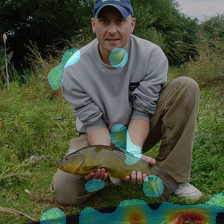} \\

\rotatebox{90}{max} &
\includegraphics[width=\sz,height=\sz]{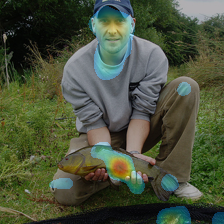} &
\includegraphics[width=\sz,height=\sz]{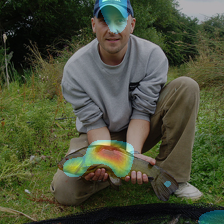} &
\includegraphics[width=\sz,height=\sz]{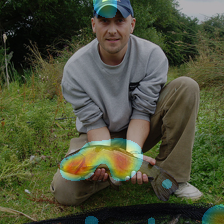} &
\includegraphics[width=\sz,height=\sz]{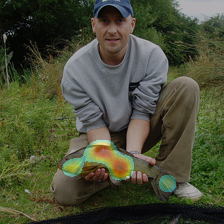} &
\includegraphics[width=\sz,height=\sz]{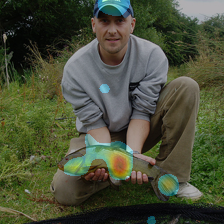} &
\includegraphics[width=\sz,height=\sz]{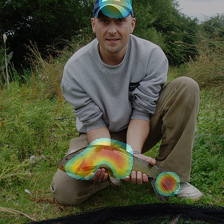} &
\includegraphics[width=\sz,height=\sz]{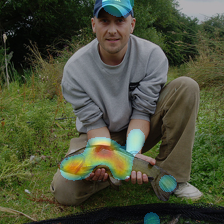} &
\includegraphics[width=\sz,height=\sz]{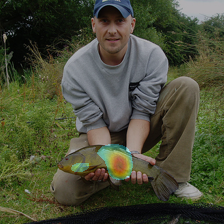} &
\includegraphics[width=\sz,height=\sz]{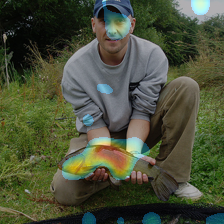}&
\includegraphics[width=\sz,height=\sz]{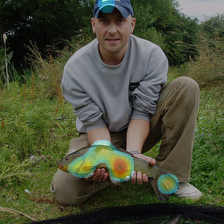} \\

&
\cls &
CAE &
CaiT &
CLIP &
CoCa &
ViT &
V-JEPA &
SigLIP &
AIM &
EP$_{16}$\\

&
 &
 &
 &
 &
 &
 &
 &
 &
 &
 \\

\end{tabular}
\vspace{-10pt}
\caption{\emph{Attention maps of multi-head attention pooling methods} for different attention predictor aggregators: mean, standard deviation (std), minimum (min), and maximum (max). MAE ViT-B pre-trained on ImageNet-1K. Images: ImageNet-1k validation set. EP$_{16}$: \our (\OUR) with 16 queries.}
\label{fig:attention-maps-multihead}
\end{figure*}
%------------------------------------------------------------------------------
%------------------------------------------------------------------------------
\begin{figure*}
\scriptsize
\centering
\setlength{\tabcolsep}{1.2pt}
\newcommand{\sz}{1.5cm}
\begin{tabular}{cccccccc}

\includegraphics[width=\sz,height=\sz]{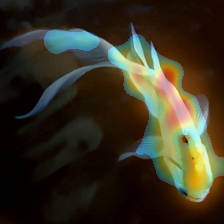} &
\includegraphics[width=\sz,height=\sz]{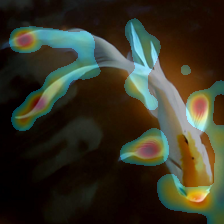} &
\includegraphics[width=\sz,height=\sz]{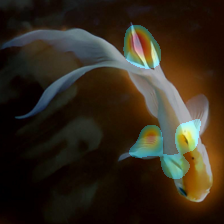} &
\includegraphics[width=\sz,height=\sz]{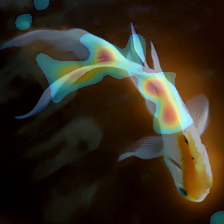} &
\includegraphics[width=\sz,height=\sz]{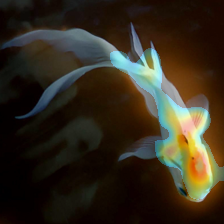} &
\includegraphics[width=\sz,height=\sz]{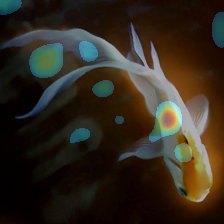} &
\includegraphics[width=\sz,height=\sz]{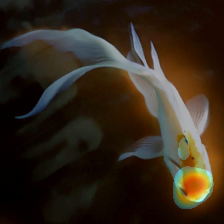} &
\includegraphics[width=\sz,height=\sz]{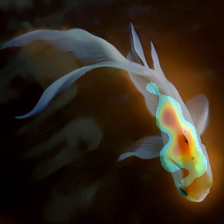} \\

\includegraphics[width=\sz,height=\sz]{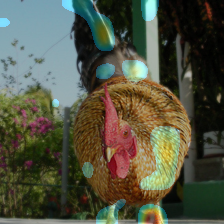} &
\includegraphics[width=\sz,height=\sz]{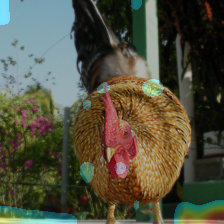} &
\includegraphics[width=\sz,height=\sz]{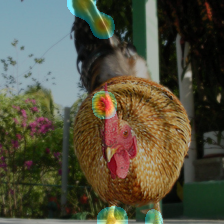} &
\includegraphics[width=\sz,height=\sz]{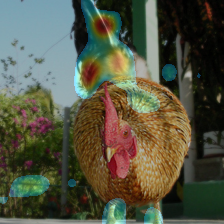} &
\includegraphics[width=\sz,height=\sz]{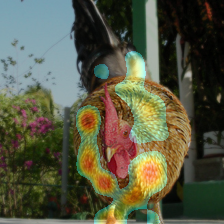} &
\includegraphics[width=\sz,height=\sz]{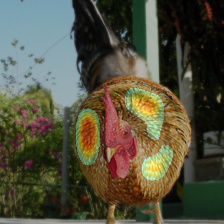} &
\includegraphics[width=\sz,height=\sz]{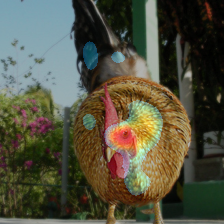} &
\includegraphics[width=\sz,height=\sz]{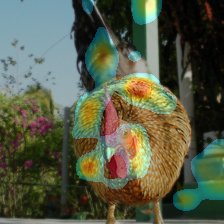} \\

\includegraphics[width=\sz,height=\sz]{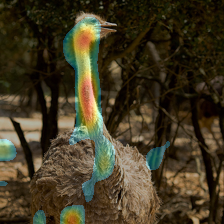} &
\includegraphics[width=\sz,height=\sz]{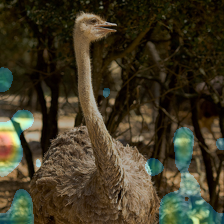} &
\includegraphics[width=\sz,height=\sz]{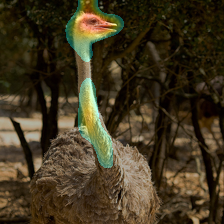} &
\includegraphics[width=\sz,height=\sz]{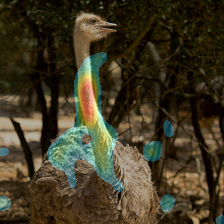} &
\includegraphics[width=\sz,height=\sz]{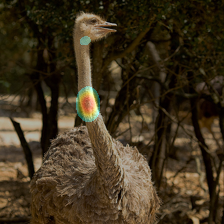} &
\includegraphics[width=\sz,height=\sz]{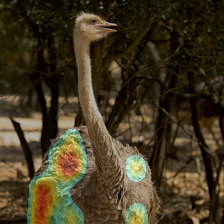} &
\includegraphics[width=\sz,height=\sz]{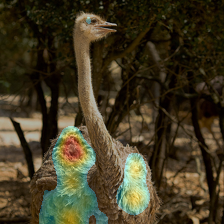} &
\includegraphics[width=\sz,height=\sz]{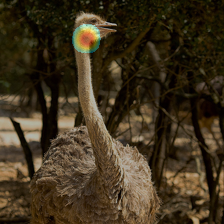} \\

\includegraphics[width=\sz,height=\sz]{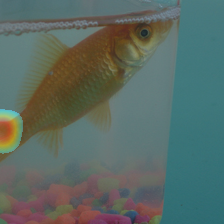} &
\includegraphics[width=\sz,height=\sz]{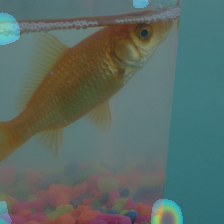} &
\includegraphics[width=\sz,height=\sz]{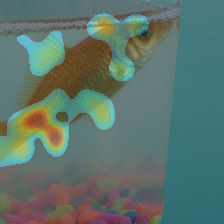} &
\includegraphics[width=\sz,height=\sz]{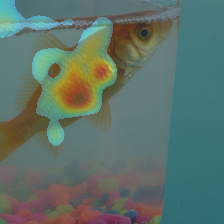} &
\includegraphics[width=\sz,height=\sz]{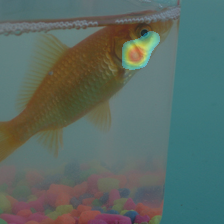} &
\includegraphics[width=\sz,height=\sz]{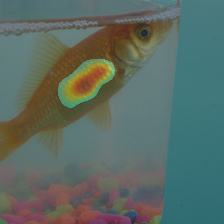} &
\includegraphics[width=\sz,height=\sz]{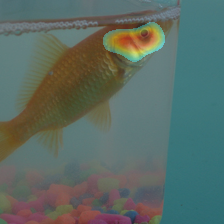} &
\includegraphics[width=\sz,height=\sz]{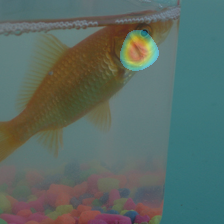} \\

$\vq_1$ &
$\vq_2$ &
$\vq_3$ &
$\vq_4$ &
$\vq_5$ &
$\vq_6$ &
$\vq_7$ &
$\vq_8$\\

\end{tabular}
\vspace{-2pt}
\caption{\emph{Attention maps of \our (\OUR)} with 8 queries. Each query $\vq_i$ learns to focus on distinct and complementary regions, capturing diverse spatial and semantic information. MAE ViT-B pre-trained on ImageNet-1K, probed with \OUR. Images: ImageNet-1k validation set.}
\label{fig:attention-maps-ours}
\end{figure*}
%------------------------------------------------------------------------------

%------------------------------------------------------------------
%\begin{figure*}[t]
%\vspace{-6pt}
%\centering
%\scriptsize
%\input{tab/imagenet_val_attmaps_q8_supp}
%\vspace{-4pt}
%\caption{\emph{Attention maps of} $\mOUR_{8}$. Joint visualization of 8 query attention distributions on diverse bird images, where each query is assigned a distinct color. Consistent semantic correspondences emerge (\eg, \textcolor{qcol3}{tails}, \textcolor{qcol4}{beaks}, \textcolor{qcol2}{feet}), with each query capturing complementary regions and enabling a structured decomposition of visual cues. MAE ViT-B pre-trained on IN-1K, probed with EP.}
%\label{fig:attention-maps-ours-dogs}
%\end{figure*}
%------------------------------------------------------------------

To better understand the behavior of different attentive pooling/probing methods, we present visualizations of attention maps across various configurations.

\autoref{fig:attention-maps-ours2} explores the effect of varying the number of queries in \OUR, visualizing configurations with 1, 2, 4, and more queries. When only a single query (EP$_1$) is used, the attention map tends to capture a coarse, global representation of the object. As the number of queries increases, the attention becomes more fine-grained and spatially distributed, with each query specializing in distinct object regions. This highlights the flexibility of \OUR in controlling the granularity of attention: fewer queries encourage holistic coverage, while more queries promote detailed, part-based localization.

\autoref{fig:attention-maps-singlehead} shows the attention maps obtained from four single-head attention probing methods (CBAM, AbMILP, DELF, and SimPool) using an ImageNet-1K pretrained MAE ViT-B model. Among them, CBAM exhibits poor localization, often failing to focus on the target object, which is consistent with its low classification accuracy across datasets. In contrast, AbMILP, DELF, and SimPool produce more precise and meaningful attention, highlighting relevant object regions while suppressing background noise. Due to their single-head nature, these methods are compelled to concentrate all semantic information into a single attention vector, which encourages a global view of the input image rather than fine-grained discrimination.

\autoref{fig:attention-maps-multihead} compares attention maps from multi-head probing methods. Rather than visualizing just the average attention across heads,which can obscure useful per-head behavior, we show the minimum, maximum, and standard deviation across attention heads. The first column contains maps from the \cls token of the pretrained MAE ViT-B model. The remaining columns display maps from CAE, CaiT, CLIP, CoCa, ViT, V-JEPA, SigLIP, and AIM, alongside \OUR using 16 learnable queries (EP$_{16}$). Notably, \OUR produces high-quality attention maps that rival the best-performing methods in both clarity and relevance, while retaining computational efficiency.

\autoref{fig:attention-maps-ours} presents the attention maps corresponding to each individual query in \OUR. We observe that each query $\vq_i$ attends to distinct, complementary regions of the object (\eg, head, torso, boundaries), illustrating how \OUR distributes attention cooperatively across salient features without redundancy. This diversity among queries reveals the model’s capacity to decompose complex objects into meaningful sub-parts.

\section{Limitations and Future Work}
\label{sec:limitations}

While our study provides the first systematic benchmark of attentive probing and introduces a lightweight yet effective alternative, several limitations remain. First, our evaluation focuses exclusively on frozen backbones. Although this setting isolates the effect of probing mechanisms, it leaves open how efficient probing might interact with lightweight fine-tuning strategies or adapter-based methods. Exploring the synergy between probing and parameter-efficient fine-tuning could offer a broader view of scalable evaluation.

Second, our current experimental protocol always performs a full forward pass of the backbone during training, even though the backbone is frozen. In practice, one could pre-compute and store patch-token features once and run probing purely on top of these cached features. Systematically studying this “frozen-feature’’ protocol—its memory/IO trade-offs, impact on optimization dynamics, and applicability across architectures—is an interesting direction for future work.

Third, while we introduced complementarity metrics (average and max similarity) to study the diversity of attention predictors, our analysis remains largely diagnostic. A deeper theoretical understanding of why attentive probing tends to yield more complementary maps—and whether this property can be explicitly optimized—could further extend probing beyond evaluation into representation refinement.

Fourth, our experiments concentrate on image classification benchmarks. Attentive probing may also benefit other tasks that naturally require part-level reasoning, such as detection, segmentation, or retrieval, where its complementary attention maps could act as implicit part detectors. Extending probing to such structured tasks is a promising direction.

Fifth, although we examined variants such as attention mixing and Matryoshka probing, both were limited to ImageNet-scale experiments. Their potential on larger models and multimodal settings (e.g., vision–language tasks) remains underexplored. Similarly, the alternative protocol of probing with pre-stored frozen features, requiring only one backbone forward pass, is left for future work.

Finally, the broader implications of probing as more than an evaluation protocol deserve attention. Our results suggest that attentive probing exhibits emerging properties—such as diversity and interpretability—that are not trivially inherited from the backbone. Understanding whether these properties generalize across modalities and can be exploited for tasks like explainability, robustness, or adaptive computation opens up an exciting line of future research.

\section{Use of Large Language Models}

This paper has made limited use of large language models (LLMs), specifically to aid in the polishing and refinement of writing. LLMs were not used for ideation, technical contributions, experimental design, analysis, or related work retrieval. All research ideas, methodology, experiments, and conclusions presented are solely the work of the authors.

\end{document}